\documentclass[letterpaper, 10 pt, conference]{ieeeconf}  

\IEEEoverridecommandlockouts                              

\usepackage{cite}
\usepackage{amsmath}
\usepackage{amssymb}
\usepackage{array}
\usepackage{subfiles} 
\usepackage{algorithm}
\usepackage{subfigure}
\usepackage{algorithm}
\usepackage{algpseudocode}
\usepackage{graphicx}
\usepackage{textcomp}
\usepackage{caption}

\usepackage{booktabs}
\usepackage{multirow}
\usepackage{cancel}

\newcommand{\multiline}[1]{%
	\begin{tabularx}{\dimexpr\linewidth-\ALG@thistlm}[t]{@{}X@{}}
		#1
	\end{tabularx}
}

\newlength\myindent
\newcommand{\argmax}{\operatornamewithlimits{argmax}}
\newcommand{\argmin}{\operatornamewithlimits{argmin}}

\newcommand{\vectori}{\operatornamewithlimits{vec}}

\title{\LARGE \bf
	Online Gaussian Process State-Space Model: Learning and Planning for Partially Observable Dynamical Systems
}

\author{Soon-Seo~Park\authorrefmark{1}, Young-Jin~Park\authorrefmark{2}, Youngjae Min\authorrefmark{3} and Han-Lim Choi\authorrefmark{4}
	\thanks{{{Soon-Seo~Park and Han-Lim~Choi are with the Department of Aerospace Engineering \& KI for Robotics, KAIST. 
				Young-Jin~Park is with NAVER R\&D Center, NAVER Corporation.
				Youngjae Min is with School of Electrical Engineering, KAIST.}}}    
	\thanks{{{\authorrefmark{4}Corresponding author: hanlimc@kaist.ac.kr}}}    
}

\begin{document}

	\maketitle
	\thispagestyle{empty}
	\pagestyle{empty}

\begin{abstract}
	This paper proposes an online learning method of Gaussian process state-space model (GP-SSM). GP-SSM is a probabilistic representation learning scheme that represents unknown state transition and/or measurement models as Gaussian processes (GPs). While the majority of prior literature on learning of GP-SSM are focused on processing a given set of time series data, data may arrive and accumulate sequentially over time in most dynamical systems. Storing all such sequential data and updating the model over entire data incur large amount of computational resources in space and time. To overcome this difficulty, we propose a practical method, termed \textit{onlineGPSSM}, that incorporates stochastic variational inference (VI) and online VI with novel formulation. The proposed method mitigates the computational complexity without catastrophic forgetting and also support adaptation to changes in a system and/or a real environments. Furthermore, we present application of onlineGPSSM into the reinforcement learning (RL) of partially observable dynamical systems by integrating onlineGPSSM with Bayesian filtering and trajectory optimization algorithms. Numerical examples are presented to demonstrate applicability of the proposed method.
\end{abstract}

	\maketitle
	
\section{Introduction}\label{sec:intro}
System identification is learning dynamical systems from measured data by representing unknown transition and/or measurement mappings as parameterized models \cite{ljung1998system}. The models obtained through system identification have enabled regenerating or predicting the measurements as well as controlling the system \cite{li2017identification,li2017neuro,eleftheriadis2017identification,ko2009gpkf,boedecker2014approximate,pan2014probabilistic,deisenroth2015PILCO,mcallister2017data,doerr2018PRSSM,sternberg2017identification,frigola2014variational}.
State-Space Model (SSM) is the most general representation of dynamical systems with a transition model and a measurement model. It has been successfully applied in a variety of applications such as robotics, ecology, finance, and neuroscience \cite{brown1998statistical,polydoros2017survey}.

For non-linear probabilistic SSM, Gaussian processes (GP) have been popularly employed as they can robustly learn dynamics models with a small number of data. Many GP-based algorithms assume that the states are fully observable and learn a regression model between the state and its changes over time. However, in real systems, the state cannot be measured directly but is inferred from a series of noisy measurements.
This partially observable and noisy dynamical system can be represented by Gaussian process state-space model (GP-SSM) by modeling unknown transition and/or measurement models as GPs
\cite{doerr2018PRSSM,frigola2013bayesian, frigola2014identification, svensson2017flexible,  frigola2014variational,sternberg2017identification,eleftheriadis2017identification}.
GP-SSM is non-parametric, so it is effective in learning from a small number of time series data that may be partially observable.
By using GPs for latent transitions and/or measurement models, we can get away with an approximate model and learn a distribution over functions. This formulation allows us to account for model uncertainty.
Consequently, the system model will not become overconfident in regions of the space where data are scarce.

Learning an unknown dynamical system accompanies several essential difficulties in practice.
First, the data for the system identification is obtained sequentially and accumulates over time in most dynamical systems, because the data is often collected through interacting with real environments.
Secondly, the system may subject to resource constraints in memory and computation time.
Storing all the data for learning the system requires a large memory, and using it all for model learning can be computationally infeasible.
Lastly, system models may change gradually along time due to unknown factors such as abrasion, temperature, and humidity, and the systems can also be locally affected by disturbances such as wind.
Such unknown factors and disturbances cause a difference between the expected system output and the actual output. These can degrade the ability of the system and, in the worst cases, lead to failure of the desired goals.
Mitigating this effect requires an algorithm that can adapt to changes in the system or environment without catastrophic forgetting.
This paper presents the online inference method, onlineGPSSM, to learn the GP-SSM
with tackling the challenges above. Its online learning approach effectively deals with the sequentially arriving data without both stacking it in the memory and consuming a large amount of time to use entire data in learning.
The method enables the online learning by applying online VI \cite{ghahramani2000online,sato2001online,broderick2013streaming} upon a tractable variational approximation of true posterior based on stochastic VI \cite{salimbeni2017doubly,doerr2018PRSSM}.
This method constrains the drastic change of the hyperparameters and, thus, prevents catastrophic forgetting without revisiting the previous measurements.
{To the best of our knowledge, it is the first time to apply online VI to the online learning of recurrent state-space models while it has been successfully utilized in static settings such as regression, classification, and deep generative models \cite{nguyen2017variational,bui2017streaming}.}
We apply our method to the reinforcement learning of partially observable dynamic systems.
The onlineGPSSM is integrated with trajectory optimization and Bayesian filtering algorithms, which can generate a robust control policy for control/planning.
The validity of the proposed method is demonstrated with
numerical examples on a synthetic toy and on a fixed-wing unmanned aerial vehicle in a wind field.

\section{Related work}

Online learning is a common approach when it is infeasible or inappropriate to learn over the entire data. Nonetheless, a simple solution to newly learn over only recent data causes catastrophic forgetting\cite{sternberg2017identification}.
Streaming Variational Bayes (streaming-VB) \cite{broderick2013streaming} is known as the most general method to online learning, and its application to GP-SSM is conceptually presented in \cite{frigola2014variational}. However, such a vanilla application of streaming-VB prevents online updating of hyperparameters in model learning and, consequently, causes catastrophic forgetting.
{Berntorp \cite{berntorp2019recursive} developed a recursive method for online learning of GP-SSM by extending the reduced-rank formulation of GP-SSM to the online setting and employing particle Markov chain Monte-Carlo (PMCMC) \cite{svensson2017flexible,frigola2013bayesian} for learning. Although such PMCMC- or Expectation Maximization (EM)\cite{frigola2014identification}-based learning methods find relatively accurate solutions, they are practically limited due to their extreme computational burden. Thus, the practicality calls for approximations and uses of inducing points through variational inference framework \cite{frigola2014variational, eleftheriadis2017identification, doerr2018PRSSM,salimbeni2017doubly}, as in our method.}

Recently, stochastic optimization based on the reparameterization trick\cite{kingma2013auto} has been successfully applied to model inference and learning methods.
{Towards this approach, Eleftheriadis et al. \cite{eleftheriadis2017identification} suggest an inference method to approximate the posterior of the latent state with a Markov-structured Gaussian distribution parameterized by a recognition model (i.e., bi-RNN).}
Doerr et al. \cite{doerr2018PRSSM} propose an inference scheme by utilizing the stochastic variational inference for deep GPs presented in \cite{salimbeni2017doubly}.
They follow a sparse variational approximation and incorporate the true dynamics into the approximate posterior by combining sampling/gradient-based learning.
Specifically, they obtain the approximate posterior distribution (i.e., variational distribution) of latent states by sampling in the true dynamics.
Our framework also adopts the stochastic variational inference scheme.
However, rather than employing a recognition model to find an informative model initialization as in \cite{salimbeni2017doubly}, the proposed framework obtains an informative prior distribution of the initial state using Bayesian filtering and the results of online learning.


Gaussian process (GP) has been widely used for model-based reinforcement learning and control problems \cite{kober2013reinforcement,polydoros2017survey}.
For instance, the work by Deisenroth et al.\cite{deisenroth2015PILCO} introduces the GP-based policy search framework called PILCO.
Also, various methods have been proposed to avoid policy parameterization and reduce computation time\cite{pan2014probabilistic,boedecker2014approximate,mitrovic2010adaptive,morimoto2003minimax}, where GP is combined with the trajectory optimization method called Differential Dynamic Programming (DDP).
However, such GP-based algorithms assume that the states are fully observable, which
may not hold under significant noise \cite{mcallister2017data}.
To deal with noisy measurements, GP-SSM is combined with policy search\cite{mcallister2017data} or trajectory optimization method\cite{park2019bayesian}. However, they are focused on handling given time series data rather than sequentially arriving data. 
Our RL framework shares some similarities with Probabilistic Differential Dynamic Programming (PDDP)\cite{pan2014probabilistic}.
However, while PDDP learns a fully observable system model and controls the learned model through iterative Linear Quadratic Gaussian Regulator (iLQR),
our framework learn a system model from partially observable and/or noisy measurements and controls the model via iterative Linear Quadratic Gaussian in belief space (belief space iLQG)\cite{van2012BiLQG} .

\section{Problem Formulation}\label{subsec:alg1}
This paper solves a problem of learning a discrete-time partially observable dynamical system described as a state-space model (SSM):
\begin{equation}
\begin{split}
\boldsymbol{x}_{t+1} &= f(\boldsymbol{x}_{t},\boldsymbol{u}_{t})+\boldsymbol{w}_{f}, \\
\boldsymbol{y}_{t} &= g(\boldsymbol{x}_{t})+\boldsymbol{w}_{g},
\end{split} \label{eq: problem}
\end{equation}
where $\boldsymbol{x}_t \in \mathbb{R}^{D}$, $\boldsymbol{u}_t \in \mathbb{R}^{P}$, and $\boldsymbol{y}_t \in \mathbb{R}^{O}$ are latent states, control inputs, and measurements at time $t$, respectively.
$\boldsymbol{w}_{(\cdot)} \!\sim\! \mathcal{N}(\boldsymbol{0}, \boldsymbol{\Sigma}_{(\cdot)})$ are Gaussian system/measurement noises. The goal of the problem is learning
the unknown transition model $f:\mathbb{R}^{D+P}\to\mathbb{R}^{D}$ from sequentially arriving measurements along with the control inputs, while the measurement model $g:\mathbb{R}^{D}\to\mathbb{R}^{O}$ is known \cite{doerr2018PRSSM}.
Through GP-SSM, we represent the transition model $f$ as a GP, which is a probability distribution over functions\cite{frigola2015bayesian,frigola2014variational}.


{ The GP $f$ can be specified completely through its mean function $m:\mathbb{R}^{D+P}\to\mathbb{R}^{D}$ and covariance/kernel function $k:\mathbb{R}^{D+P}\times\mathbb{R}^{D+P}\to\mathbb{R}$ with a covariance matrix $\boldsymbol{V}$ \cite{GP,park2018deep}}. This work particularly employs matrix-variate GP to further consider the covariance among the multiple dimensions of the output.
Let $\tilde{\boldsymbol{x}}_{t} = [\boldsymbol{x}_{t}^T,\boldsymbol{u}_{t}^T]^T$ denote the input of the transition model at time $t$. Then, for a series of input data $\boldsymbol{X}=\tilde{\boldsymbol{x}}_{1:T}$, any finite subset of their function evaluations $\boldsymbol{f}_{1:T}=[f(\tilde{\boldsymbol{x}}_{t})]_{t=1}^{T}$ are jointly Gaussian:
\begin{align}
\begin{split}
p(\boldsymbol{f}_{1:T} \mid \boldsymbol{X})
&= \mathcal{MN}(\boldsymbol{f}_{1:T} ; \boldsymbol{m}_{\boldsymbol{X}},\boldsymbol{K}_{\boldsymbol{X}\boldsymbol{X}}, \boldsymbol{V})\\
&= \mathcal{N}( \vectori({\boldsymbol{f}_{1:T}}) ; \vectori(\boldsymbol{m}_{\boldsymbol{X}}), \boldsymbol{V} \otimes \boldsymbol{K}_{\boldsymbol{X} \boldsymbol{X}}),
\end{split}
\end{align}
where $\boldsymbol{m}_{\boldsymbol{X}} = [m(\tilde{\boldsymbol{x}}_{t})]_{t=1}^{T}$ is the mean matrix, and $\boldsymbol{K}_{\boldsymbol{X} \boldsymbol{X}}$ and $\boldsymbol{V}$ are the covariance matrices such that $(\boldsymbol{K}_{\boldsymbol{X}\boldsymbol{X}})_{i,j}=k(\tilde{\boldsymbol{x}}_i,\tilde{\boldsymbol{x}}_j)$.
$\vectori(\cdot)$ and $\otimes$ denote the vectorization and the Kronecker product, respectively.
This work represent the output covariance matrix as the free form, which is more expressive than kernelized forms \cite{park2018deep}.

\begin{figure}[t]
	\centering
	\includegraphics[width=.5\columnwidth]{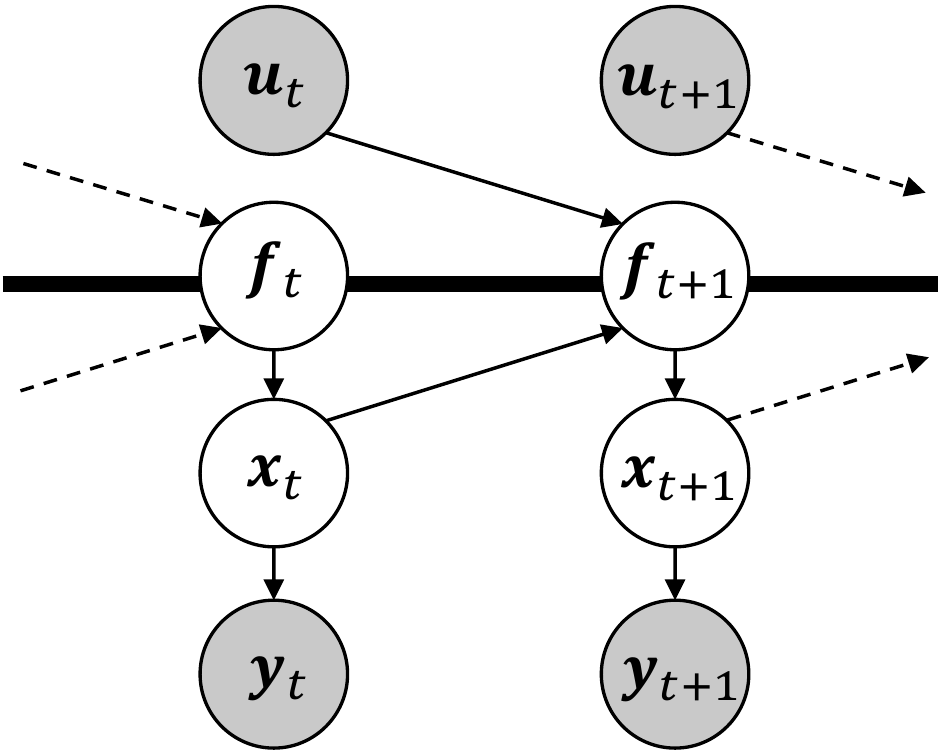}
	\caption{Graphical model of GP-SSM. White and gray nodes represent latent and observed variables, respectively. The thick lines indicate sets of fully connected variables, which are jointly Gaussian under the GP prior.}
	\label{fig: graphicalmodel}
\end{figure}

The main goal of learning $f$ is to be able to predict the function value $\boldsymbol{f}^*$, i.e., to assess the probability distribution of $\boldsymbol{f}^*$, at an arbitrary input $\tilde{\boldsymbol{x}}^{*}$. Given the input data $\boldsymbol{X}$ and their function evaluations $\boldsymbol{f}_{1:T}$, the GP predictive distribution at the new input $\tilde{\boldsymbol{x}}^{*}$ can be easily computed as:
\begin{equation}
\begin{split}
p(\boldsymbol{f}^{*} \mid \tilde{\boldsymbol{x}}^{*}, \boldsymbol{f}_{1:T}, \boldsymbol{X})
= \mathcal{N}(\boldsymbol{f}^{*} ; \boldsymbol{\mu}^*, \boldsymbol{\Sigma}^*),
\end{split}
\end{equation}
\vspace*{-.6cm}\\
with
\begin{equation}
\begin{split}
\boldsymbol{\mu}^* &= m(\tilde{\boldsymbol{x}}^{*})+\boldsymbol{k}_{\tilde{\boldsymbol{x}}^{*}\boldsymbol{X}} \boldsymbol{K}_{\boldsymbol{X} \boldsymbol{X}}^{-1}(\boldsymbol{f}_{1:T}-\boldsymbol{m}_{\boldsymbol{X}}),\\
\boldsymbol{\Sigma}^* &= (k(\tilde{\boldsymbol{x}}^{*},\tilde{\boldsymbol{x}}^{*}) - \boldsymbol{k}_{\tilde{\boldsymbol{x}}^{*}\boldsymbol{X}} \boldsymbol{K}_{\boldsymbol{X}\boldsymbol{X}}^{-1} \boldsymbol{k}_{\boldsymbol{X}\tilde{\boldsymbol{x}}^{*}})\boldsymbol{V},
\end{split}
\end{equation}
where $\boldsymbol{k}_{A,B}$ denotes the $|A|\times|B|$ matrix of the covariances evaluated at all pairs of the sets $A$ and $B$. ($|\tilde{\boldsymbol{x}}^{*}|=1$)
However, unlike general supervised learning problems, the training data $\boldsymbol{X}$ and $\boldsymbol{f}_{1:T}$ are latent in our problem. Thus, it is required to infer both the latent data and the GP $f$ (its mean and kernel functions) to evaluate the predictive distribution.

Due to the high computational complexity of the problem, the zero mean function and the squared exponential kernel function are generally adopted to specify a GP. When a rough model of the dynamics is available, it can be incorporated, instead, for the mean function to obtain meaningful distributions over the state trajectories from the first learning iteration. Also, note that any other differentiable kernel function can be employed for the gradient-based inference scheme (cf. Section \ref{sec:learning}).

With this conventional setting, we also adopt a sparse approximation scheme to alleviate the computational cost \cite{titsias2009variational}.
We approximate the GP predictive distribution with $M$ pseudo inducing points $\boldsymbol{\eta} = [\eta^{m}]_{m=1}^{M}$ and corresponding inducing outputs $\boldsymbol{z} = [f(\eta^{m})]_{m=1}^{M}$:
\begin{equation}
\begin{split}
p(\boldsymbol{f}^{*} \mid \tilde{\boldsymbol{x}}^{*}, \boldsymbol{f}_{1:T}, \boldsymbol{X})
\approx p(\boldsymbol{f}^{*} \mid \tilde{\boldsymbol{x}}^{*}, \boldsymbol{z}, \boldsymbol{\eta}).
\end{split}
\end{equation}
With respect to the inducing points, the full joint distribution of the GP-SSM can be represented by:
\begin{equation}
\begin{split}
p(\boldsymbol{y}_{0:T\!-\!1},\boldsymbol{x}_{0:T},\boldsymbol{f}_{1:T}, \boldsymbol{z})
= p(\boldsymbol{x}_{0:T}, \boldsymbol{f}_{1:T}, \boldsymbol{z})p(\boldsymbol{y}_{0:T\!-\!1} \mid \boldsymbol{x}_{0:T\!-\!1})\\
=p(\boldsymbol{x}_{1:T}, \boldsymbol{f}_{1:T} \mid \boldsymbol{z},\boldsymbol{x}_{0})p(\boldsymbol{z})p(\boldsymbol{x}_{0})p(\boldsymbol{y}_{0:T\!-\!1} \mid \boldsymbol{x}_{0:T\!-\!1}),
\end{split} \label{eq:GP-SSM}
\end{equation}
where
\begin{equation}
\begin{split}
&p(\boldsymbol{x}_{1:T},\boldsymbol{f}_{1:T} \mid \boldsymbol{z}, \boldsymbol{x}_{0}) = \prod_{t=1}^{T} p(\boldsymbol{x}_{t} \mid \boldsymbol{f}_{t})
p(\boldsymbol{f}_{t} \mid \boldsymbol{z}, \tilde{\boldsymbol{x}}_{t\!-\!1}),\\
&p(\boldsymbol{z}) = \mathcal{MN}(\boldsymbol{z} ; \boldsymbol{m}_{\boldsymbol{\eta}}, \boldsymbol{K}_{\boldsymbol{\eta} \boldsymbol{\eta}},\boldsymbol{V}),\\
&p(\boldsymbol{y}_{0:T\!-\!1} \mid \boldsymbol{x}_{0:T\!-\!1}) = \prod_{t=0}^{T\!-\!1} p(\boldsymbol{y}_{t} \mid \boldsymbol{x}_{t}).       
\end{split}
\end{equation}
{Based on these formulations, our goal is to estimate the pseudo inducing points, corresponding inducing outputs (or their statistics) and noise covariances as well as the parameters for the GP (its kernel function).}

\section{Stochastic Variational Inference in GP-SSM}\label{sec:learning}
{This section presents a conventional non-online inference method for learning the GP-SSM on which our algorithm is based.} Learning the GP-SSM is generally intractable due to the nonlinearity and non-Gaussianity of the dynamics model in a latent state. However, instead of directly optimizing the log marginal likelihood $\log p(\boldsymbol{y})$, its tractable Evidence Lower BOund (ELBO) can be incorporated.
In the following, we form an approximation of true posterior of GP-SSM based on the stochastic variational inference scheme\cite{salimbeni2017doubly,doerr2018PRSSM} and derive ELBO via Jensen's inequality.

Following \cite{salimbeni2017doubly,doerr2018PRSSM}, we form a variational approximation to the true posterior distribution $p(\boldsymbol{x}_{0:T},\boldsymbol{f}_{1:T}, \boldsymbol{z} \mid \boldsymbol{y}_{0:T\!-\!1})$ with reflecting the true transition model through the sparse GP approximation from (\ref{eq:GP-SSM}).
We approximate the posterior for $\boldsymbol{z}$ as $q(\boldsymbol{z}) = \mathcal{MN}(\boldsymbol{z} ; \boldsymbol{\mu}_{\boldsymbol{z}}, \boldsymbol{\Sigma}_{\boldsymbol{z}}, \boldsymbol{V})$ with parameters $\{\boldsymbol{\mu}_{\boldsymbol{z}}, \boldsymbol{\Sigma}_{\boldsymbol{z}}, \boldsymbol{V}\}$.
Then, the variational distribution is:
\begin{equation}
\begin{split}
&q(\boldsymbol{x}_{0:T},\boldsymbol{f}_{1:T}, \boldsymbol{z}) =
p(\boldsymbol{x}_{1:T},\boldsymbol{f}_{1:T} \mid \boldsymbol{z},\boldsymbol{x}_{0}) q(\boldsymbol{z}) q(\boldsymbol{x}_{0}),
\end{split}\label{eq:variationdis}
\end{equation}
with $q(\boldsymbol{x}_{0}) = \mathcal{N}(\boldsymbol{x}_{0} ; \boldsymbol{\mu}_{\boldsymbol{x}_{0}}, \boldsymbol{\Sigma}_{\boldsymbol{x}_{0}})$.
Note that the variational distribution does not impose independence between output dimension, unlike the mean-field variational approximation \cite{park2018deep} in existing approaches.
We learn these hyper-parameters $\vartheta = \{\boldsymbol{\mu}_{\boldsymbol{x}_{0}},\boldsymbol{\Sigma}_{\boldsymbol{x}_{0}}, \boldsymbol{\Sigma}_{g},
\boldsymbol{\Sigma}_{f},\boldsymbol{\mu}_{\boldsymbol{z}}, \boldsymbol{\Sigma}_{\boldsymbol{z}}, \theta \}$
through maximizing ELBO where $\theta$ denotes the hyperparameters for GP consisting of the parameters for the kernel model, {the covariance matrix $\boldsymbol{V}$, and the pseudo inducing points $\boldsymbol{\eta}$.}

\subsubsection{Evidence Lower Bound (ELBO) and Stochastic ELBO evaluation}
With the standard variational inference methodology, the evidence lower bound (ELBO) is given by: \\
\begin{equation}
\hspace{-0.25cm}
\begin{split}
&\log p(\boldsymbol{y}_{0:T\!-\!1} \mid \vartheta)\\
&\geq \mathbb{E}_{q(\boldsymbol{x}_{0:T},\boldsymbol{f}_{1:T}, \boldsymbol{z} \mid \vartheta)}
\bigg[\log \dfrac{p(\boldsymbol{y}_{0:T\!-\!1},\boldsymbol{x}_{0:T},\boldsymbol{f}_{1:T}, \boldsymbol{z} \mid \vartheta)}{q(\boldsymbol{x}_{0:T},\boldsymbol{f}_{1:T}, \boldsymbol{z} \mid \vartheta)}  \bigg] \\
&= \log p(\boldsymbol{y}_{0:T\!-\!1} \mid \vartheta)\\ 
&\quad -\mathcal{D}_{KL}(q(\boldsymbol{x}_{0:T}, \boldsymbol{f}_{1:T}, \boldsymbol{z}\mid \vartheta) \lVert
p(\boldsymbol{x}_{0:T}, \boldsymbol{f}_{1:T}, \boldsymbol{z} \mid \boldsymbol{y}_{0:T\!-\!1},\vartheta)). \label{eq:MLE}
\end{split}
\end{equation}
Then, we maximize the ELBO by finding the closest $q$ to the true posterior in terms of KL divergence.
From equations (\ref{eq:GP-SSM}) and (\ref{eq:variationdis}), the ELBO can be transformed into\footnote{For notational simplicity, we denote $p(\boldsymbol{y}_{t} \mid \boldsymbol{x}_{t}, \vartheta)$ as $p(\boldsymbol{y}_{t} \mid \boldsymbol{x}_{t})$.}:
\begin{equation}
\begin{split}
\mathcal{L}(\vartheta) &= \sum_{t=0}^{T\!-\!1} \mathbb{E}_{q(\boldsymbol{x}_{t})} \big[\log p(\boldsymbol{y}_{t} \mid \boldsymbol{x}_{t}) \big]\\
&\quad-\mathcal{D}_{KL}(q(\boldsymbol{x}_{0}) \lVert p(\boldsymbol{x}_{0}))
-\mathcal{D}_{KL}(q(\boldsymbol{z}) \lVert p(\boldsymbol{z} \mid \theta)),\label{eq:ELBO}			
\end{split}
\end{equation}
where $q(\boldsymbol{x}_{t})$ denotes the marginalized distribution of \eqref{eq:variationdis}.
The first term encourages the state trajectory to achieve high likelihood for the measured data. 
The second term prompts the state trajectory to start from a region of the state space with high prior.
{Here, we assume a given prior distribution of the initial latent state $\boldsymbol{x}_{0}$.
	As many system control/planning problems are specified with uncertain prior knowledge of the system's initial state given by the probability distribution $p(\boldsymbol{x}_{0})$, it can be effectively utilized to learn the system model.}
The last part of the ELBO regularizes the approximate posterior for $\boldsymbol{z}$ to remain close to the GP prior.
A full derivation is provided in Appendix A.
For a more detailed explanation, we refer readers to \cite{salimbeni2017doubly,doerr2018PRSSM} and
the references therein.


Learning $\vartheta$ is done through the gradient descent method.
The second and third terms of the ELBO can be analytically computed as they are KL-divergences between two Gaussian distributions.
However, the first term is not analytically tractable with the expectation for latent state distribution $q(\boldsymbol{x})$.
The Markovian structure of the latent states and the sparse GP approximation can be used to enable a differentiable sampling-based estimation of the expectation term \cite{doerr2018PRSSM}.

Samples $\boldsymbol{x}_{t} \sim q(\boldsymbol{x}_{t})$ are recursively drawn from an approximate GP posterior for time $t=1,...,T$. {From the sparse GP approximation, the predictive distribution for $\boldsymbol{f}_{t}$ is represented as Gaussian with the mean and covariance given by:}
\begin{equation}
\begin{split}
\boldsymbol{\mu}_p &= m_{\tilde{\boldsymbol{x}}_{t\!-\!1}} + \boldsymbol{k}_{\tilde{\boldsymbol{x}}_{t\!-\!1},\boldsymbol{\eta}} \boldsymbol{K}_{\boldsymbol{\eta},\boldsymbol{\eta}}^{-1} \big[\boldsymbol{\mu}_{\boldsymbol{z}}-\boldsymbol{m}_{\boldsymbol{\eta}}\big],\\
\boldsymbol{\Sigma}_p &= \Big[k_{\tilde{\boldsymbol{x}}_{t\!-\!1},\tilde{\boldsymbol{x}}_{t\!-\!1}}-
\alpha(\tilde{\boldsymbol{x}}_{t\!-\!1})	\big[\boldsymbol{K}_{\boldsymbol{\eta},\boldsymbol{\eta}}-\boldsymbol{\Sigma}_{\boldsymbol{z}} \big] \alpha(\tilde{\boldsymbol{x}}_{t\!-\!1})^{\top} \Big] \boldsymbol{V},
\end{split} \label{eq:varationalGP}
\end{equation}
where $\alpha(\tilde{\boldsymbol{x}}_{t\!-\!1}) = \boldsymbol{k}_{\tilde{\boldsymbol{x}}_{t\!-\!1},\boldsymbol{\eta}} \boldsymbol{K}_{\boldsymbol{\eta},\boldsymbol{\eta}}^{-1}$.
Note that the distribution of the current latent state $q(\boldsymbol{x}_{t})$ is independent of past time steps, given $q(\boldsymbol{x}_{T\!-\!1})$ and the explicit representation of GP inducing points.

By using the reparametrization trick\cite{kingma2013auto}, the operation of drawing samples can be differentiable for the hyper-parameters $\vartheta$. We draw $\boldsymbol{\varepsilon} \sim \mathcal{N}(0,\boldsymbol{I})$ and then compute the following equation:
{\begin{equation}
	\hat{\boldsymbol{x}}_{t+1} = \boldsymbol{\mu}_p(\tilde{\boldsymbol{x}}_{t})+ \boldsymbol{\varepsilon}L, \label{eq:repara}
	\end{equation} 
	where $LL^T={\boldsymbol{\Sigma}_p(\tilde{\boldsymbol{x}}_{t},\tilde{\boldsymbol{x}}_{t})+\boldsymbol{\Sigma}_{f}}$,} $\tilde{\boldsymbol{x}}_{t} = (\hat{\boldsymbol{x}}_{t}, \boldsymbol{u}_{t})$ and $\hat{\boldsymbol{x}}_{0} \sim q(\boldsymbol{x}_{0})$. With this trick, the gradient can be back-propagated along time. From (\ref{eq:repara}), an unbiased estimator of the first term in the ELBO is represented by:
\begin{equation}
\mathbb{E}_{q(\boldsymbol{x}_{t})} \big[\log p(\boldsymbol{y}_{t} \mid \boldsymbol{x}_{t}) \big] \approx
\frac{1}{N} \sum_{i=1}^{N} \log p(\boldsymbol{y}_{t} \mid \hat{\boldsymbol{x}}_{t}^{(i)}). \label{eq:stocastic}
\end{equation}



\section{Main Result: Online Learning of GP-SSM}\label{sec:online}

We then present our online inference method, onlineGPSSM, to efficiently handle sequentially arriving data\footnote{In this paper, we use the terms `data' and `measurements' interchangeably.} in GP-SSM.
The method follows a Bayesian approach but employs online VI to approximate the intractable posterior distribution.
In most dynamical systems, measurement arrives in a sequential manner.
Storing and utilizing all data is impractical in terms of memory and computation time.
Specifically, for the time $T$ and $M$ inducing points, the storage demand is $\mathcal{O}(TD)$, and the computational complexity is $\mathcal{O}(TM^{2})$ for each sample trajectory. They increase in proportion to the amount of data.
Thus, an essential goal of the online learning is to learn a system model that can be used for anytime prediction and control.

{With the limited resources and sequentially arriving data, we assume that only the new measurements $\boldsymbol{y}_{T:T'\!-\!1}$ and control inputs $\boldsymbol{u}_{T:T'\!-\!1}$ are directly accessible after the learning is done with the data until $T\!-\!1$. Instead, our method indirectly reflects the old measurements $\boldsymbol{y}_{0:T\!-\!1}$ and control inputs $\boldsymbol{u}_{0:T\!-\!1}$ in the new learning process by assessing the current posterior using the previous approximate posterior. For later use, let $T_0$ denote the previous checkpoint right before time $T$ as $T$ before $T'$ ($T_0<T<T'$). Note that our general notion allows unfixed time intervals with arbitrary sizes including a single-step size.
	
	In our sequential learning scheme, the pseudo inducing points evolve along the learning process as new regions of input space are gradually revealed. Let $\boldsymbol{z}_0, \boldsymbol{z}$, and $\boldsymbol{z'}$ denote the inducing outputs corresponding to the inducing points $\boldsymbol{\eta}_0, \boldsymbol{\eta}$, and $\boldsymbol{\eta'}$ at the learning process at $T_0, T$, and $T'$, respectively.
	For the initial case, we set $T_0=0$ with empty $\boldsymbol{\eta}_0$ and $\boldsymbol{z}_0$.
	Unlike \eqref{eq:variationdis}, our trick for the sequential learning is considering the joint posterior including the previous inducing outputs:
	\begin{equation}
	\begin{split}
	&p(\boldsymbol{x}_{0:T'}, \boldsymbol{f}_{1:T'}, \boldsymbol{z}, \boldsymbol{z'} \mid \boldsymbol{y}_{0:T'\!-\!1}, \vartheta')\\
	&= \dfrac{p(\boldsymbol{x}_{0:T'}, \boldsymbol{f}_{1:T'}, \boldsymbol{z}, \boldsymbol{z'} \mid \vartheta')}{p(\boldsymbol{y}_{0:T'\!-\!1} \mid \vartheta')}
	p(\boldsymbol{y}_{0:T'\!-\!1} \mid \boldsymbol{x}_{0:T'\!-\!1})\\
	&= \dfrac{p(\boldsymbol{x}_{0:T\!-\!1}, \boldsymbol{f}_{1:T}, \boldsymbol{z}, \boldsymbol{z'} \mid \vartheta')}{p(\boldsymbol{y}_{0:T'\!-\!1} \mid \vartheta')}
	p(\boldsymbol{x}_{T} \mid \boldsymbol{x}_{0:T\!-\!1}, \boldsymbol{f}_{1:T}, \boldsymbol{z}, \boldsymbol{z'}, \vartheta') \\
	&\quad \times p(\boldsymbol{x}_{T+1:T'}, \boldsymbol{f}_{T+1:T'} \mid \boldsymbol{x}_{0:T}, \boldsymbol{f}_{1:T}, \boldsymbol{z}, \boldsymbol{z'},\vartheta')\\
	&\quad \times p(\boldsymbol{y}_{0:T\!-\!1} \mid \boldsymbol{x}_{0:T\!-\!1}) p(\boldsymbol{y}_{T:T'\!-\!1} \mid \boldsymbol{x}_{T:T'\!-\!1}). 
	\end{split}\label{eq:qnew}
	\end{equation}
	
	Now, we can incorporate the previous approximate posterior in \eqref{eq:qnew}. As in (\ref{eq:MLE}), the variational distribution $q(\boldsymbol{x}_{0:T}, \boldsymbol{f}_{1:T}, \boldsymbol{z}_0, \boldsymbol{z} \mid \vartheta)$ approximates the true posterior at the previous step:
	\begin{equation}
	\begin{split}
	&q(\boldsymbol{x}_{0:T}, \boldsymbol{f}_{1:T}, \boldsymbol{z}_0, \boldsymbol{z} \mid \vartheta)
	\approx p(\boldsymbol{x}_{0:T}, \boldsymbol{f}_{1:T}, \boldsymbol{z}_0, \boldsymbol{z} \mid \boldsymbol{y}_{0:T\!-\!1}, \vartheta)\\
	&\quad \qquad\qquad= \dfrac{p(\boldsymbol{x}_{0:T}, \boldsymbol{f}_{1:T}, \boldsymbol{z}_0, \boldsymbol{z} \mid \vartheta)}{p(\boldsymbol{y}_{0:T\!-\!1} \mid \vartheta)} p(\boldsymbol{y}_{0:T\!-\!1} \mid \boldsymbol{x}_{0:T\!-\!1}). \end{split}\label{eq:qold}
	\end{equation}
	Then, we can approximate the term $p(\boldsymbol{y}_{0:T\!-\!1} \mid \boldsymbol{x}_{0:T\!-\!1})$ in \eqref{eq:qnew} by inverting (\ref{eq:qold}).
	In addition, we approximate $p(\boldsymbol{x}_{T} \mid \boldsymbol{x}_{0:T\!-\!1}, \boldsymbol{f}_{1:T}, \boldsymbol{z}, \boldsymbol{z'}, \vartheta')$ with the state estimation $\hat{p}(\boldsymbol{x}_{T})$ obtained by applying Bayesian filtering on the previously learned system model (see Section \ref{sec:planning}).
	Then, (\ref{eq:qnew}) is transformed into:
	\begin{equation}
	\begin{split}
	&\hat{p}(\boldsymbol{x}_{0:T'},\boldsymbol{f}_{1:T'}, \boldsymbol{z}, \boldsymbol{z'} \mid \boldsymbol{y}_{0:T'\!-\!1},\vartheta') \\
	&= \dfrac{p(\boldsymbol{y}_{0:T\!-\!1} \mid \vartheta)}{p(\boldsymbol{y}_{0:T'\!-\!1} \mid \vartheta')}
	p(\boldsymbol{x}_{0:T\!-\!1}, \boldsymbol{f}_{1:T},\boldsymbol{z},\boldsymbol{z'} \mid \vartheta') \hat{p}(\boldsymbol{x}_{T})\\
	&\quad \times p(\boldsymbol{x}_{T+1:T'}, \boldsymbol{f}_{T+1:T'} \mid \boldsymbol{x}_{0:T}, \boldsymbol{f}_{1:T}, \boldsymbol{z}, \boldsymbol{z'},\vartheta') \\
	&\quad \times p(\boldsymbol{y}_{T:T'\!-\!1} \mid \boldsymbol{x}_{T:T'\!-\!1})\frac{q(\boldsymbol{x}_{0:T},\boldsymbol{f}_{1:T},\boldsymbol{z}_0,\boldsymbol{z} \mid \vartheta)}{p(\boldsymbol{x}_{0:T},\boldsymbol{f}_{1:T},\boldsymbol{z}_0,\boldsymbol{z} \mid \vartheta)}.
	\end{split}\label{eq:exactpost}
	\end{equation}
}

Since directly optimizing the posterior requires intractable computation, we consider variational updates that employ the projection through KL minimization as in Section \ref{sec:learning}. This approach considers \eqref{eq:exactpost} in a tractable form using new hyperparameters
in a similar way to the online variational update in GP \cite{bui2017streaming}.
{We choose our variational distribution form such that the previous approximate posterior is given by:
	\begin{equation}
	\begin{split}
	&q(\boldsymbol{x}_{0:T},\boldsymbol{f}_{1:T}, \boldsymbol{z}_0, \boldsymbol{z} \!\mid\! \vartheta)=
	p(\boldsymbol{x}_{0:T_0\!-\!1},\boldsymbol{f}_{1:T_0}, \boldsymbol{z}_0 \!\mid\! \boldsymbol{z}, \vartheta)\\
	&\quad \times p(\boldsymbol{x}_{T_0\!+\!1:T},\boldsymbol{f}_{T_0\!+\!1:T} \!\mid\! \boldsymbol{x}_{0:T_0},\boldsymbol{f}_{1:T_0},\boldsymbol{z}_0,\boldsymbol{z}, \vartheta)
	q(\boldsymbol{z}) q(\boldsymbol{x}_{T_0}),
	\end{split}\label{eq:newvar}
	\end{equation}
	where $q(\boldsymbol{z})=\mathcal{MN}(\boldsymbol{z}; \boldsymbol{\mu}_{\boldsymbol{z}}, \boldsymbol{\Sigma}_{\boldsymbol{z}}, \boldsymbol{V})$
	and $q(\boldsymbol{x}_{T_0})= \mathcal{N}(\boldsymbol{x}_{T_0}; \boldsymbol{\mu}_{\boldsymbol{x}_{T_0}}, \boldsymbol{\Sigma}_{\boldsymbol{x}_{T_0}})$.
	Note that \eqref{eq:newvar} is reduced to \eqref{eq:variationdis} for the initial case with the empty $\boldsymbol{z}_0$ and $T_0=0$.
	Accordingly, the new posterior approximation is represented with new hyperparameters as:
	\begin{equation}
	\begin{split}
	&q(\boldsymbol{x}_{0:T'},\boldsymbol{f}_{1:T'},\boldsymbol{z},\boldsymbol{z'} \!\mid\! \vartheta') = 
	p(\boldsymbol{x}_{0:T\!-\!1},\boldsymbol{f}_{0:T} ,\boldsymbol{z} \!\mid\! \boldsymbol{z'}, \vartheta')\\
	&\quad \times p(\boldsymbol{x}_{T\!+\!1:T'},\boldsymbol{f}_{T\!+\!1:T'} \!\mid\! \boldsymbol{x}_{0:T}, \boldsymbol{f}_{1:T}, \boldsymbol{z}, \boldsymbol{z'},\vartheta')
	q(\boldsymbol{z'}) q(\boldsymbol{x}_{T}),
	\end{split}
	\end{equation}
	where $q(\boldsymbol{z'})=\mathcal{MN}(\boldsymbol{z'}; \boldsymbol{\mu}_{\boldsymbol{z'}}, \boldsymbol{\Sigma}_{\boldsymbol{z'}}, \boldsymbol{V'})$ and $q(\boldsymbol{x}_{T})= \mathcal{N}(\boldsymbol{x}_{T}; \boldsymbol{\mu}_{\boldsymbol{x}_{T}}, \boldsymbol{\Sigma}_{\boldsymbol{x}_{T}})$.}
Then, optimizing \eqref{eq:exactpost} can be transformed into minimizing a KL divergence through the variational inference:
\begin{align} 
\begin{split}
&\mathcal{D}_{KL}(q(\cdot\!\mid\!\vartheta') \lVert
\hat{p}(\boldsymbol{x}_{0:T'},\boldsymbol{f}_{1:T'}, \boldsymbol{z}, \boldsymbol{z'}\!\mid\!\boldsymbol{y}_{0:T'\!-\!1},\vartheta'))\\
&= \log \dfrac{p(\boldsymbol{y}_{0:T'\!-\!1}\!\mid\!\vartheta')}{p(\boldsymbol{y}_{0:T\!-\!1}\!\mid\!\vartheta)}
+\mathbb{E}_{q(\cdot \mid \vartheta')}
\left[\log \dfrac{\tilde{p}(\boldsymbol{x}_{T_0}) p(\boldsymbol{z} \mid \theta)}{q(\boldsymbol{x}_{T_0}) q(\boldsymbol{z})} \right]\\
&+\mathbb{E}_{q(\cdot \mid \vartheta')}
\left[\log \dfrac{q(\boldsymbol{x}_{T}) q(\boldsymbol{z'})}
{\hat{p}(\boldsymbol{x}_{T}) p(\boldsymbol{z'}\!\mid\!\theta') p(\boldsymbol{y}_{T:T'\!-\!1}\!\mid\!\boldsymbol{x}_{T:T'\!-\!1})}
\right],
\end{split} \label{eq:onlineKL}
\end{align}
where $q(\cdot\!\mid\!\vartheta')$ represents $q(\boldsymbol{x}_{0:T'},\boldsymbol{f}_{1:T'},\boldsymbol{z},\boldsymbol{z'} \!\mid\! \vartheta')$.

Since the KL divergence is non-negative, the expectation terms in \eqref{eq:onlineKL} form a negative approximate lower bound of the log marginal likelihood represented as
$\log p(\boldsymbol{y}_{0:T'\!-\!1} \!\mid\! \vartheta')/p(\boldsymbol{y}_{0:T\!-\!1} \!\mid\! \vartheta) \approx \log p(\boldsymbol{y}_{T:T'\!-\!1} \!\mid\! \boldsymbol{y}_{0:T\!-\!1})$.
The negative lower bound is derived similarly as the equations (\ref{eq:MLE}) and (\ref{eq:ELBO}) with considering our new variational distribution form.
{The term related to $\boldsymbol{x}_{T_0}$ can be neglected if $\tilde{p}(\boldsymbol{x}_{T_0})\approx q(\boldsymbol{x}_{T_0})$ is achieved in the previous learning step.} Assuming this, the negative evidence lower bound (negative ELBO) of the approximate online log marginal likelihood is represented by:	
\begin{align}
\begin{split}
\mathcal{NL}(\vartheta') &=
- \sum_{t=T}^{T'\!-\!1} \mathbb{E}_{q({\boldsymbol{x}_{t}})} \big[\log p(\boldsymbol{y}_{t} \mid \boldsymbol{x}_{t})\big]\\
&+ \mathcal{D}_{KL}(q(\boldsymbol{x}_{T}) \lVert \hat{p}(\boldsymbol{x}_{T}))
+ \mathcal{D}_{KL}(q(\boldsymbol{z'}) \lVert p(\boldsymbol{z'} \mid \theta'))\\		
&+ \mathcal{D}_{KL}(q'(\boldsymbol{z}) \lVert q(\boldsymbol{z}))
-  \mathcal{D}_{KL}(q'(\boldsymbol{z}) \lVert p(\boldsymbol{z} \mid \theta)),
\end{split} \label{eq:NELBO}
\end{align}
where
\begin{equation*}
\begin{split}
q'(\boldsymbol{z})
&= \int\int\int q(\boldsymbol{x}_{0:T'},\boldsymbol{f}_{1:T'},\boldsymbol{z},\boldsymbol{z'} \mid \vartheta')
d\boldsymbol{x}_{0:T'}d\boldsymbol{f}_{1:T'}d\boldsymbol{z'}\\
&= \int p(\boldsymbol{z} \mid \boldsymbol{z'},\theta')q(\boldsymbol{z'}) d\boldsymbol{z'},\\
&= \mathcal{MN}(\boldsymbol{z} ; \boldsymbol{m}_{\boldsymbol{z}\mid \boldsymbol{z'}},
\boldsymbol{S}_{\boldsymbol{z}\mid \boldsymbol{z'}}, \boldsymbol{V'}),\\
\boldsymbol{m}_{\boldsymbol{z}\mid \boldsymbol{z'}}
&= \boldsymbol{m}_{\boldsymbol{\eta}} +
\boldsymbol{K}_{\boldsymbol{\eta}\boldsymbol{\eta}'}
\boldsymbol{K}_{\boldsymbol{\eta}'\boldsymbol{\eta}'}^{-1}
\big[\boldsymbol{\mu}_{\boldsymbol{z'}}-\boldsymbol{m}_{\boldsymbol{\eta}'}\big],\\ \boldsymbol{S}_{\boldsymbol{z}\mid \boldsymbol{z'}}
&= \boldsymbol{K}_{\boldsymbol{\eta}\boldsymbol{\eta}} -\boldsymbol{K}_{\boldsymbol{\eta}\boldsymbol{\eta}'}
\boldsymbol{K}_{\boldsymbol{\eta}'\boldsymbol{\eta}'}^{-1} (\boldsymbol{I}+\boldsymbol{\Sigma}_{\boldsymbol{z'}}
\boldsymbol{K}_{\boldsymbol{\eta}'\boldsymbol{\eta}'}^{-1})
\boldsymbol{K}_{\boldsymbol{\eta}'\boldsymbol{\eta}}.
\end{split}
\end{equation*}
The first three terms form the variational bound considering only the new measurements data as in \eqref{eq:ELBO}. {The last two terms lead the approximate posterior to take into account the old likelihood through the approximate posterior and the prior.}
See Appendix B for details about the derivation of \eqref{eq:onlineKL} and \eqref{eq:NELBO}.

\begin{figure*}[t]
	\centering
	\includegraphics[width=1.9\columnwidth]{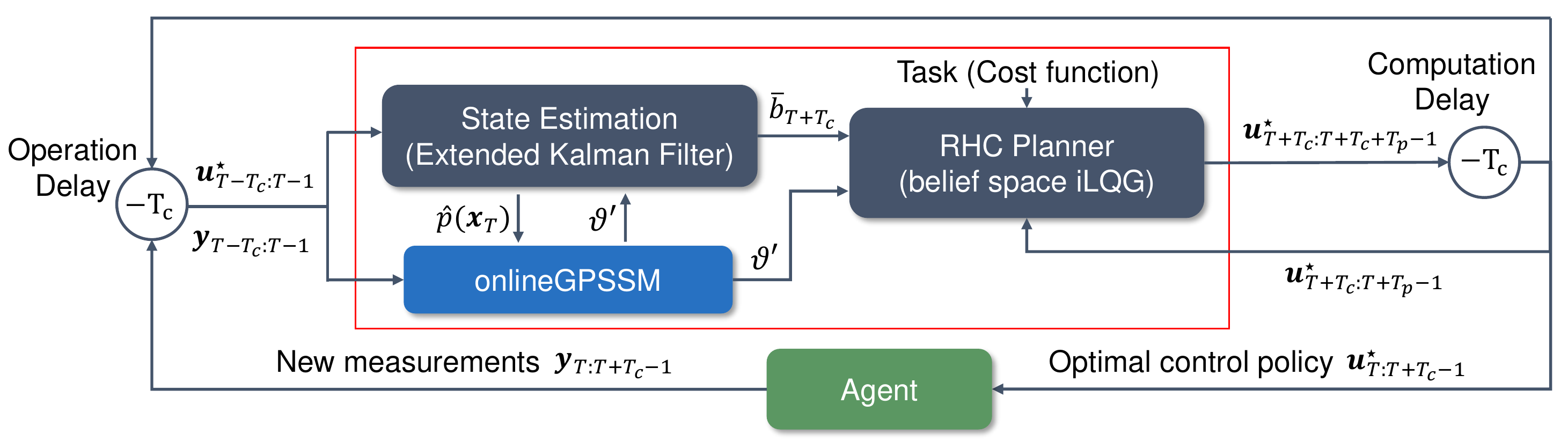}
	\caption{Proposed online learning and planning framework.}
	\label{fig:rlframework}
\end{figure*}

Note that the onlineGPSSM generalizes the vanilla application of streaming variational Bayes (streaming-VB) \cite{broderick2013streaming,frigola2014variational}.
For the GP-SSM setting with fixed hyperparameters (i.e., the positions of inducing inputs and kernel parameters), the previous posterior plays the role of the new prior for new measurements \cite{frigola2014variational}. Then, the onlineGPSSM becomes equivalent to the vanilla application of streaming-VB.
\section{Application: GP-SSM-based  Reinforcement  Learning Framework}\label{sec:planning}
In this section, we introduce our online learning and planning, i.e., reinforcement learning (RL), framework that employs a combination of probabilistic inference, Bayesian filtering, and trajectory optimization as shown in Fig. \ref{fig:rlframework}.

As in general optimal control problem, our goal of RL is to find the optimal control policy $\pi(\boldsymbol{x}_{t},t)$ that minimizes the expected cost:
\begin{equation}
\mathcal{J}^{\pi}(\boldsymbol{x}_{T})=\mathbb{E}_{\boldsymbol{x}}
\bigg[\phi(\boldsymbol{x}_{T+T_P})+ \sum_{t=T}^{T+T_P-1}  \ell(\boldsymbol{x}_{t},\pi(\boldsymbol{x}_{t},t)) \bigg], \label{eq:originalcost}
\end{equation}
where $\phi(\boldsymbol{x}_{T+T_P})$ and $\ell(\boldsymbol{x}_{t},\pi(\boldsymbol{x}_{t},t))$ are the final and instantaneous costs, respectively, and $\boldsymbol{u}_{t}=\pi(\boldsymbol{x}_{t},t)$  maps states and time to control inputs. The cost $\mathcal{J}^{\pi}(\boldsymbol{x}_{T})$ is defined as the expected total cost accumulated from $T$ to $T+T_P$. To compute \eqref{eq:originalcost}, it is essential to predict future states given the distribution of the current state $\boldsymbol{x}_{T}$.

\subsection{State Prediction and Estimation via Bayesian Filtering}\label{subsec:filtering}

The multi-step prediction can be done through Bayesian filtering.
Let $p(\boldsymbol{x}_{t},\boldsymbol{u}_{t}) = p(\tilde{\boldsymbol{x}}_{t}) \sim \mathcal{N}(\tilde{\boldsymbol{\mu}}_{t}, \tilde{\boldsymbol{P}}_{t})$ be the joint distribution over the state-control pair at $t$.
While the update of state distributions is formulated as $p(\tilde{\boldsymbol{x}}_{t\!+\!1}) \!=\! \int p(f(\tilde{\boldsymbol{x}}_{t}) \!\mid\! \tilde{\boldsymbol{x}}_{t})p(\boldsymbol{x}_{t})d\boldsymbol{x}_{t}$ with $\boldsymbol{u}_{t}\!=\!\pi(\boldsymbol{x}_{t},t)$, its computation is intractable due to the non-linearity of $p(f(\tilde{\boldsymbol{x}}_{t}) \!\mid\! \tilde{\boldsymbol{x}}_{t})$. {Thus, we approximate the update through linearization based on the Extended Kalman filter (EKF) as in \cite{ko2009gpkf}.}
Then, given $p(\boldsymbol{x}_{t}) \sim \mathcal{N}(\boldsymbol{x}_{t};\boldsymbol{\mu}_{t}, \boldsymbol{P}_{t})$, the state distribution at $t+1$ is represented as a Gaussian with mean and covariance:
\begin{equation}
\begin{split}
\boldsymbol{\mu}_{t+1} &= m_{\tilde{\boldsymbol{\mu}}_{t}} + \boldsymbol{k}_{\tilde{\boldsymbol{\mu}}_{t},\boldsymbol{\eta}} \boldsymbol{K}_{\boldsymbol{\eta},\boldsymbol{\eta}}^{-1} \big[\boldsymbol{\mu}_{\boldsymbol{z}}-\boldsymbol{m}_{\boldsymbol{\eta}}\big],\\
\boldsymbol{P}_{t+1} &= (\boldsymbol{I}-\boldsymbol{K}_{t}\boldsymbol{H}_{t}) \bar{\boldsymbol{P}}_{t+1} \label{eq:riccati},
\end{split}
\end{equation}
where
\begin{equation}
\begin{split}
\bar{\boldsymbol{P}}_{t+1} &= \boldsymbol{A}_{t} \boldsymbol{P}_{t} \boldsymbol{A}_{t}^{\top} + \boldsymbol{\Sigma}_{f}+\boldsymbol{\Sigma}_{p}',\\
\boldsymbol{K}_{t} &= \bar{\boldsymbol{P}}_{t+1} \boldsymbol{H}_{t}^{\top} (\boldsymbol{H}_{t} \bar{\boldsymbol{P}}_{t+1} \boldsymbol{H}_{t}^{\top} + \boldsymbol{\Sigma}_{g})^{-1},\\
\boldsymbol{H}_{t} &= \left. {\partial g(\tilde{\boldsymbol{x}}_{t})}/{\partial \tilde{\boldsymbol{x}}_{t}} \right\vert_{\tilde{\boldsymbol{\mu}}_{t}},
\boldsymbol{A}_{t} = {\partial \boldsymbol{\mu}_p(\tilde{\boldsymbol{x}}_{t})}/{\partial \tilde{\boldsymbol{x}}_{t}}\vert_{\tilde{\boldsymbol{\mu}}_{t}},\\
\tilde{\boldsymbol{\mu}}_{t} &= [\boldsymbol{\mu}_t,\pi(\boldsymbol{\mu}_{t},t)].
\end{split}
\end{equation}
$\boldsymbol{\Sigma}_{p}'$ is the latent dynamics uncertainty as in (\ref{eq:varationalGP}) evaluated at $\tilde{\boldsymbol{\mu}}_{t}$ .
$\boldsymbol{K}_{t}$ is often called Kalman gain\cite{chen2003bayesian}.

{
	While \eqref{eq:riccati} recursively predicts future state distributions without any measurements, it is also required to estimate the new current state distribution after applying the optimized control inputs for solving \eqref{eq:originalcost} in the next time frame. This estimation is also developed through EKF with a slight change from \eqref{eq:riccati}. Given the measurement $\boldsymbol{y}_{t}$, the estimation for the distribution of the state $\boldsymbol{x}_{t}$ is obtained with new mean
	\begin{equation}
	\boldsymbol{\mu}_{t} = m_{\tilde{\boldsymbol{\mu}}_{t\!-\!1}} + \boldsymbol{k}_{\tilde{\boldsymbol{\mu}}_{t\!-\!1},\boldsymbol{\eta}} \boldsymbol{K}_{\boldsymbol{\eta},\boldsymbol{\eta}}^{-1}
	\big[\boldsymbol{\mu}_{\boldsymbol{z}}-\boldsymbol{m}_{\boldsymbol{\eta}}\big] + \boldsymbol{K}_{t} \big[\boldsymbol{y}_{t} - g(\tilde{\boldsymbol{x}}_{t})\big] \label{eq:ekf}
	\end{equation}
	and the same covariance as \eqref{eq:riccati}. The estimated distribution is also utilized as $\hat{p}(\boldsymbol{x}_{T})$ in our online learning method (Section \ref{sec:online}).
	Note that any other nonlinear filters, such as particle filter and unscented Kalman filter, can be incorporated instead for the state estimation\cite{chen2003bayesian}. For the optimization technique in Section \ref{subsec:ilqg}, the future state prediction further requires differentiable belief dynamics.}

\subsection{Probabilistic Trajectory Optimization} \label{subsec:ilqg}

Based on the multi-step prediction, we now consider the control/planning problem \eqref{eq:originalcost} for the system model learned through onlineGPSSM.
In order to incorporate model uncertainty explicitly, we adopt belief space iterative Linear Quadratic Gaussian (belief space iLQG \cite{van2012BiLQG}) and perform trajectory optimization in the belief space.
The belief space iLQG finds a locally optimal trajectory for known belief dynamics and cost function using an iterative procedure.
It proceeds by, first, linearizing belief dynamics forward in time around the nominal trajectory and, then, computing a locally optimal control law backward in time. These steps are repeated until convergence to the locally optimal trajectory.

\subsubsection{Approximate Belief Dynamics and Cost Function}
The belief space iLQG assumes a given dynamics model and the stochasticity from the process and sensor noises, but we use the learned dynamics model and nominal trajectory distribution that is a state distribution-control pair $(p(\bar{\boldsymbol{x}}_{t}),\bar{\boldsymbol{u}}_{t})$.
To incorporate uncertainty explicitly into the local model, we define the belief state as $\boldsymbol{b}_{t}=[\boldsymbol{\mu}_{t}, \textmd{vec}({\boldsymbol{P}}_{t})]^{\top} \in \mathcal{R}^{D+D(D+1)/2}$ ($\boldsymbol{P}_{t}$ is symmetric).
Based on (\ref{eq:varationalGP}) and (\ref{eq:riccati}), the belief dynamics with the augmented state $\boldsymbol{b}_{t}$ is given by:
\begin{equation}
\boldsymbol{b}_{t+1} = \mathcal{F}(\boldsymbol{b}_{t},\boldsymbol{u}_{t}).
\end{equation}
Then, the original cost function (\ref{eq:originalcost}) is reformulated as:	
\begin{equation}
\mathcal{J}^{\pi}(\boldsymbol{b}_{T})= \phi(\boldsymbol{b}_{T+T_P})+ \sum\nolimits_{t=T}^{T+T_P-1}  \ell(\boldsymbol{b}_{t},\boldsymbol{u}_{t}), \label{eq:beliefcost}
\end{equation}	
where
\begin{equation}
\begin{split}
\phi(\boldsymbol{b}_{T+T_P})
&= \mathbb{E}[\phi(\boldsymbol{x}_{T+T_P})]\\
&= tr(Q_{T+T_P}\boldsymbol{P}_{T+T_P})\\
& \quad + (\boldsymbol{\mu}_{T+T_P}-\boldsymbol{x}^{+})^{\top} \tilde{Q}_{T+T_P} (\boldsymbol{\mu}_{T+T_P}-\boldsymbol{x}^{+}), \\
\ell(\boldsymbol{b}_{t},\boldsymbol{u}_{t}) &= \mathbb{E}[\ell(\boldsymbol{x}_{t},\pi(\boldsymbol{x}_{t}))]\\
&= tr(Q\boldsymbol{P}_{t}) + \boldsymbol{u}_{t}^{\top} R \boldsymbol{u}_{t} + \alpha(\boldsymbol{\mu}_{t}),
\end{split}
\end{equation}	
for given $Q_{t} \geq 0$ and $R_{t} > 0$. The term $\phi(\boldsymbol{b}_{T+T_P})$ encodes the final cost of arriving at the goal $\boldsymbol{x}^{+}$, $\boldsymbol{u}_{t}^{T} R_{t} \boldsymbol{u}_{t}$ penalizes the control effort along the trajectory,
$tr(Q_{t}\boldsymbol{P}_{t})$ penalizes the uncertainty, and $\alpha(\boldsymbol{\mu}_{t})$ encodes the state constraints (e.g., motion and collision).

\subsubsection{Control Policy}
By linearizing the belief dynamics around the nominal trajectory distribution, the approximate dynamics are expressed as:
\begin{equation}
\begin{split}
\boldsymbol{b}_{t+1} - \bar{\boldsymbol{b}}_{t+1} &\approx \mathcal{F}_{\boldsymbol{b},t}(\boldsymbol{b}_{t} - \bar{\boldsymbol{b}}_{t})
+ \mathcal{F}_{\boldsymbol{u},t}(\boldsymbol{u}_{t} - \bar{\boldsymbol{u}}_{t}),\\
W_{(i)}(\boldsymbol{b}_{t},\boldsymbol{u}_{t}) &\approx \boldsymbol{e}_t^i + \mathcal{F}_{\boldsymbol{b},t}^{(i)}(\boldsymbol{b}_{t} - \bar{\boldsymbol{b}}_{t})
+ \mathcal{F}_{\boldsymbol{u},t}^{(i)}(\boldsymbol{u}_{t} - \bar{\boldsymbol{u}}_{t}),
\end{split} \label{eq:linear_dyn}
\end{equation}
where
\begin{equation*}
\begin{split}
&\mathcal{F}_{\boldsymbol{b},t} = \frac{\partial \mathcal{F}}{\partial \boldsymbol{b}}(\bar{\boldsymbol{b}}_{t},\bar{\boldsymbol{u}}_{t}),
\enspace
\mathcal{F}_{\boldsymbol{u},t} = \frac{\partial \mathcal{F}}{\partial \boldsymbol{u}}(\bar{\boldsymbol{b}}_{t},\bar{\boldsymbol{u}}_{t}),\\
&\enspace
\boldsymbol{e}_t^i = W_{(i)}(\bar{\boldsymbol{b}}_{t},\bar{\boldsymbol{u}}_{t}),
\enspace
\mathcal{F}_{\boldsymbol{b},t}^{i} = \frac{\partial W_{(i)}}{\partial \boldsymbol{b}}(\bar{\boldsymbol{b}}_{t},\bar{\boldsymbol{u}}_{t}), \\
&\enspace
\mathcal{F}_{\boldsymbol{u},t}^{i} = \frac{\partial W_{(i)}}{\partial \boldsymbol{u}}(\bar{\boldsymbol{b}}_{t},\bar{\boldsymbol{u}}_{t}). 				
\end{split}
\end{equation*}
$W_{(i)}(\boldsymbol{b}_{t},\boldsymbol{u}_{t})$ is the $i$-th column of matrix $W(\boldsymbol{b}_{t},\boldsymbol{u}_{t})$. Note that $W_{(i)}(\boldsymbol{b}_{t},\boldsymbol{u}_{t})$ has $n$ columns, where $n$ is the dimension of the state.
For a general nonquadratic cost function, we approximate it as a quadratic function along the nominal belief and control trajectory	$(\bar{\boldsymbol{b}}, \bar{\boldsymbol{u}})$,
\begin{equation}
\begin{split}
\ell(\boldsymbol{b}_{t},\boldsymbol{u}_{t})
\approx&  \dfrac{1}{2}
\begin{bmatrix}
\delta\boldsymbol{b}_{t} \\ \delta\boldsymbol{u}_{t}
\end{bmatrix}^{\top}
\begin{bmatrix}
\ell_{\boldsymbol{b}\boldsymbol{b},t} & \ell_{\boldsymbol{b}\boldsymbol{u},t} \\ 	\ell_{\boldsymbol{u}\boldsymbol{b},t} & \ell_{\boldsymbol{u}\boldsymbol{u},t}
\end{bmatrix}
\begin{bmatrix}
\delta\boldsymbol{b}_{t} \\ \delta\boldsymbol{u}_{t}
\end{bmatrix} \\
&+
\begin{bmatrix}
\delta\boldsymbol{b}_{t} \\ \delta\boldsymbol{u}_{t}
\end{bmatrix}^{\top}
\begin{bmatrix}
\ell_{\boldsymbol{b},t} \\ \ell_{\boldsymbol{u},t}
\end{bmatrix}
+ \ell_{0,t},
\end{split} \label{eq:quadratic_cost}
\end{equation}	
where $\ell_{0,t} = \ell(\bar{\boldsymbol{b}}_{t},\bar{\boldsymbol{u}}_{t})$.
$\delta\boldsymbol{b}_{t} = \boldsymbol{b}_{t} - \bar{\boldsymbol{b}}_{t}$,  $\delta\boldsymbol{u}_{t} = \boldsymbol{u}_{t} - \bar{\boldsymbol{u}}_{t}$ are deviations from the nominal trajectory and the terms with subscripts denote Jacobian and Hessian matrices of their respective functions.

Given the linear dynamics \eqref{eq:linear_dyn} and quadratic cost \eqref{eq:quadratic_cost}, we can obtain a quadratic approximation of the value function along the nominal trajectory $\bar{\boldsymbol{b}}_t$:
\begin{equation}
\begin{split}
\boldsymbol{V}_{t}(\boldsymbol{b}_{t}) &\approx \dfrac{1}{2} \delta\boldsymbol{b}_{t}^{\top} \boldsymbol{V}_{\boldsymbol{b}\boldsymbol{b},t} \delta\boldsymbol{b}_{t}
+ \delta\boldsymbol{b}_{t}^{\top} \boldsymbol{V}_{\boldsymbol{b},t} + \boldsymbol{V}_{0,t}.\\
\end{split} \label{eq:ori_value_function}
\end{equation}	
Following the dynamic programming principle \cite{jacobson1970DDP}, the Bellman equation for the value function $\boldsymbol{V}_{t}(\boldsymbol{b}_{t})$ and control policy $\pi_{t}(\boldsymbol{b}_{t})$ in discrete-time are specified as:
\begin{equation}
\begin{split}
\boldsymbol{V}_{t}(\boldsymbol{b}_{t})
&= \min_{\boldsymbol{u}_t} \Big( \ell(\boldsymbol{b}_{t},\boldsymbol{u}_{t}) + \mathbb{E} \big[ \boldsymbol{V}_{t+1}(\mathcal{F}(\boldsymbol{b}_{t},\boldsymbol{u}_{t}) + \textbf{w}_t) \big] \Big) \\
&= \min_{\boldsymbol{u}_t} \Big( \ell(\boldsymbol{b}_{t},\boldsymbol{u}_{t}) +
\dfrac{1}{2} \delta\boldsymbol{b}_{t+1}^{\top} \boldsymbol{V}_{\boldsymbol{b}\boldsymbol{b},t+1} \delta\boldsymbol{b}_{t+1} \\
&\quad\quad\quad+ \delta\boldsymbol{b}_{t+1}^{\top} \boldsymbol{V}_{\boldsymbol{b},t+1} + \boldsymbol{V}_{0,t+1} \\
&\quad\quad\quad+\dfrac{1}{2} tr
\big[W(\boldsymbol{b}_{t},\boldsymbol{u}_{t})^{\top} \boldsymbol{V}_{\boldsymbol{b}\boldsymbol{b},t+1} W(\boldsymbol{b}_{t},\boldsymbol{u}_{t}) \big] \Big) \\
&= \min_{\boldsymbol{u}_t} \boldsymbol{Q}(\boldsymbol{b}_{t}, \boldsymbol{u}_t), \\
\pi_{t}(\boldsymbol{b}_{t})	&= \argmin_{\boldsymbol{u}_t} \Big( \ell(\boldsymbol{b}_{t},\boldsymbol{u}_{t}) + \mathbb{E} \big[ \boldsymbol{V}_{t+1}(\mathcal{F}(\boldsymbol{b}_{t},\boldsymbol{u}_{t}) + \textbf{w}_t) \big] \Big),
\end{split} \label{eq:value_function}
\end{equation}	
where
\begin{equation*}
tr \big[W(\boldsymbol{b}_{t},\boldsymbol{u}_{t})^{\top} \boldsymbol{V}_{\boldsymbol{b}\boldsymbol{b},t+1} W(\boldsymbol{b}_{t},\boldsymbol{u}_{t}) \big] = \sum_{i=1}^{m} W_{(i)}(\boldsymbol{b}_{t},\boldsymbol{u}_{t}).
\end{equation*}
By substituting equations (\ref{eq:linear_dyn}) and (\ref{eq:quadratic_cost}) into (\ref{eq:value_function}), the Q-function is given by:
\begin{equation}
\begin{split}				
&\boldsymbol{Q}_{t}(\bar{\boldsymbol{b}}_{t}+\delta\boldsymbol{b}_{t},\bar{\boldsymbol{u}}_{t}+\delta\textbf{u}_{t})\\
&\quad= \dfrac{1}{2}
\begin{bmatrix}
\delta\boldsymbol{b}_{t} \\ \delta\boldsymbol{u}_{t}
\end{bmatrix}^{\top}
\begin{bmatrix}
Q_{\boldsymbol{b}\boldsymbol{b},t} & Q_{\boldsymbol{b}\boldsymbol{u},t} \\ 	Q_{\boldsymbol{u}\boldsymbol{b},t} & Q_{\boldsymbol{u}\boldsymbol{u},t}
\end{bmatrix}
\begin{bmatrix}
\delta\boldsymbol{b}_{t} \\ \delta\boldsymbol{u}_{t}
\end{bmatrix}\\
&\quad\quad+
\begin{bmatrix}
\delta\boldsymbol{b}_{t} \\ \delta\boldsymbol{b}_{t}
\end{bmatrix}^{\top}
\begin{bmatrix}
Q_{\boldsymbol{b},t} \\ Q_{\boldsymbol{u},t}
\end{bmatrix}
+ Q_{0,t},						
\end{split} \label{eq:q_fuction}
\end{equation}		
where
\begin{equation} \small
\begin{split}
&\boldsymbol{Q}_{\boldsymbol{b}\boldsymbol{b},t} = \ell_{\boldsymbol{b}\boldsymbol{b},t} + \mathcal{F}_{\boldsymbol{b},t}^{\top} \boldsymbol{V}_{\boldsymbol{b}\boldsymbol{b},t+1}\mathcal{F}_{\boldsymbol{b},t}
+ \sum_{i=1}^{m} \mathcal{F}_{\boldsymbol{b},t}^{i{\top}} \boldsymbol{V}_{\boldsymbol{b}\boldsymbol{b},t+1} \mathcal{F}_{\boldsymbol{b},t}^{i},\\
&\boldsymbol{Q}_{\boldsymbol{b},t} = \ell_{\boldsymbol{b},t} + \mathcal{F}_{\boldsymbol{b},t}^{\top} \boldsymbol{V}_{\boldsymbol{b},t+1}
+ \sum_{i=1}^{m} \mathcal{F}_{\boldsymbol{b},t}^{i{\top}} \boldsymbol{V}_{\boldsymbol{b}\boldsymbol{b},t+1} \boldsymbol{e}_{t}^{i},\\
&\boldsymbol{Q}_{\boldsymbol{u}\boldsymbol{u},t} = \ell_{\boldsymbol{u}\boldsymbol{u},t} + \mathcal{F}_{\boldsymbol{u},t}^{\top} \boldsymbol{V}_{\boldsymbol{b}\boldsymbol{b},t+1}\mathcal{F}_{\boldsymbol{u},t}
+ \sum_{i=1}^{m} \mathcal{F}_{\boldsymbol{u},t}^{i{\top}} \boldsymbol{V}_{\boldsymbol{b}\boldsymbol{b},t+1} \mathcal{F}_{\boldsymbol{u},t}^{i},\\
&\boldsymbol{Q}_{\boldsymbol{u},t} = \ell_{\boldsymbol{u},t} + \mathcal{F}_{\boldsymbol{u},t}^{\top} \boldsymbol{V}_{\boldsymbol{b},t+1}
+ \sum_{i=1}^{m} \mathcal{F}_{\boldsymbol{u},t}^{i{\top}} \boldsymbol{V}_{\boldsymbol{b}\boldsymbol{b},t+1} \boldsymbol{e}_{t}^{i},\\
&\boldsymbol{Q}_{\boldsymbol{u}\boldsymbol{b},t} = \ell_{\boldsymbol{u}\boldsymbol{v},t} + \mathcal{F}_{\boldsymbol{u},t}^{\top} \boldsymbol{V}_{\boldsymbol{b}\boldsymbol{b},t+1}\mathcal{F}_{\boldsymbol{b},t}
+ \sum_{i=1}^{m} \mathcal{F}_{\boldsymbol{u},t}^{i{\top}} \boldsymbol{V}_{\boldsymbol{b}\boldsymbol{b},t+1} \mathcal{F}_{\boldsymbol{b},t}^{i},\\	
&\boldsymbol{Q}_{0,t} = \ell_{0,t} + \boldsymbol{V}_{0,t+1} + \sum_{i=1}^{n} \boldsymbol{e}_t^{i{\top}} \boldsymbol{V}_{\boldsymbol{b}\boldsymbol{b},t+1} \boldsymbol{e}_t^{i}.	
\end{split} \label{eq:q_fuction2}
\end{equation}	

In order to find the optimal control policy, we compute the local variations in control $\delta\boldsymbol{\hat{u}}$ that minimize the quadratic approximation of the $Q$-function in (\ref{eq:value_function}):
\begin{equation}
\begin{split}
\delta\boldsymbol{u}^{\star}_{t} &= \argmax_{\delta\boldsymbol{u}_{t}} \big[ \boldsymbol{Q}_{t}(\bar{\boldsymbol{b}}_{t}+\delta\boldsymbol{b}_{t},\bar{\boldsymbol{u}}_{t}+\delta\boldsymbol{u}_{t}) \big]\\
&= -\boldsymbol{Q}_{\boldsymbol{u}\boldsymbol{u},t}^{-1}\boldsymbol{Q}_{\boldsymbol{u},t} -\boldsymbol{Q}_{\boldsymbol{u}\boldsymbol{u},t}^{-1}\boldsymbol{Q}_{\boldsymbol{u}\boldsymbol{b},t}\delta\boldsymbol{b}_{t}.
\end{split}\label{eq:optimal_policy}
\end{equation}
The optimal control can be found as $\boldsymbol{u}^{\star}_{t} = \bar{\boldsymbol{u}}_{t} + \delta\boldsymbol{u}^{\star}_{t}$.
Substituting (\ref{eq:optimal_policy}) into (\ref{eq:value_function}) gives the value function $\boldsymbol{V}_{t}(\boldsymbol{b}_{t})$ as a function of only $\boldsymbol{b}_{t}$ in the form of (\ref{eq:ori_value_function}):
\begin{equation}
\begin{split}
\boldsymbol{V}_{\boldsymbol{b}\boldsymbol{b},t} &= \boldsymbol{Q}_{\boldsymbol{b}\boldsymbol{b},t}
-\boldsymbol{Q}_{\boldsymbol{u}\boldsymbol{b},t}^{\top}\boldsymbol{Q}_{\boldsymbol{u}\boldsymbol{u},t}^{-1} \boldsymbol{Q}_{\boldsymbol{u}\boldsymbol{b},t},\\
\boldsymbol{V}_{\boldsymbol{b},t} &= \boldsymbol{Q}_{\boldsymbol{b},t}
-\boldsymbol{Q}_{\boldsymbol{u}\boldsymbol{b},t}^{\top}\boldsymbol{Q}_{\boldsymbol{u}\boldsymbol{u},t}^{-1} \boldsymbol{Q}_{\boldsymbol{u},t},\\
\boldsymbol{V}_{0,t} &= \boldsymbol{Q}_{0,t}
-\boldsymbol{Q}_{\boldsymbol{u},t}^{\top}\boldsymbol{Q}_{\boldsymbol{u}\boldsymbol{u},t}^{-1} \boldsymbol{Q}_{\boldsymbol{u},t}.			
\end{split}
\end{equation}
This recursion continues by computing a control policy for time step $T+T_P-1$.

\begin{algorithm}[ht!] 
	\caption{Online Learning and Planning}\label{alg:ouralg}
	\begin{algorithmic}[1]
		\Statex \textbf{Input:} control horizon $T_c$, planning horizon $T_p$. 
		\Statex $\triangleright$ Initialization (line 1-11)
		\State Initialize $p(\boldsymbol{x}_{0}),\vartheta,\gamma$
		\State $T \leftarrow 0$, $\hat{p}(\boldsymbol{x}_{T}) \leftarrow p(\boldsymbol{x}_{0})$
		\State Generate arbitrary control inputs $\boldsymbol{u}^{\star}_{T:T+2T_{c}-1}$
		\State Execute $\boldsymbol{u}^{\star}_{T:T+T_{c}\!-\!1}$ \& obtain $\boldsymbol{y}_{T:T+T_c\!-\!1}$
		\State $T \leftarrow T+T_c$
		\State Execute $\boldsymbol{u}^{\star}_{T:T+T_{c}\!-\!1}$ \& obtain $\boldsymbol{y}_{T:T+T_c\!-\!1}$ (in parallel)
		\State $\vartheta \leftarrow \Call{Learn1}{\vartheta, \hat{p}(\boldsymbol{x}_{T\!-\!T_c}),\boldsymbol{y}_{T\!-\!T_c:T\!-\!1},\boldsymbol{u}^{\star}_{T\!-\!T_c:T\!-\!1},\gamma}$
		\State $\hat{p}(\boldsymbol{x}_{T}) \leftarrow \Call{Estimate}{\vartheta,\hat{p}(\boldsymbol{x}_{T-T_c}),\boldsymbol{y}_{T-T_c:T-1}}$
		\State $\bar{\boldsymbol{b}}_{T+T_c} \leftarrow \Call{Predict}{\vartheta,\hat{p}(\boldsymbol{x}_{T})}$
		\State $\boldsymbol{u}^{\star}_{T\!+\!T_c:T\!+\!T_c\!+\!T_{p}\!-\!1} \leftarrow \Call{Plan}{\vartheta,\bar{\boldsymbol{b}}_{T\!+\!T_c},\boldsymbol{0}}$
		\State $T \leftarrow T+T_c$
		\Statex $\triangleright$ Main Loop (line 12-19)
		\Repeat
		\State Execute $\boldsymbol{u}^{\star}_{T:T+T_{c}\!-\!1}$ \& obtain $\boldsymbol{y}_{T:T+T_c\!-\!1}$ (in parallel)		
		\State $\vartheta \leftarrow \Call{Learn2}{\vartheta, \hat{p}(\boldsymbol{x}_{T\!-\!T_c}),\boldsymbol{y}_{T\!-\!T_c:T\!-\!1},\boldsymbol{u}^{\star}_{T\!-\!T_c:T\!-\!1},\gamma}$
		\State $\hat{p}(\boldsymbol{x}_{T}) \leftarrow \Call{Estimate}{\vartheta,\hat{p}(\boldsymbol{x}_{T-T_c}),\boldsymbol{y}_{T-T_c:T-1}}$
		\State $\bar{\boldsymbol{b}}_{T+T_c} \leftarrow \Call{Predict}{\vartheta,{p}(\boldsymbol{x}_{T})}$
		\State $\boldsymbol{u}^{\star}_{T\!+\!T_c:T\!+\!T_c\!+\!T_{p}\!-\!1}\!\leftarrow\!
		\Call{Plan}{\vartheta\!,\bar{\boldsymbol{b}}_{T\!+\!T_c}\!,[\boldsymbol{u}^{\star}_{T\!+\!T_c:T\!+\!T_p\!-\!1}\!,\boldsymbol{0}]}$
		\State $T \leftarrow T+T_c$
		\Until{Operation ends}
		
		\Statex $\triangleright$ Model Learning in Section \ref{sec:learning}
		\Function{Learn1}{$\vartheta, \hat{p}(\boldsymbol{x}_{T}),\boldsymbol{y}_{T:T+T_c-1},\boldsymbol{u}_{T:T+T_c-1},\gamma$}
		\Repeat
		\State $\vartheta \leftarrow \vartheta + \gamma\nabla\mathcal{L}(\vartheta)$
		\Statex \hspace{2cm}given $\{\hat{p}(\boldsymbol{x}_{T}),\boldsymbol{y}_{T:T+T_c-1},\boldsymbol{u}_{T:T+T_c-1}\}$
		\Until{$\mathcal{L}(\vartheta)$ converges}
		\State\Return $\vartheta$
		\EndFunction
		
		\Statex $\triangleright$ Model Learning in Section \ref{sec:online}
		\Function{Learn2}{$\vartheta, \hat{p}(\boldsymbol{x}_{T}),\boldsymbol{y}_{T:T+T_c-1},\boldsymbol{u}_{T:T+T_c-1},\gamma$}
		\State $\vartheta' \leftarrow \vartheta$
		\Repeat
		\State $\vartheta' \leftarrow \vartheta' - \gamma\nabla\mathcal{NL}(\vartheta')$
		\Statex \hspace{2cm}given $\{\vartheta, \hat{p}(\boldsymbol{x}_{T}),\boldsymbol{y}_{T:T+T_c-1},\boldsymbol{u}_{T:T+T_c-1}\}$
		\Until{$\mathcal{NL}(\vartheta')$ converges}
		\State\Return $\vartheta'$
		\EndFunction
		
		\Statex $\triangleright$ Trajectory Optimization in Section \ref{subsec:ilqg}
		\Function{Plan}{$\vartheta,\bar{\boldsymbol{b}}_{T},\bar{\boldsymbol{u}}_{T:T+T_p-1}$}
		\Repeat
		\For{$t\leftarrow T$ to $(T+T_p-1)$}
		\State $\bar{\boldsymbol{b}}_{t+1} = \mathcal{F}(\bar{\boldsymbol{b}}_{t},\bar{\boldsymbol{u}}_{t})$
		\EndFor
		\For{$t\leftarrow (T+T_p-1)$ to $T$}
		\State $\delta\boldsymbol{u}^{\star}_{t} = \argmax_{\delta\boldsymbol{u}_{t}} \big[ \boldsymbol{Q}_{t}(\bar{\boldsymbol{b}}_{t}+\delta\boldsymbol{b}_{t},\bar{\boldsymbol{u}}_{t}+\delta\boldsymbol{u}_{t}) \big]$
		\State $\bar{\boldsymbol{u}}_{t} \leftarrow \bar{\boldsymbol{u}}_{t} + \delta\boldsymbol{u}^{\star}_{t}$
		\EndFor
		\Until{$\bar{\boldsymbol{u}}_{T:T+T_p-1}$ converges}
		\State\Return $\bar{\boldsymbol{u}}_{T:T+T_p-1}$
		\EndFunction
		
	\end{algorithmic}
\end{algorithm}

\subsection{Summary of Algorithm}\label{subsec:alg2}

{Algorithm \ref{alg:ouralg} describes the proposed online learning and planning framework summarized in Fig. \ref{fig:rlframework}.}
Here, the receding horizon control (RHC), also known as model predictive control (MPC), scheme is adopted for real-time operation \cite{kuwata2004three,park2019distributed}. After planning the future $T_{p}$ steps, the system executes only the first $T_{c}$ steps $(T_c \leq T_p)$. Then, it plans the next $T_{p}$-step controls succeeding the current $T_{c}$-step plan while performing the current plan. This scheme enables the agent to move continuously without stopping to wait for the learning and planning process. $T_{c}$ is determined by the available computational resources required for model learning and trajectory optimization.

In the initialization part, the system first follows the arbitrarily generated $2T_c$-step controls to fit in with the RHC scheme.
When executing the last $T_c$ steps, it learns GP-SSM hyperparameters from the measurements obtained in the first $T_c$ steps and the given prior $\hat{p}(\boldsymbol{x}_{0})$ of the system (Section \ref{sec:learning}).
If an approximate model of the system is known, employing it as a mean prior of the GP-SSM and planning the control inputs from the beginning can accelerate the learning process.
We then perform trajectory optimization with the learned model to obtain the next optimal control inputs. 

{In the main loop, the system repeats the process of executing the current plan and obtaining new measurements while learning the model and planning the next plan.
	While executing the current plan, the system learns the GP-SSM from the recent measurements $\boldsymbol{y}_{T\!-\!T_c:T\!-\!1}$ and previous estimation  $\hat{p}(\boldsymbol{x}_{T-T_c})$ (Section \ref{sec:online}).
	Then, it utilizes $\boldsymbol{y}_{T\!-\!T_c:T\!-\!1}$ to estimate the current state distribution $\hat{p}(\boldsymbol{x}_{T})$ recursively from $\hat{p}(\boldsymbol{x}_{T-T_c})$ as \eqref{eq:ekf}. From $\hat{p}(\boldsymbol{x}_{T})$, the nominal belief $\bar{\boldsymbol{b}}_{T+T_c}$ for $p(\boldsymbol{x}_{T+T_c})$ is then predicted recursively without any measurements as \eqref{eq:riccati} (Section \ref{subsec:filtering}).
	This belief is utilized as an initial condition of the trajectory optimization to plan the next $T_p$-step controls (Section \ref{subsec:ilqg}). Note that the previous solution is utilized for warm start so that the $T_p$-step initial guess is $[\boldsymbol{u}^{\star}_{T+T_{c}},...,\boldsymbol{u}^{\star}_{T+T_{p}-1},0,...,0]$.
	This warm start enables much faster convergences than the zero-initial case in the initialization part unless the task too differs from the previous one.
	
	In addition to the current method, we can further consider the mini-batch training under the limited memory as in the literature on online/continual learning \cite{nguyen2017variational,bui2017streaming,awasthi2019continual,hoang2019collective}. This technique further helps to prevent the catastrophic forgetting in addition to our online learning formulation.
	When the limited memory allows storing $(\mathcal{M}\!=\!mT_c)$-step data, the training can be done with the partitioned mini-batches 
	$\langle \hat{p}(\boldsymbol{x}_{T\!-nT_c}), \boldsymbol{y}_{T\!-nT_c:T\!-(n\!-1)T_c\!-1},\boldsymbol{u}_{T\!-nT_c:T\!-(n\!-1)T_c\!-1} \rangle_{n = 1}^m$.
	Note that, if the number of mini-batches $m$ is set to $1$, the learning method is reduced to the recursive formulation as in Section \ref{sec:online} and \cite{berntorp2019recursive}.
}


\section{Numerical Simulations}\label{sec:simul}
In the following, we evaluate the onlineGPSSM and the GP-SSM-based RL framework using two examples: a toy and the loitering task of a fixed-wing Unmanned Aerial Vehicle (UAV) in a wind field.
In both experiments, the squared exponential kernel with ARD length-scales is used for GP, while any other differentiable kernel can be incorporated instead.
For the sparse approximation of GP for the latent transition model, we use 20 pseudo inducing points.
The onlineGPSSM, however, allows the varying number of pseudo inducing points at each online learning step.

\subsection{A simple linear system}\label{subsec:toy}

We start by analyzing the online learning approach with a simple linear system:
\begin{equation}
\begin{split}
x_{t+1} &= 0.8x_{t} + u_{t} + w_{f}\\
y_{t} &= x_{t} + w_{g}
\end{split}\label{eq:toy_real}
\end{equation}
with parameters $\Sigma_{f} = 0.1$, $\Sigma_{g} = 1$, and known control inputs $u_{t}$, which was set to -2 for the first 30 seconds, 3 for the next 30 seconds, and then to -1 for the last 40 seconds.
The online learning was performed five times with successive 20-second intervals.
The state at $t=0$ is given.

\begin{figure}[t]
	\centering
	\subfigure{
		\includegraphics[width=.95\columnwidth]{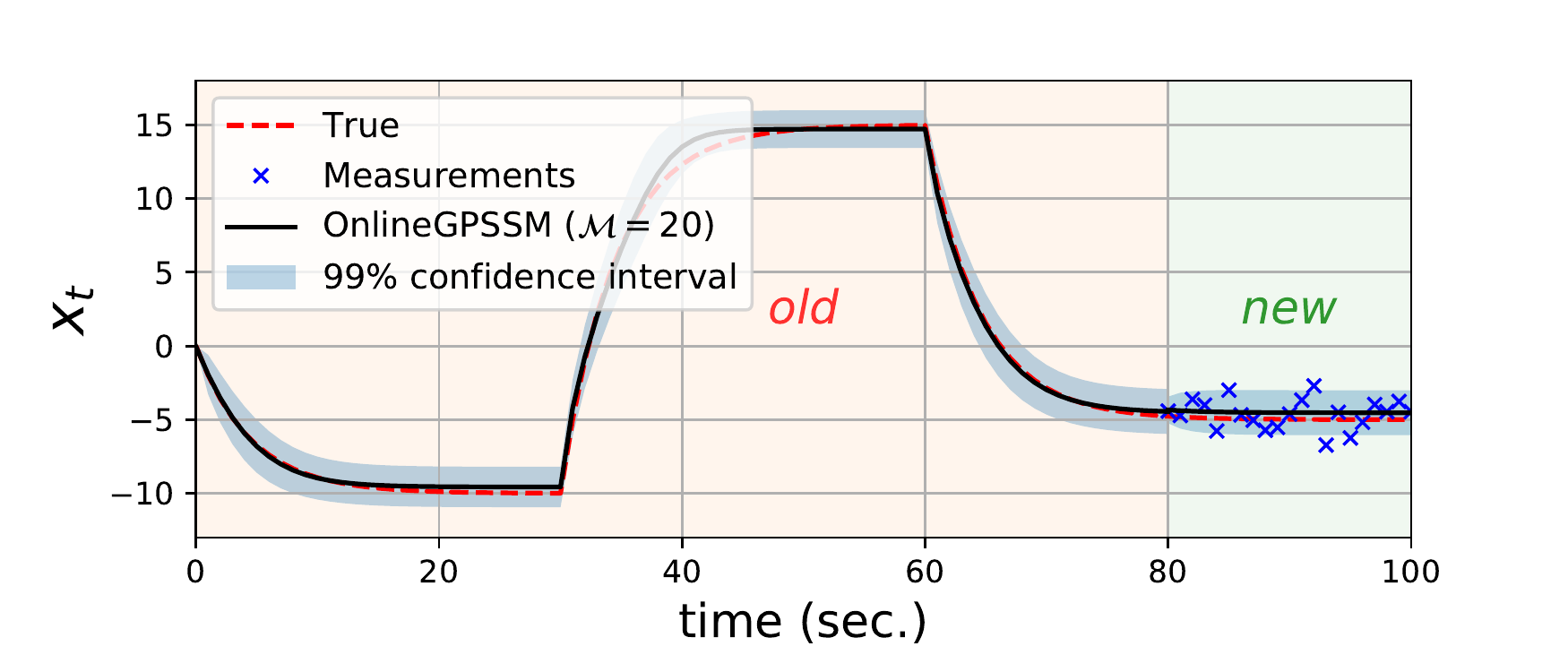}}\\
	\subfigure{
		\includegraphics[width=.95\columnwidth]{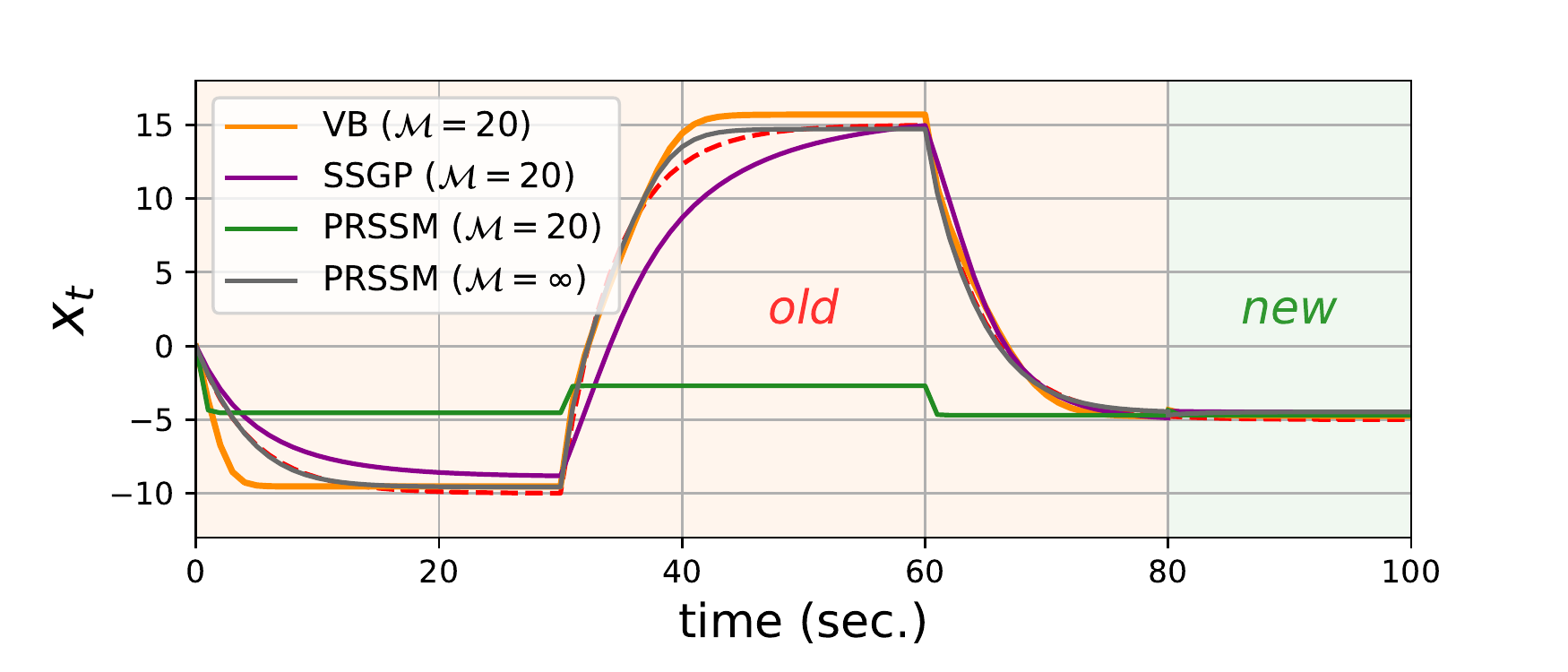}}\\\vspace*{-.3cm}
	\caption{Reconstruction results after 5-step online learning.\vspace*{-.4cm}}
	\label{fig:ex1_recon}
\end{figure}

We verify the performance of onlineGPSSM by comparing it to other online learning methods: a GP-based learning without considering noise \cite{bui2017streaming}, a vanilla application of streaming-VB \cite{frigola2015bayesian}, and learning through stochastic VI with finite and infinite memories \cite{doerr2018PRSSM}. After each learning step, the state trajectory from the initial state $\boldsymbol{x}_0$ is reconstructed through mean propagation using the learned system model and the known control inputs. Table. \ref{tbl:RMSE_comp} shows the RMSE for old, new, and total data at each step, and the final reconstruction results are shown in Fig. \ref{fig:ex1_recon}.
The result indicates that the proposed method outperforms most methods and shows similar performance with learning the entire data with unlimited memory.

\begin{figure}[t!]
	\centering
	\subfigure[Step 2]{
		\includegraphics[trim={0.1cm 0.cm 1.5cm 1.1cm},clip,width=0.45\columnwidth]{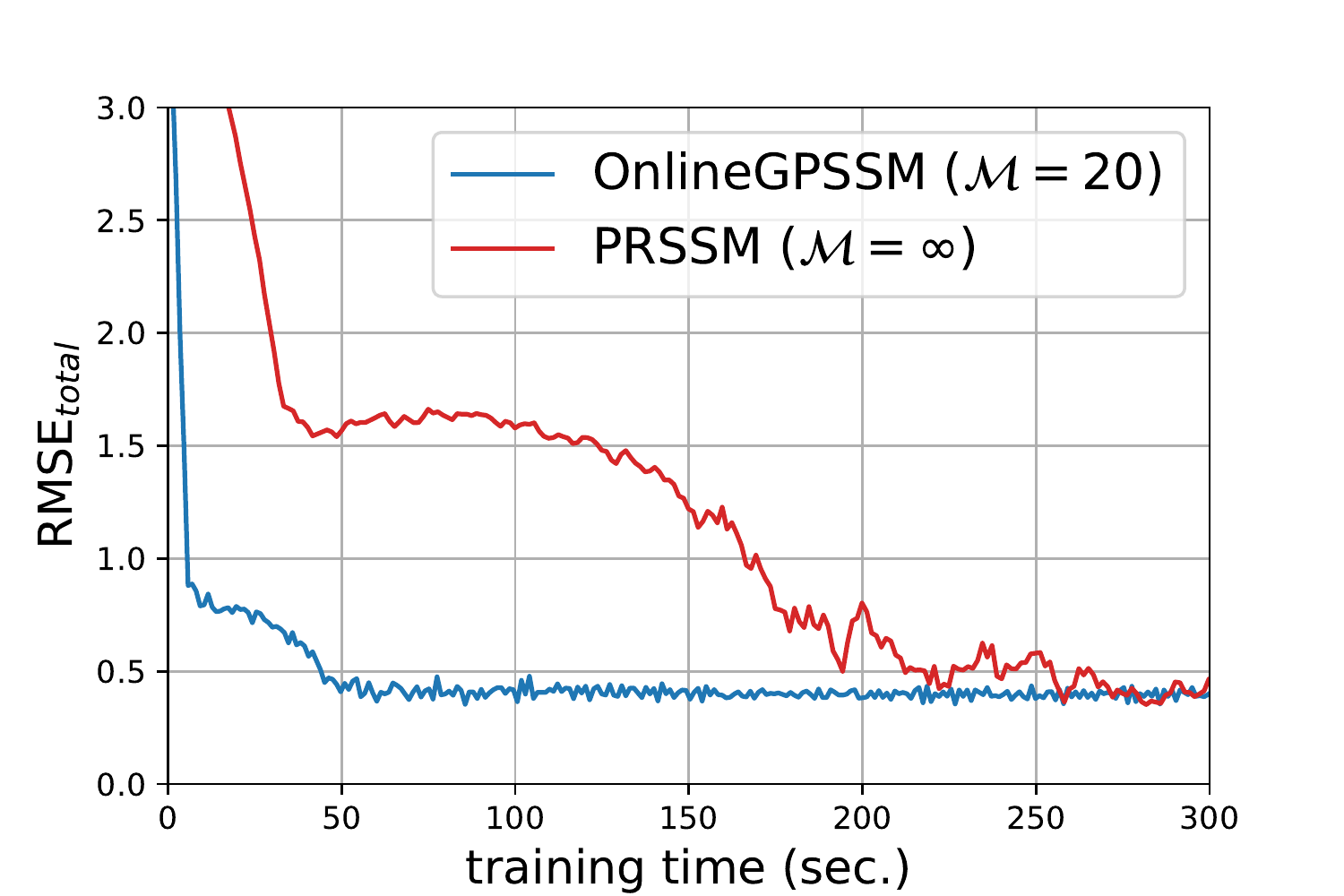}}
	\hspace{-0.2cm}
	\subfigure[Step 3]{
		\includegraphics[trim={0.1cm 0.cm 1.5cm 1.1cm},clip,width=0.45\columnwidth]{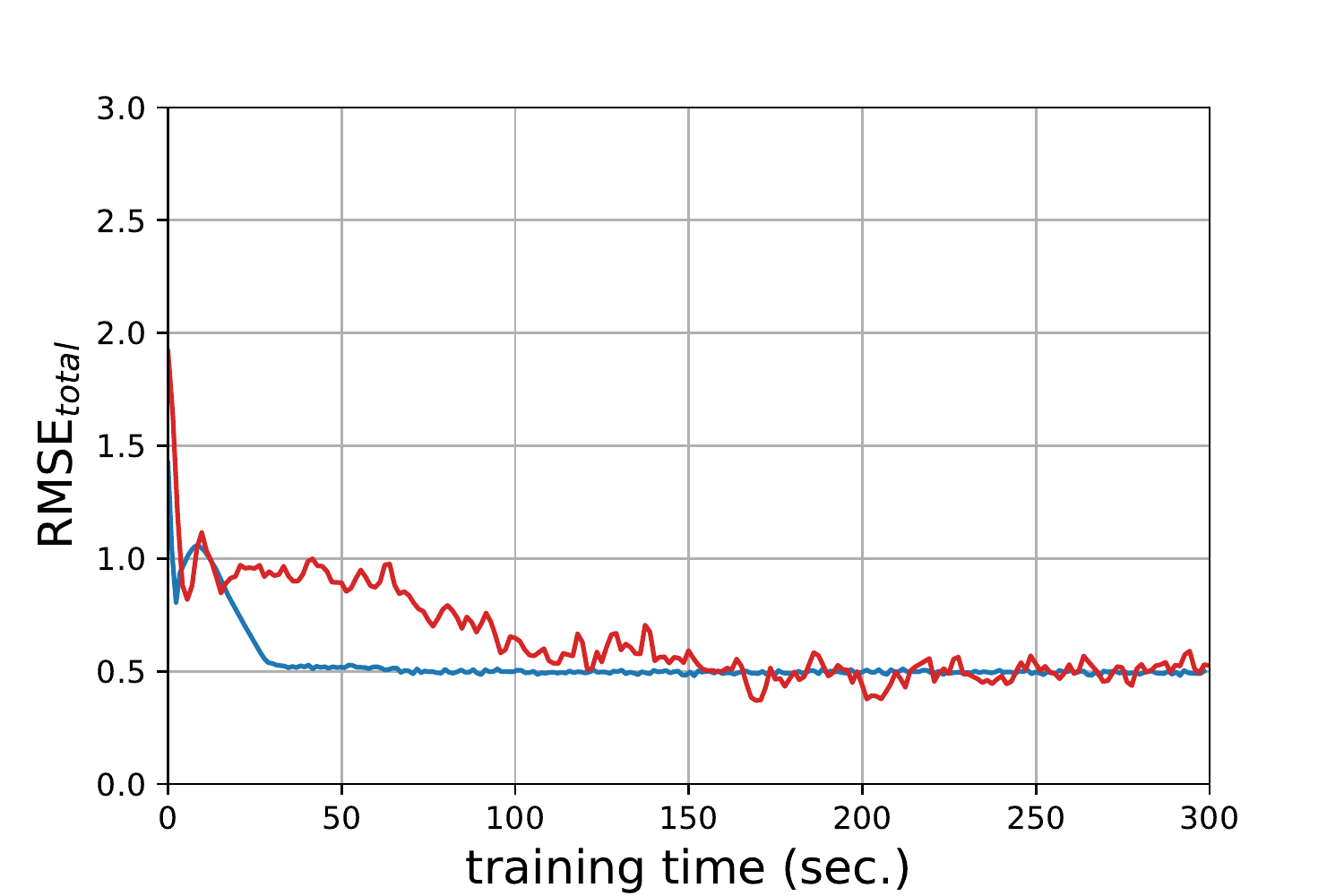}}\\
	\subfigure[Step 4]{
		\includegraphics[trim={0.1cm 0.cm 1.5cm 1.1cm},clip,width=0.45\columnwidth]{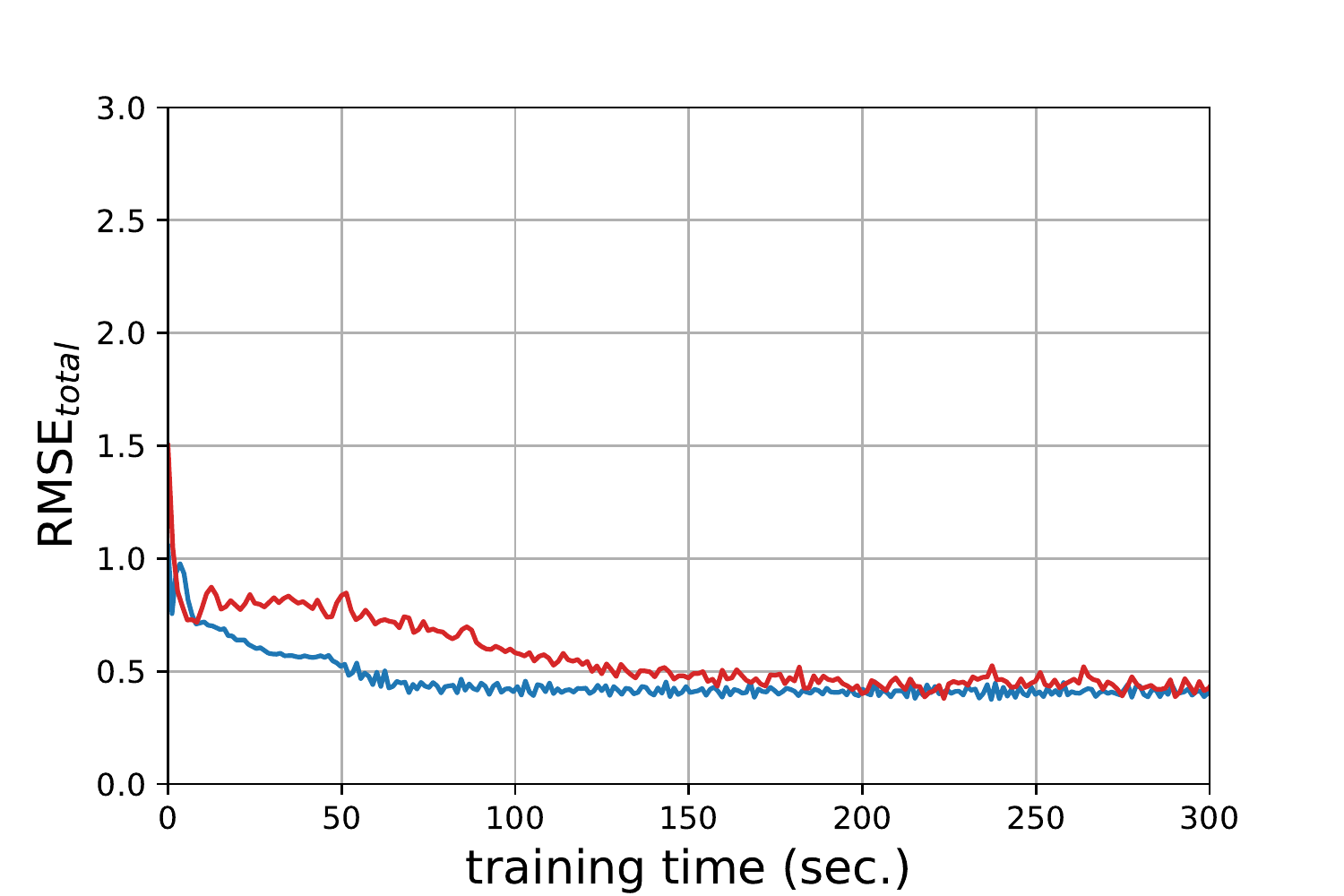}}
	\hspace{-0.2cm}
	\subfigure[Step 5]{
		\includegraphics[trim={0.1cm 0.cm 1.5cm 1.1cm},clip,width=0.45\columnwidth]{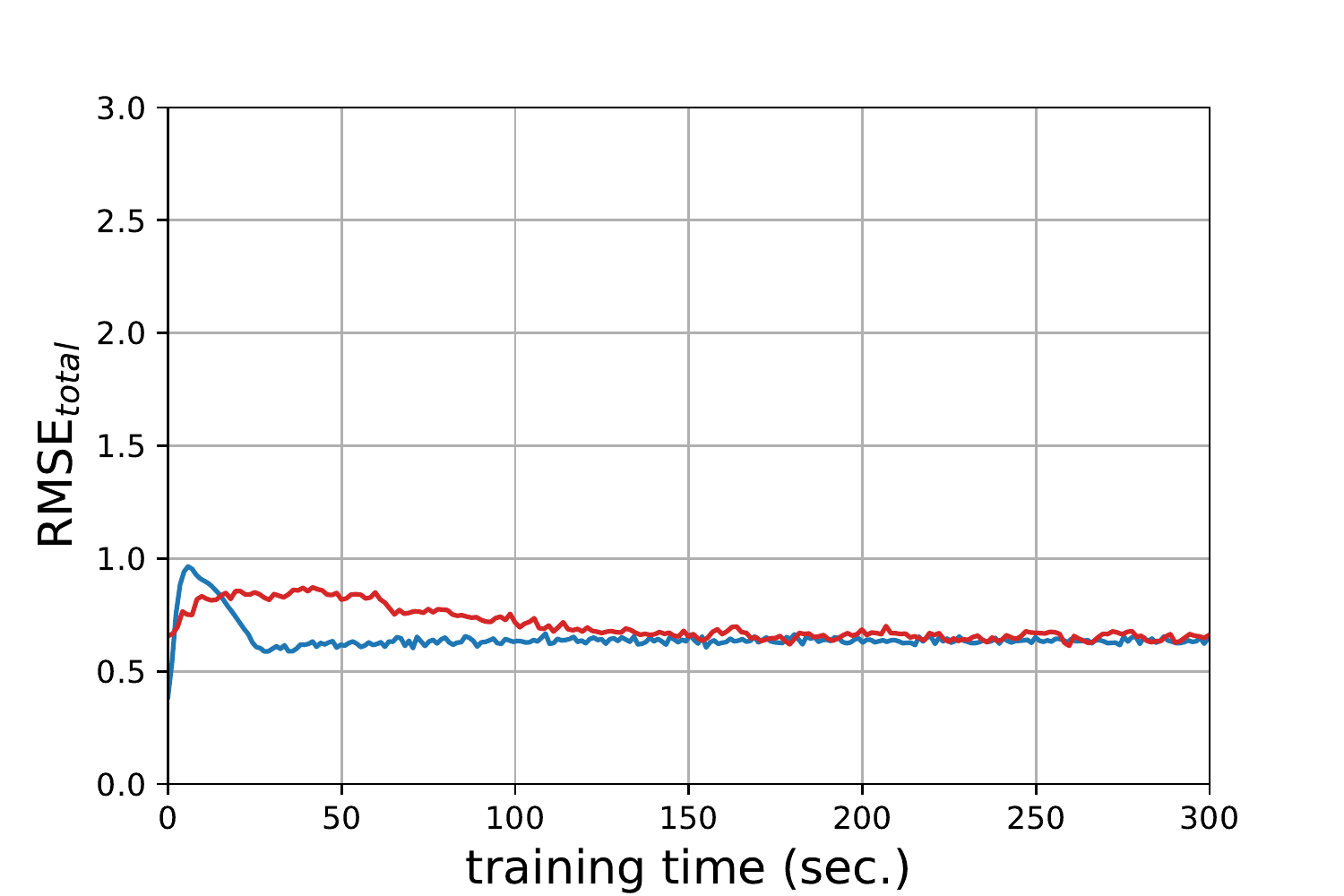}}
	\caption{Comparison of RMSE w.r.t. training time.}				
	\label{fig:ex1_time}
\end{figure}

\begin{figure}[ht!]
	\centering
	\subfigure[Step 2]{
		\includegraphics[trim={0.1cm 0.cm 1.5cm 1.1cm},clip,width=0.45\columnwidth]{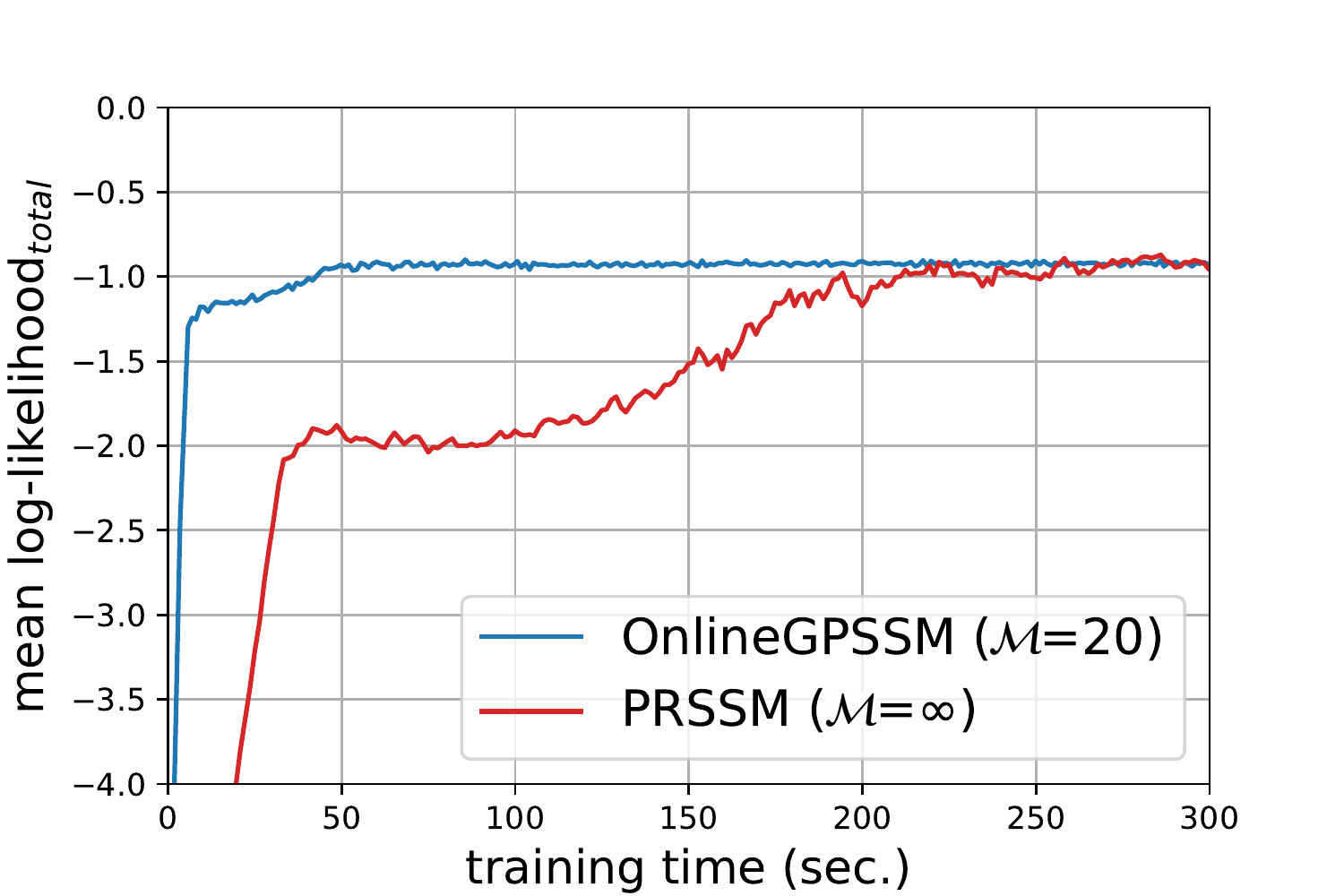}}
	\hspace{-0.2cm}
	\subfigure[Step 3]{
		\includegraphics[trim={0.1cm 0.cm 1.5cm 1.1cm},clip,width=0.45\columnwidth]{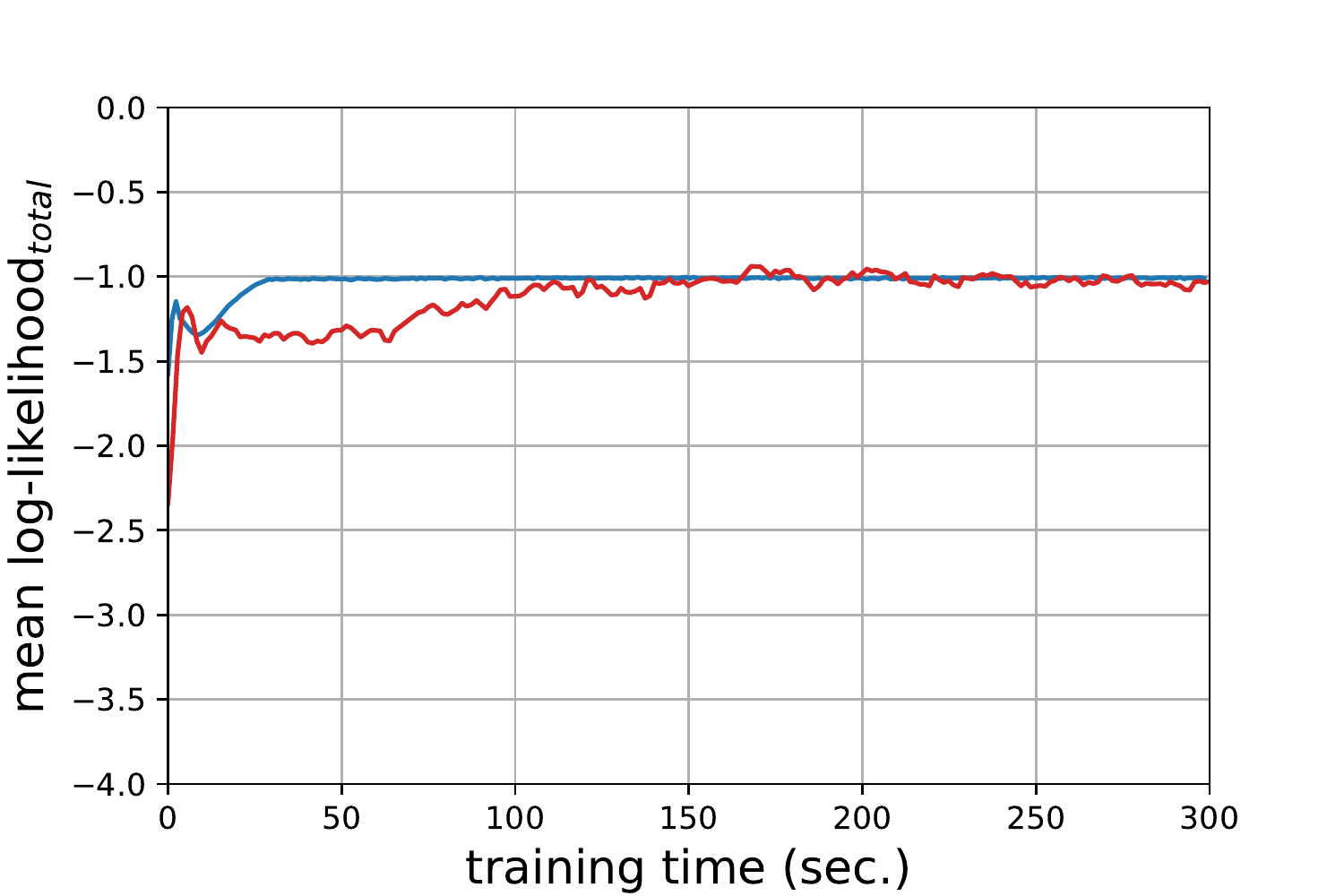}}\\
	\subfigure[Step 4]{
		\includegraphics[trim={0.1cm 0.cm 1.5cm 1.1cm},clip,width=0.45\columnwidth]{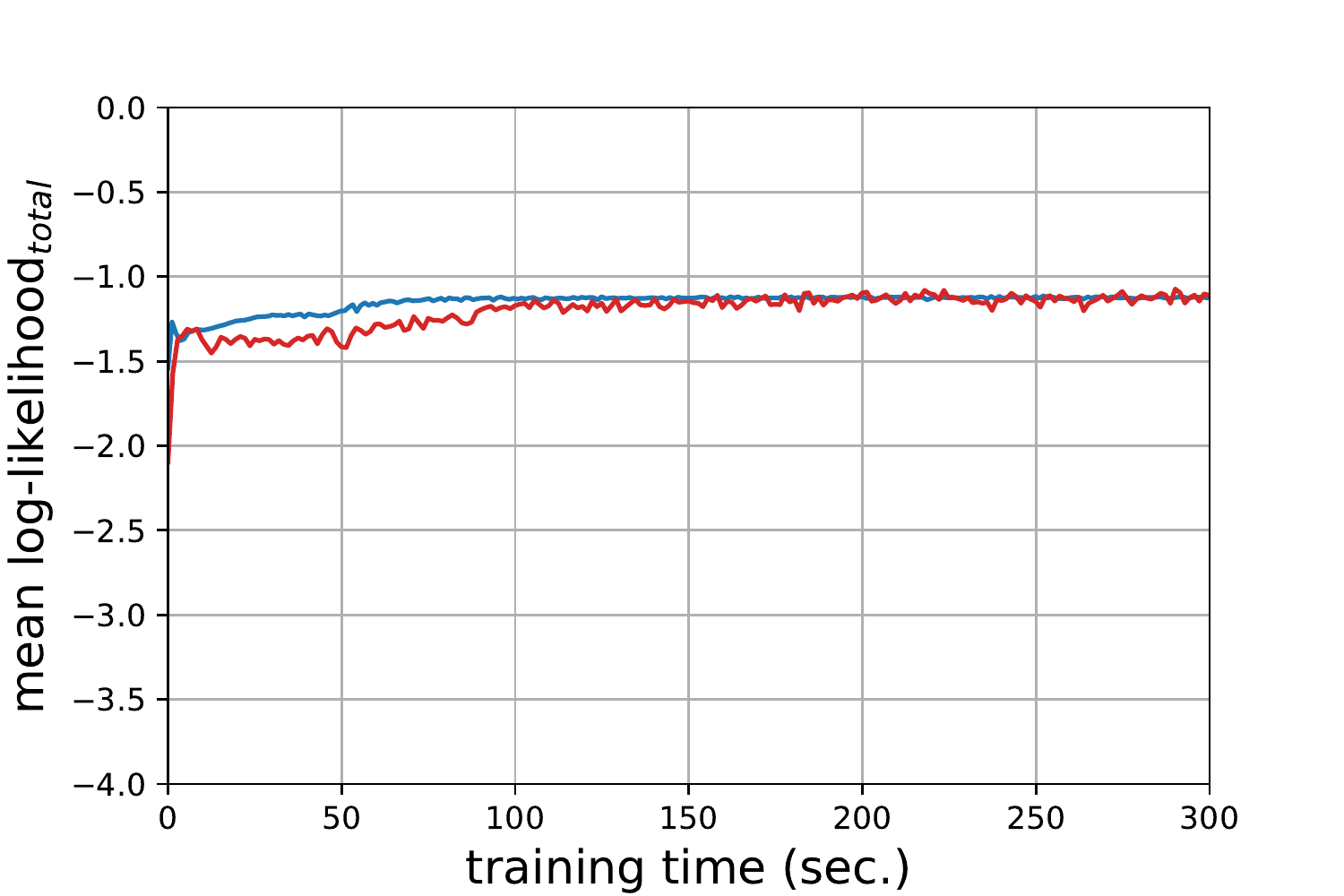}}
	\hspace{-0.2cm}
	\subfigure[Step 5]{
		\includegraphics[trim={0.1cm 0.cm 1.5cm 1.1cm},clip,width=0.45\columnwidth]{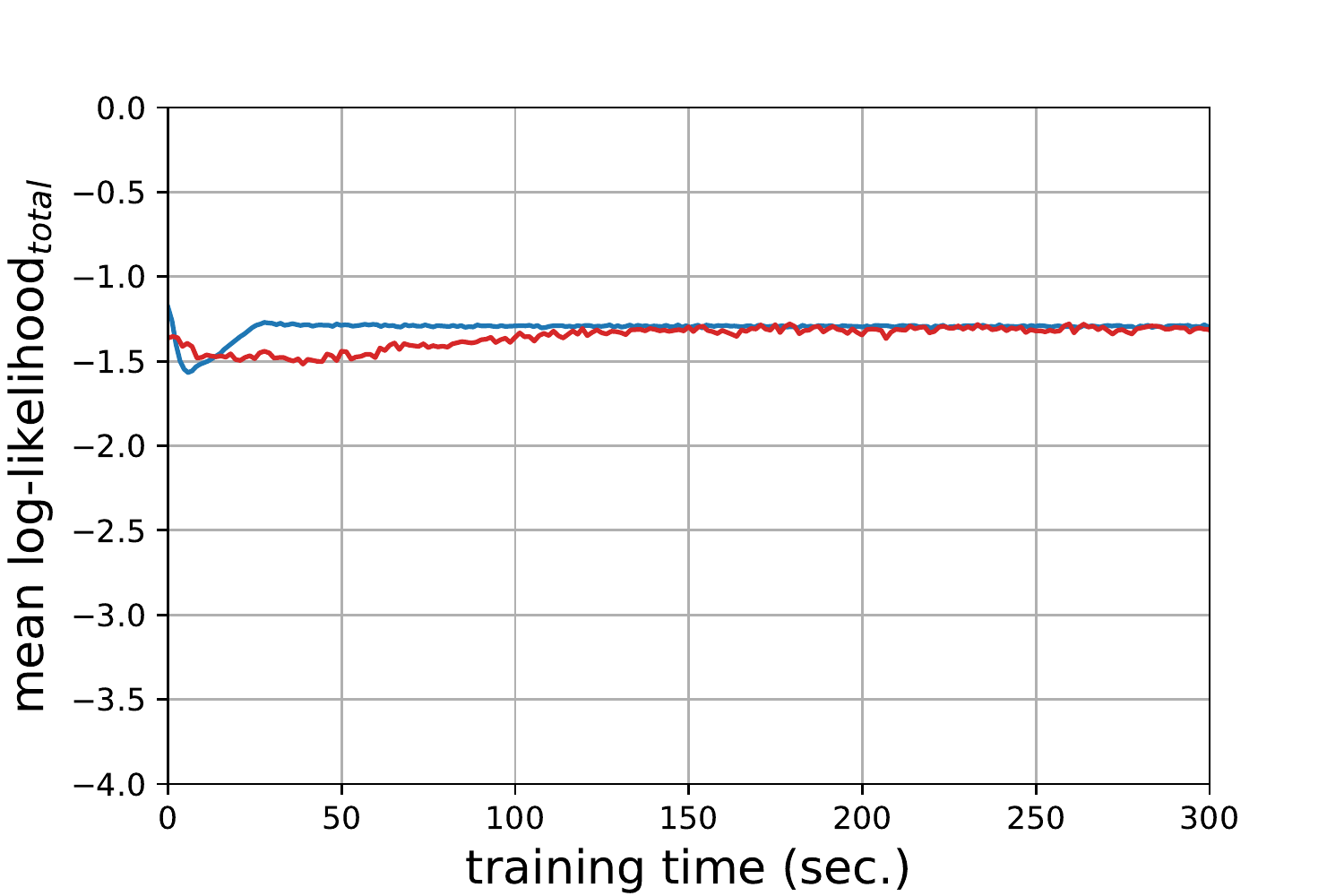}}
	\caption{Comparison of mean log-likelihood w.r.t. training time.}				
	\label{fig:ex1_loglik}
\end{figure}

{For further analysis, we check the efficiency of both algorithms by comparing the training time as shown in Figs. \ref{fig:ex1_time} and \ref{fig:ex1_loglik}. The comparison is performed on a 3.4 GHz Quad-Core Intel(TM) i7 PC. The results show that the onlineGPSSM achieves a solution with the same level of RMSE and mean log-likelihood much faster.}

\begin{figure*}[ht!]
	\centering
	\subfigure[Step 2]{
		\includegraphics[trim={.5cm 1.cm 1cm 1.5cm},clip,width=0.5\columnwidth]{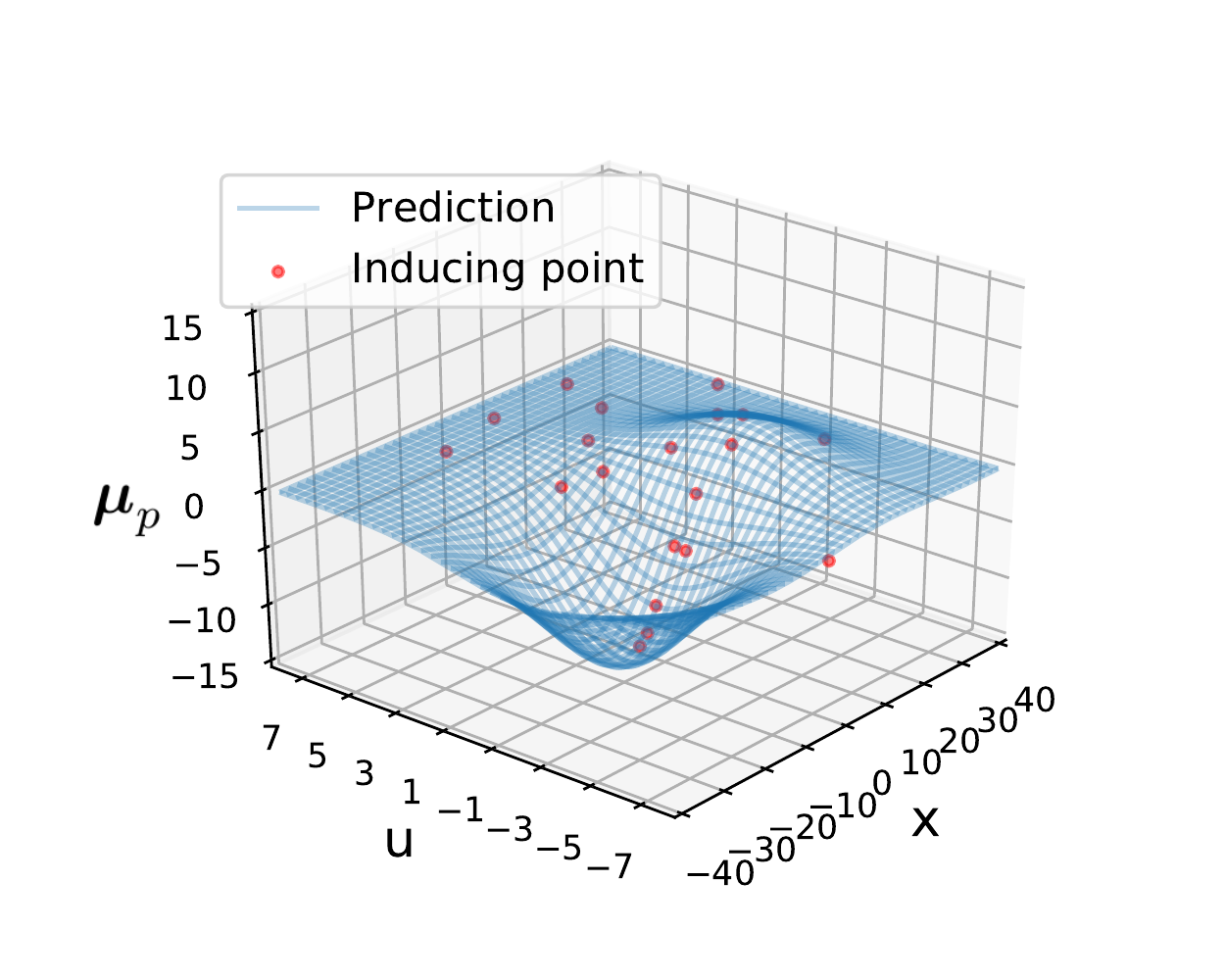}}
	\hspace*{-0.6cm}
	\subfigure[Step 3]{
		\includegraphics[trim={.5cm 1.cm 1cm 1.5cm},clip,width=0.5\columnwidth]{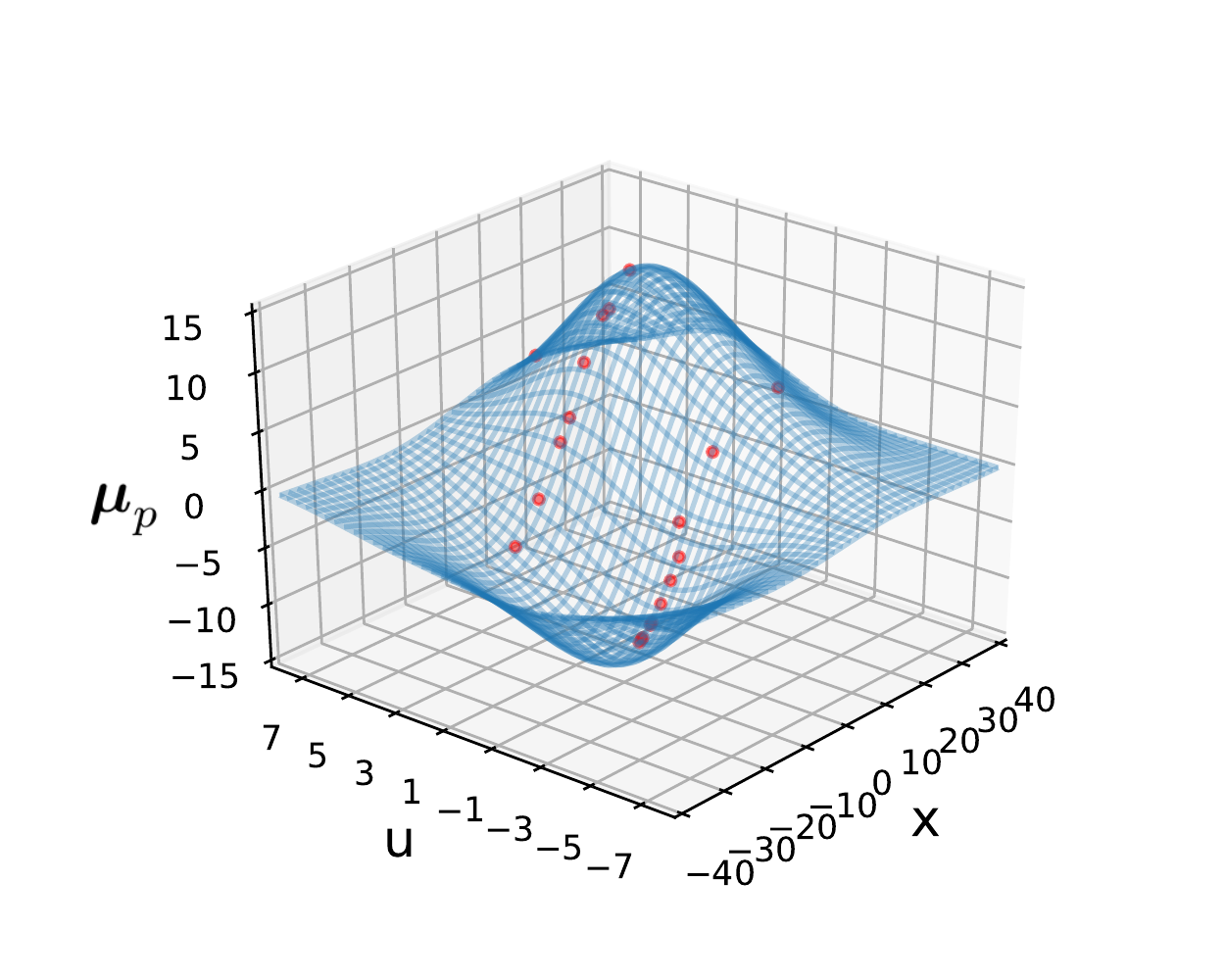}}
	\hspace*{-0.6cm}	
	\subfigure[Step 4]{
		\includegraphics[trim={.5cm 1.cm 1cm 1.5cm},clip,width=0.5\columnwidth]{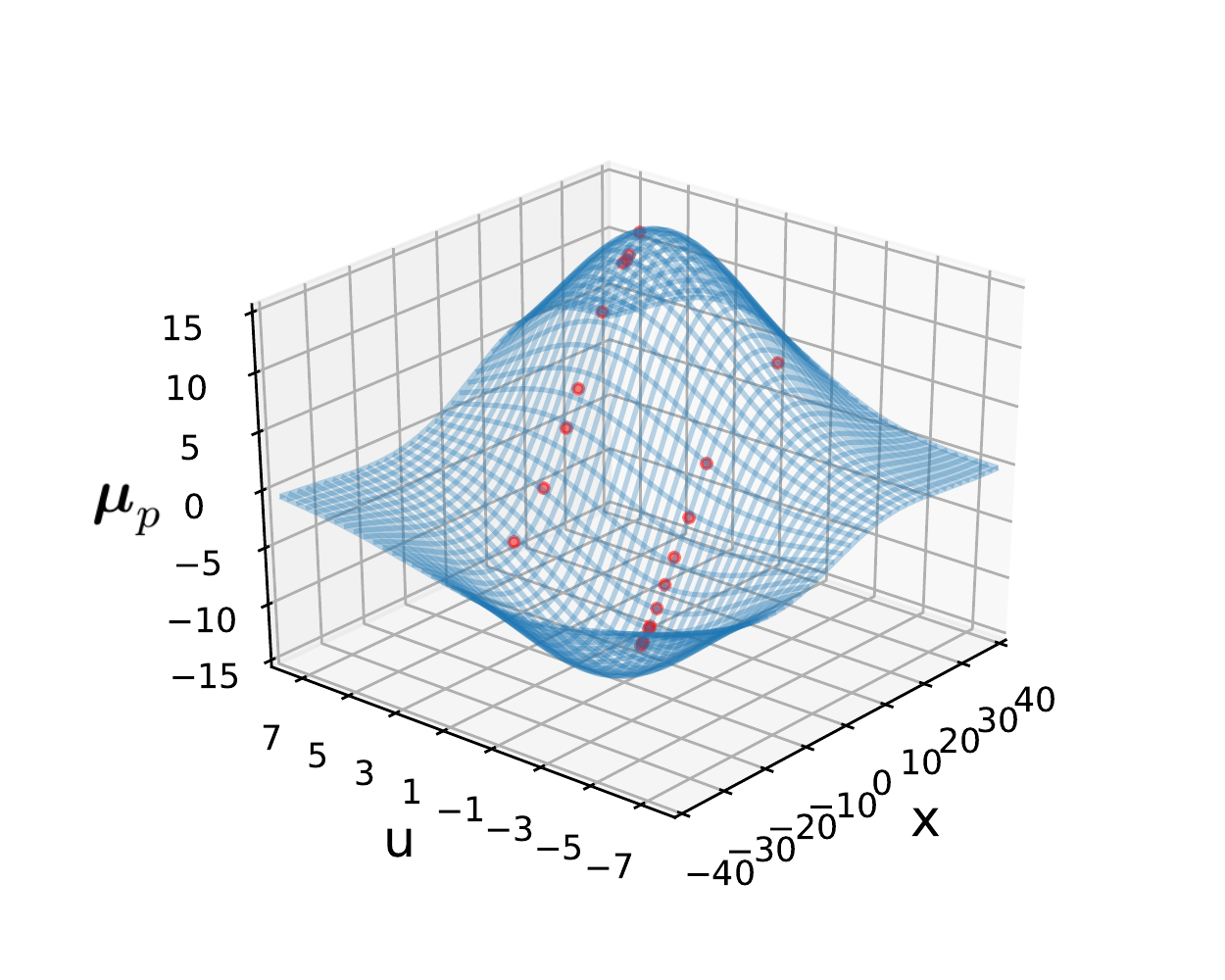}}
	\hspace*{-0.6cm}
	\subfigure[Step 5]{
		\includegraphics[trim={.5cm 1.cm 1cm 1.5cm},clip,width=0.5\columnwidth]{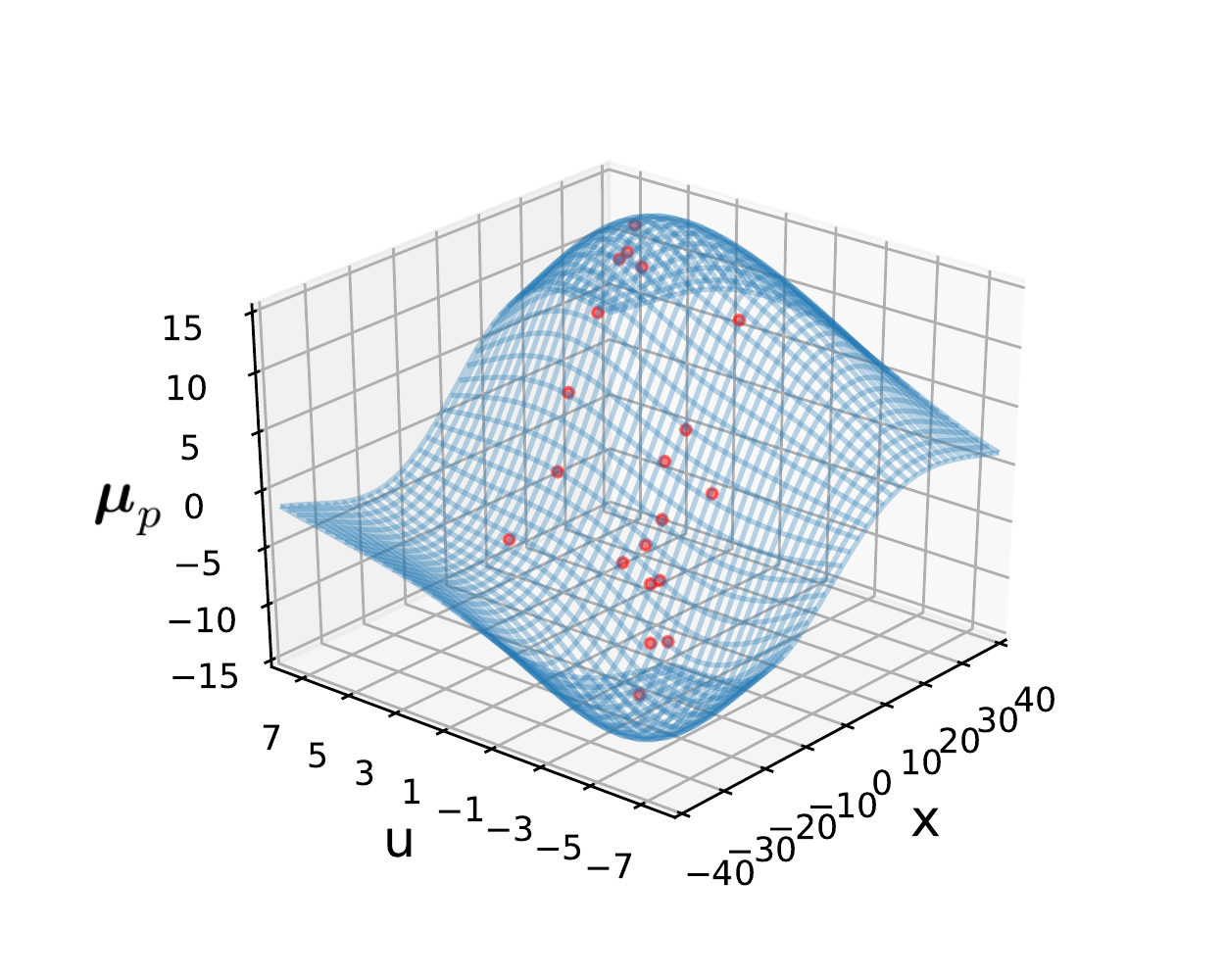}}\\
	\subfigure[Step 2]{
		\includegraphics[trim={.5cm 1.cm 1cm 1.5cm},clip,width=0.5\columnwidth]{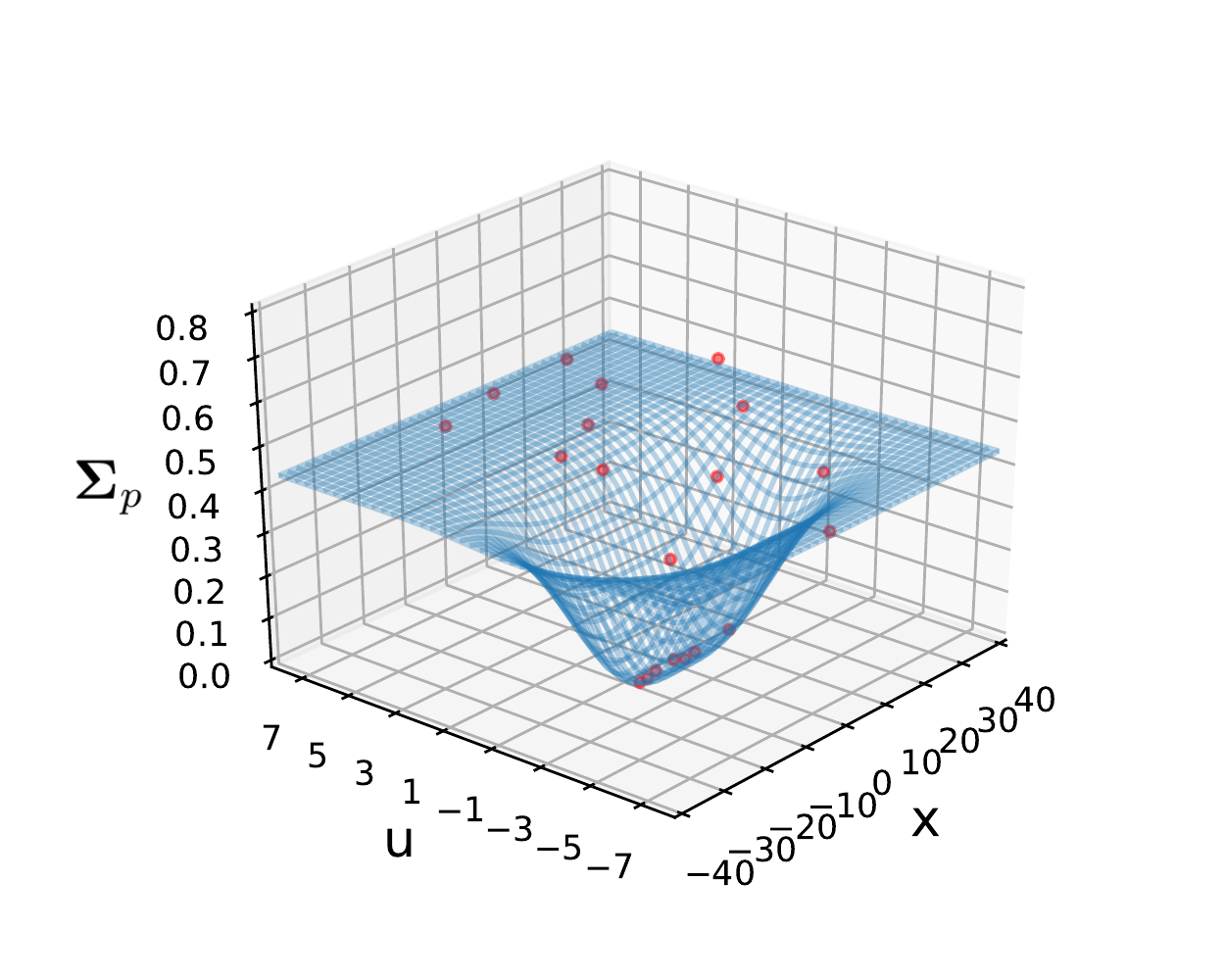}}
	\hspace*{-0.6cm}
	\subfigure[Step 3]{
		\includegraphics[trim={.5cm 1.cm 1cm 1.5cm},clip,width=0.5\columnwidth]{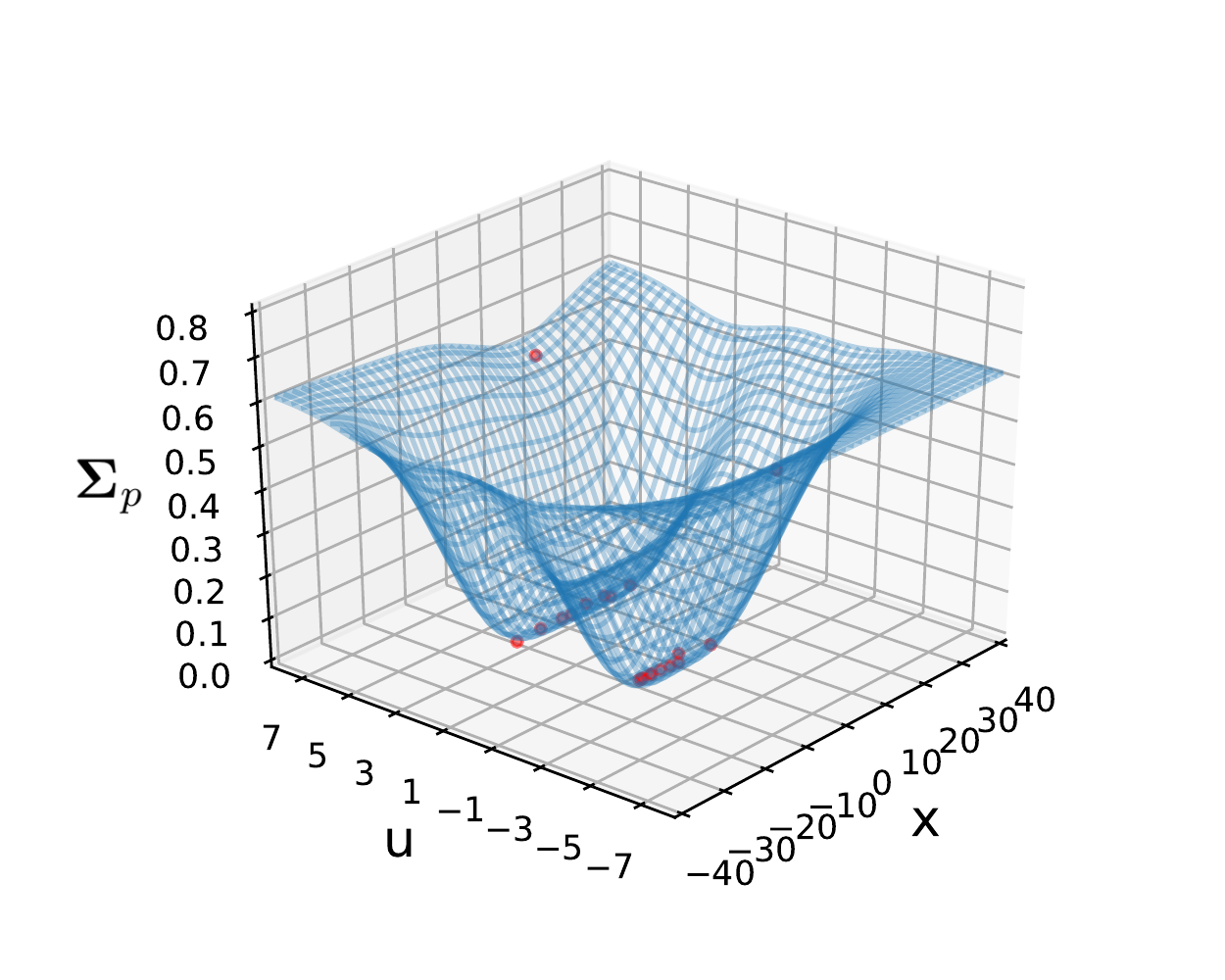}}
	\hspace*{-0.6cm}
	\subfigure[Step 4]{
		\includegraphics[trim={.5cm 1.cm 1cm 1.5cm},clip,width=0.5\columnwidth]{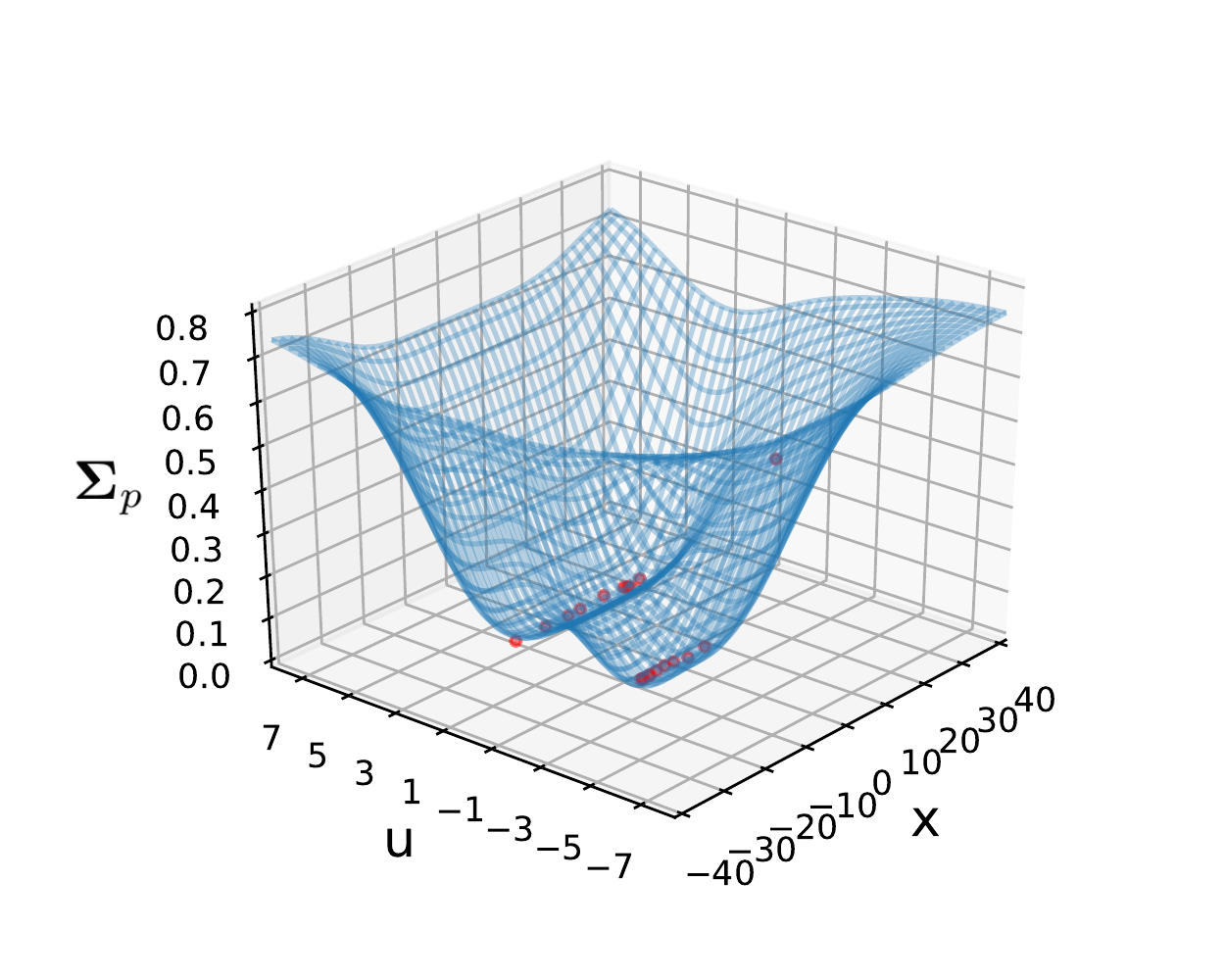}}
	\hspace*{-0.6cm}
	\subfigure[Step 5]{
		\includegraphics[trim={.5cm 1.cm 1cm 1.5cm},clip,width=0.5\columnwidth]{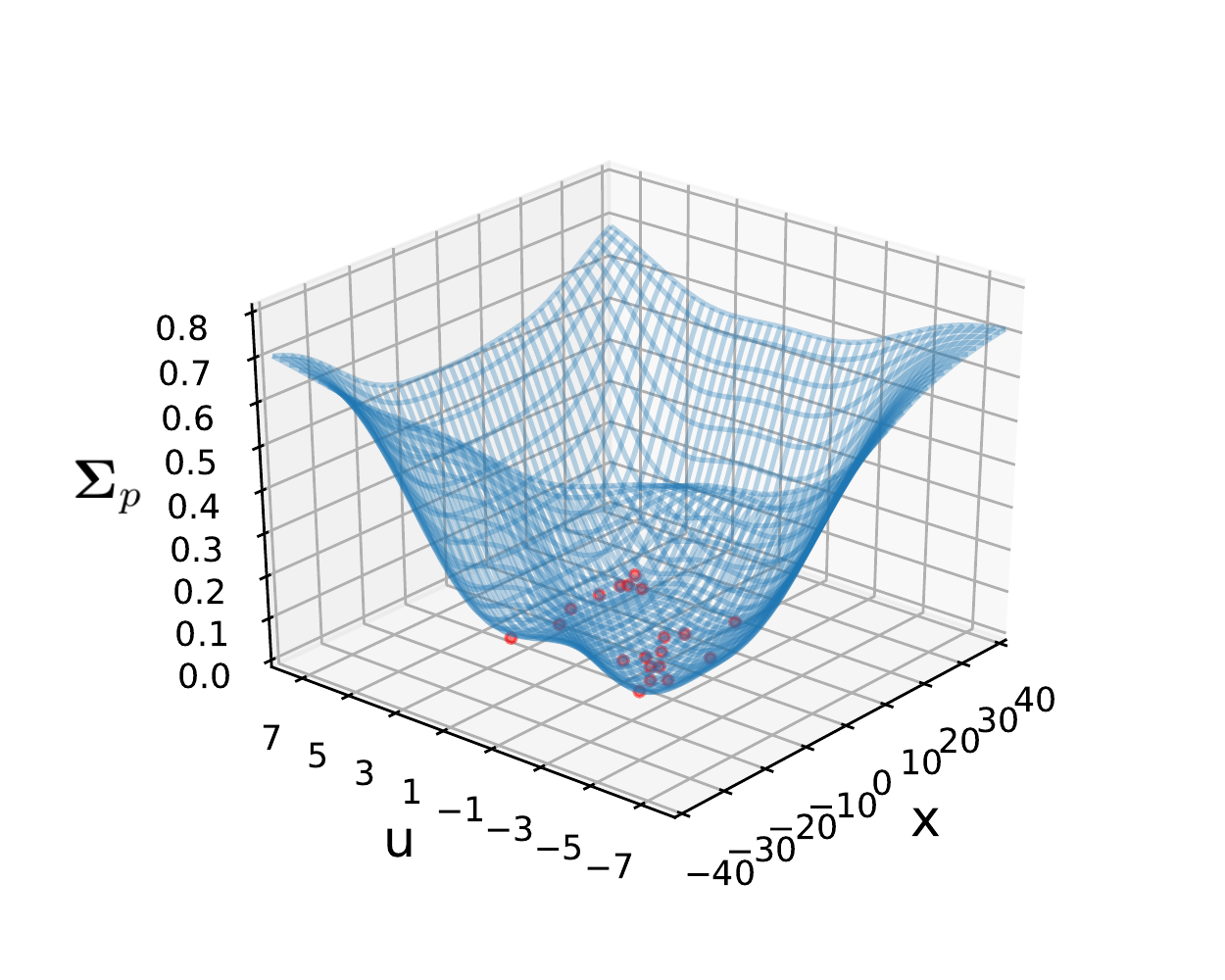}}\\
	\caption{Prediction results of onlineGPSSM ($\mathcal{M}$=20) after each online learning: (a)-(d) mean and (e)-(h) variance.}
	\label{fig:ex1_pred}
\end{figure*}

Figure \ref{fig:ex1_pred} shows the prediction results of mean and covariance after each online learning step. As the overall shapes evolve based on the previously learned shapes, the results show that the hyperparameters of the GP-SSM are optimized to generate new measurements while maintaining the learned system knowledge (i.e., the shape of mean and covariance). This gradual change is enforced by the previous learning from the past measurements as expressed in the last two terms of \eqref{eq:NELBO}.
Since the zero mean function was used for initialization in this experiment, the mean value takes zero for unexperienced areas. However, this prediction is not overconfident in the regions with scarce data as shown through the large covariance values.


\begin{table}[t]\centering
	\caption{Comparison of the RMSE.}	\label{tbl:rmse}
	\tabcolsep=0.11cm
	\begin{tabular}{c c ccccc}
		\toprule
		\multirow{2}[3]{*}{} & \multirow{2}[3]{*}{} & \multicolumn{5}{c}{Online learning step} \\
		\cmidrule(lr){3-7}
		& ~& 1 & 2            & 3 & 4           & 5          \\
		\midrule
		\multirow{2}[5]{*}{\shortstack{SSGP\cite{bui2017streaming}\\($\mathcal{M}$ = 20)}}
		& old    &   --   & 0.5657 & 1.8235 & 1.5026 & 1.8905 \\
		& new    & 0.6829 & 1.5884 & 0.6164 & 0.6380 & 0.1183 \\
		& total  & 0.6829 & 1.1923 & 1.5308 & 1.3398 & 1.6918 \\
		\midrule
		\multirow{2}[5]{*}{\shortstack{streming-VB\cite{frigola2015bayesian}\\($\mathcal{M}$ = 20)}}
		& old    &   --   & 0.5552 & 1.0874 & 1.0842 & 1.1856 \\
		& new    & 0.2584 & 0.4187 & 0.5501 & 0.2032 & \textbf{0.2195} \\
		& total  & 0.2584 & 0.4917 & 0.9429 & 0.9445 & 1.0650 \\
		\midrule
		\multirow{2}[5]{*}{\shortstack{PRSSM\cite{doerr2018PRSSM}\\($\mathcal{M}$ = 20)}}
		& old    &   --   & 2.3956 & 7.0378 & 9.0272 & 10.1632 \\
		& new    & 0.2584 & 0.3532 & \textbf{0.3431} & 0.3179 & 0.2958 \\
		& total  & 0.2584 & 1.7122 & 5.7498 & 7.8194 & 9.0912 \\
		\midrule
		\multirow{2}[5]{*}{\shortstack{PRSSM\cite{doerr2018PRSSM}\\($\mathcal{M}$ = $\infty$)}}
		& old    &   --   & 0.4049 & 0.5458 & 0.4484 & 0.7023 \\
		& new    & 0.2584 & \textbf{0.3121} & 0.3720 & \textbf{0.1507} & 0.3682 \\
		& total  & 0.2584 & 0.3765 & \textbf{0.4546} & 0.3860 & 0.6421 \\
		\midrule
		\multirow{2}[5]{*}{\shortstack{onlineGPSSM\\ ($\mathcal{M}$ = 20)}}
		& old    &   --   & \textbf{0.3868} & \textbf{0.5398} & \textbf{0.4367} & \textbf{0.6915} \\
		& new    & 0.2584 & 0.3316 & 0.3826 & 0.1538 & 0.3505 \\
		& total  & 0.2584 & \textbf{0.3557} & 0.4703 & \textbf{0.3859} & \textbf{0.6381} \\
		\bottomrule
	\end{tabular} \label{tbl:RMSE_comp}
\end{table}

\begin{figure*}[ht!]
	\centering
	\subfigure[0-140 sec.]{
		\includegraphics[width=0.6\columnwidth]{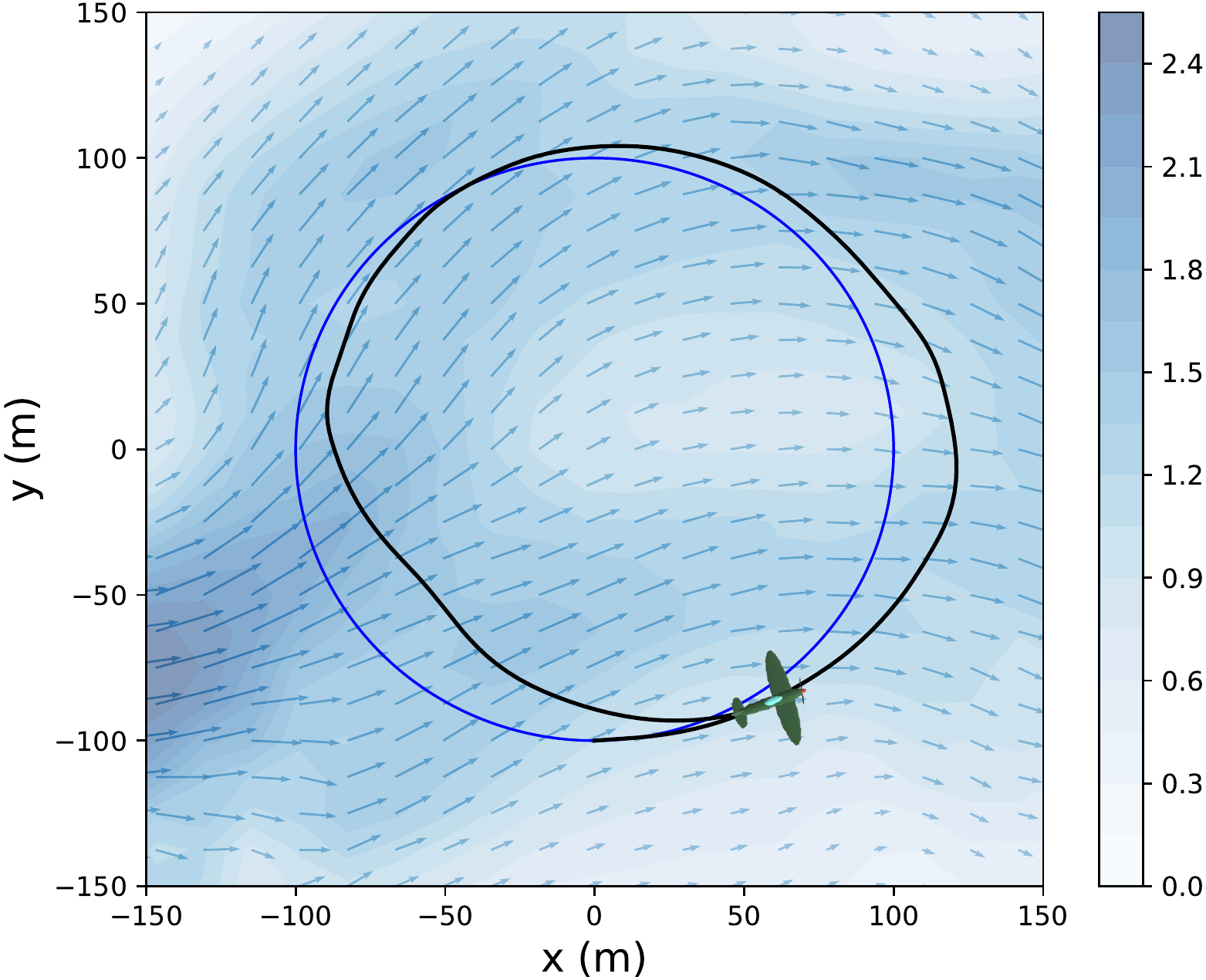}}
	\subfigure[200-340 sec.]{
		\includegraphics[width=0.6\columnwidth]{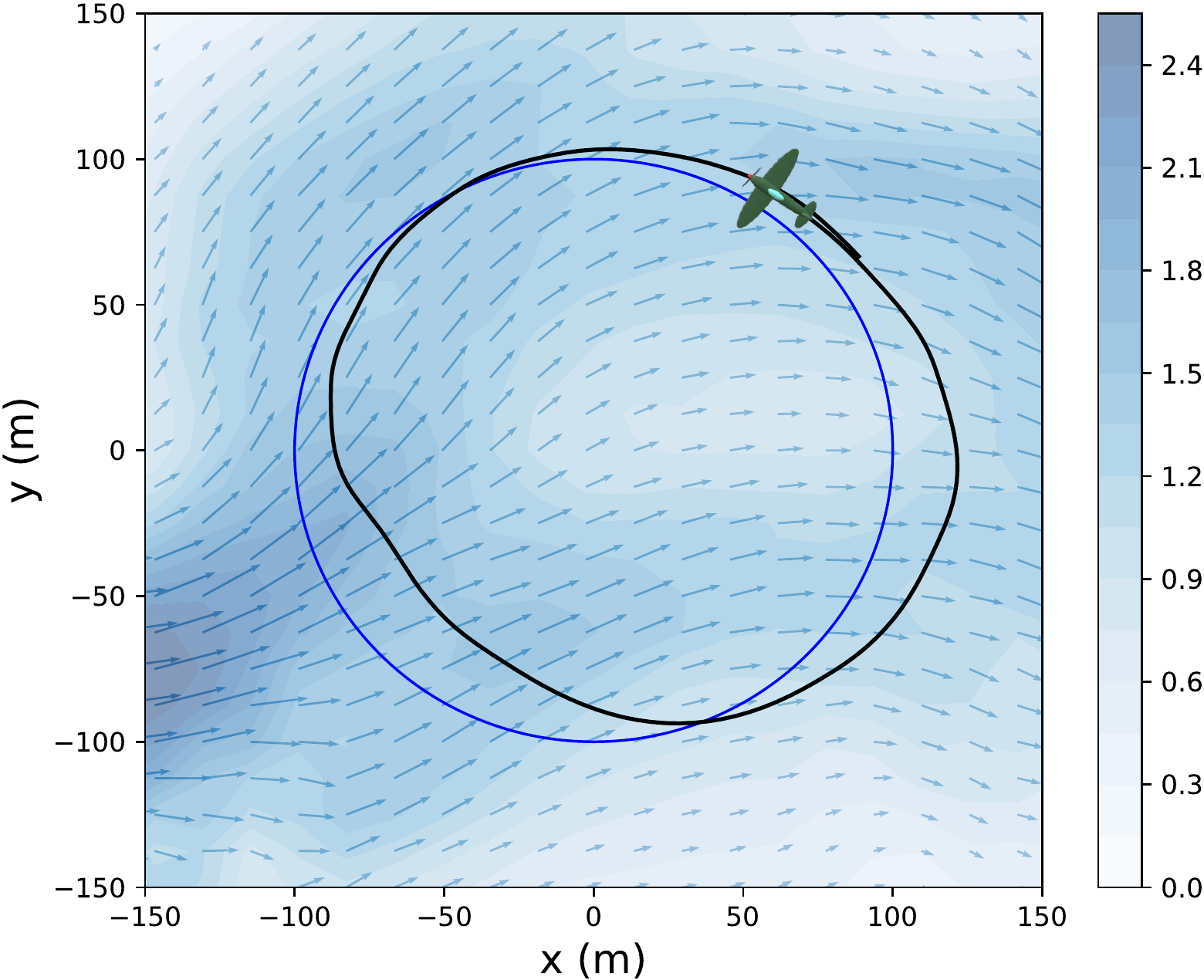}}
	\subfigure[460-600 sec.]{
		\includegraphics[width=0.6\columnwidth]{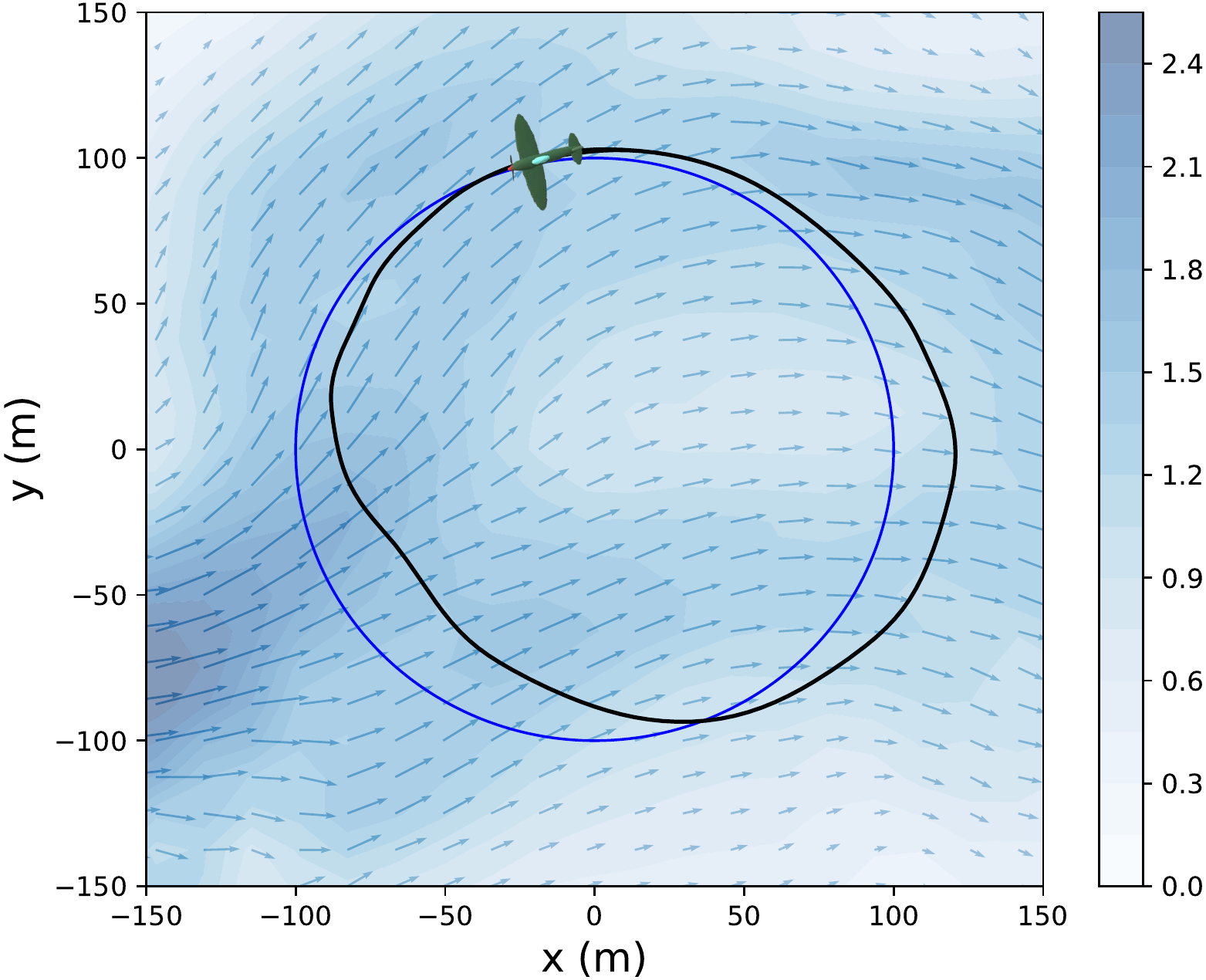}}\\
	\subfigure[0-140 sec.]{
		\includegraphics[width=0.6\columnwidth]{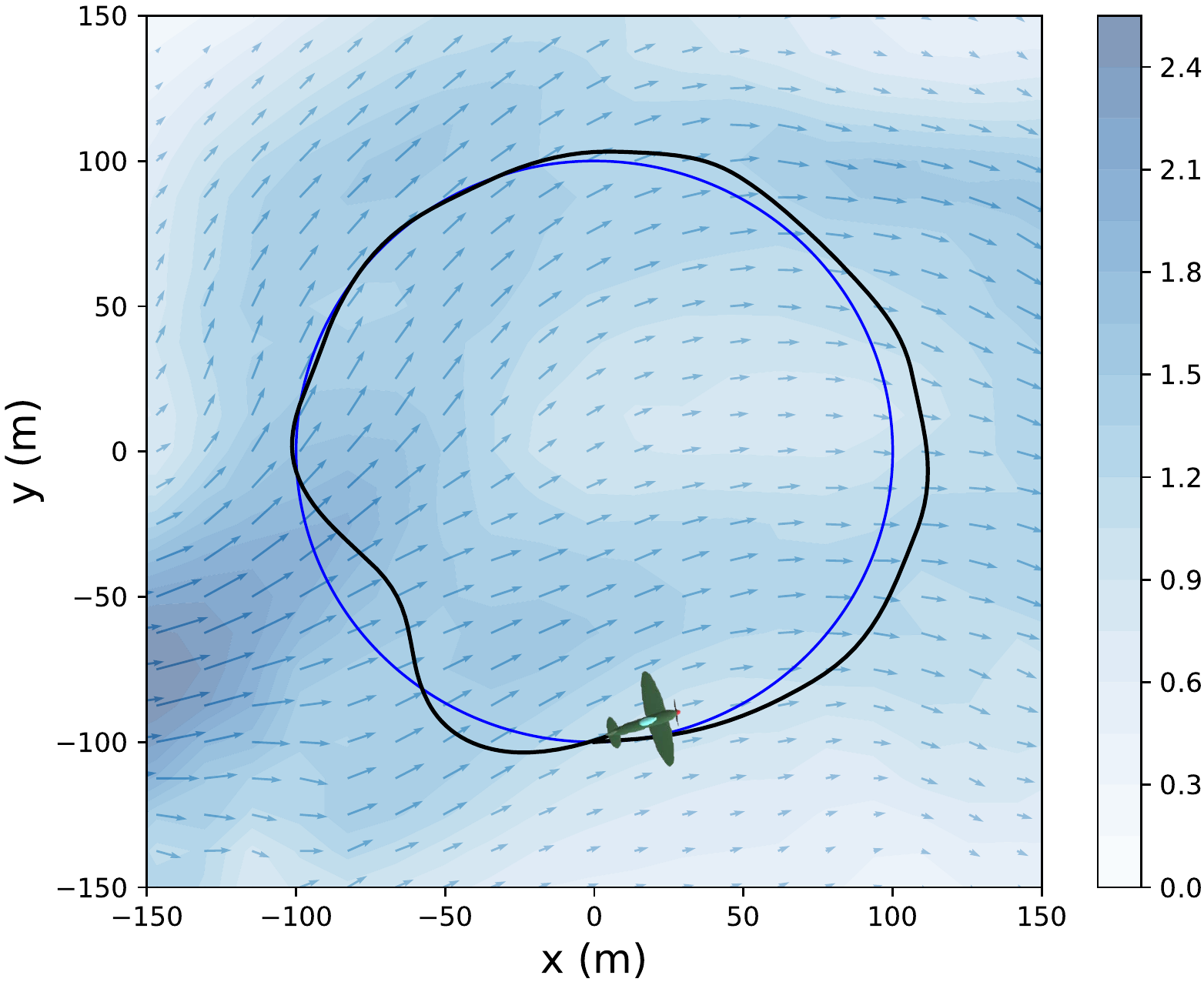}}
	\subfigure[200-340 sec.]{
		\includegraphics[width=0.6\columnwidth]{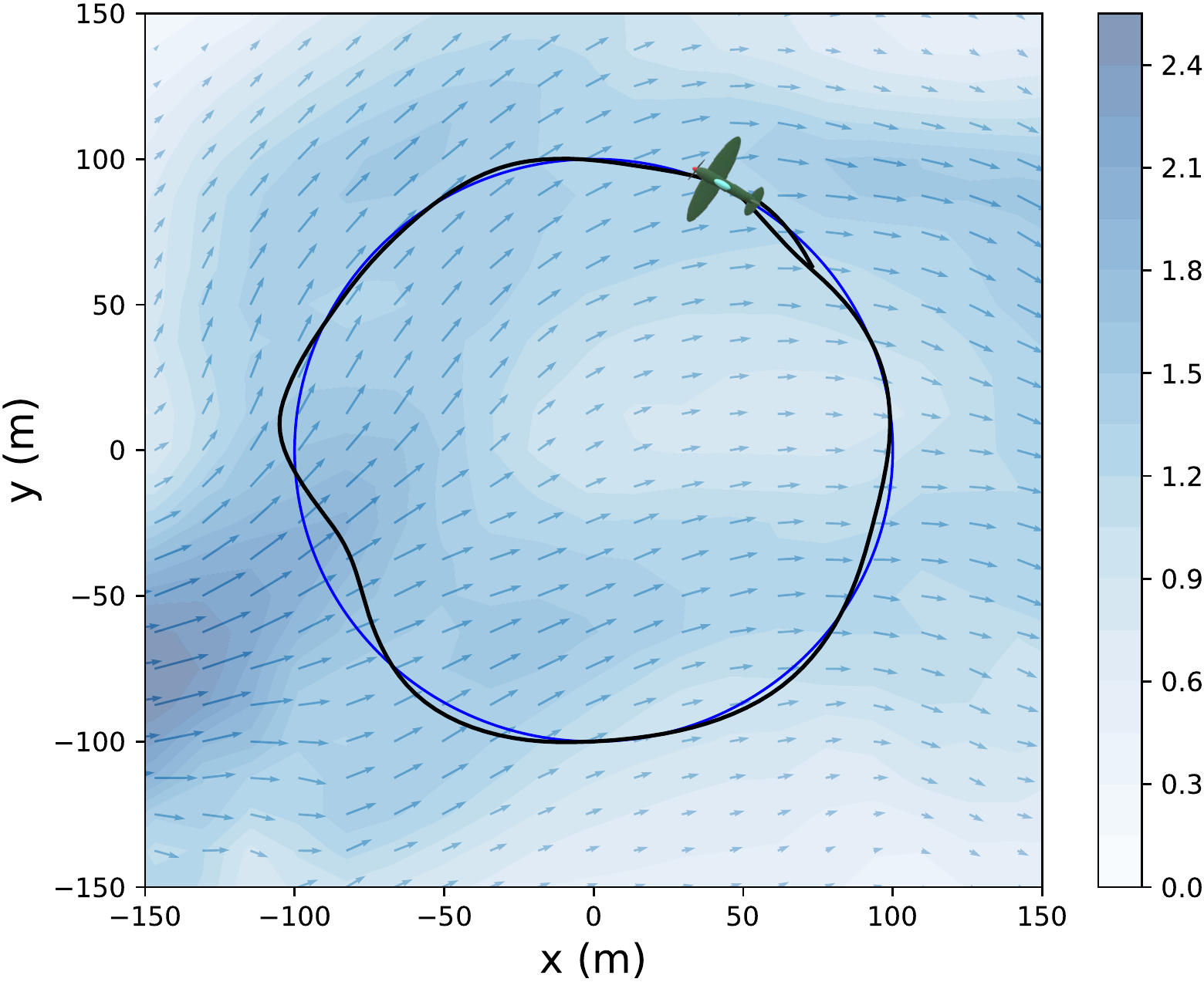}}
	\subfigure[460-600 sec.]{
		\includegraphics[width=0.6\columnwidth]{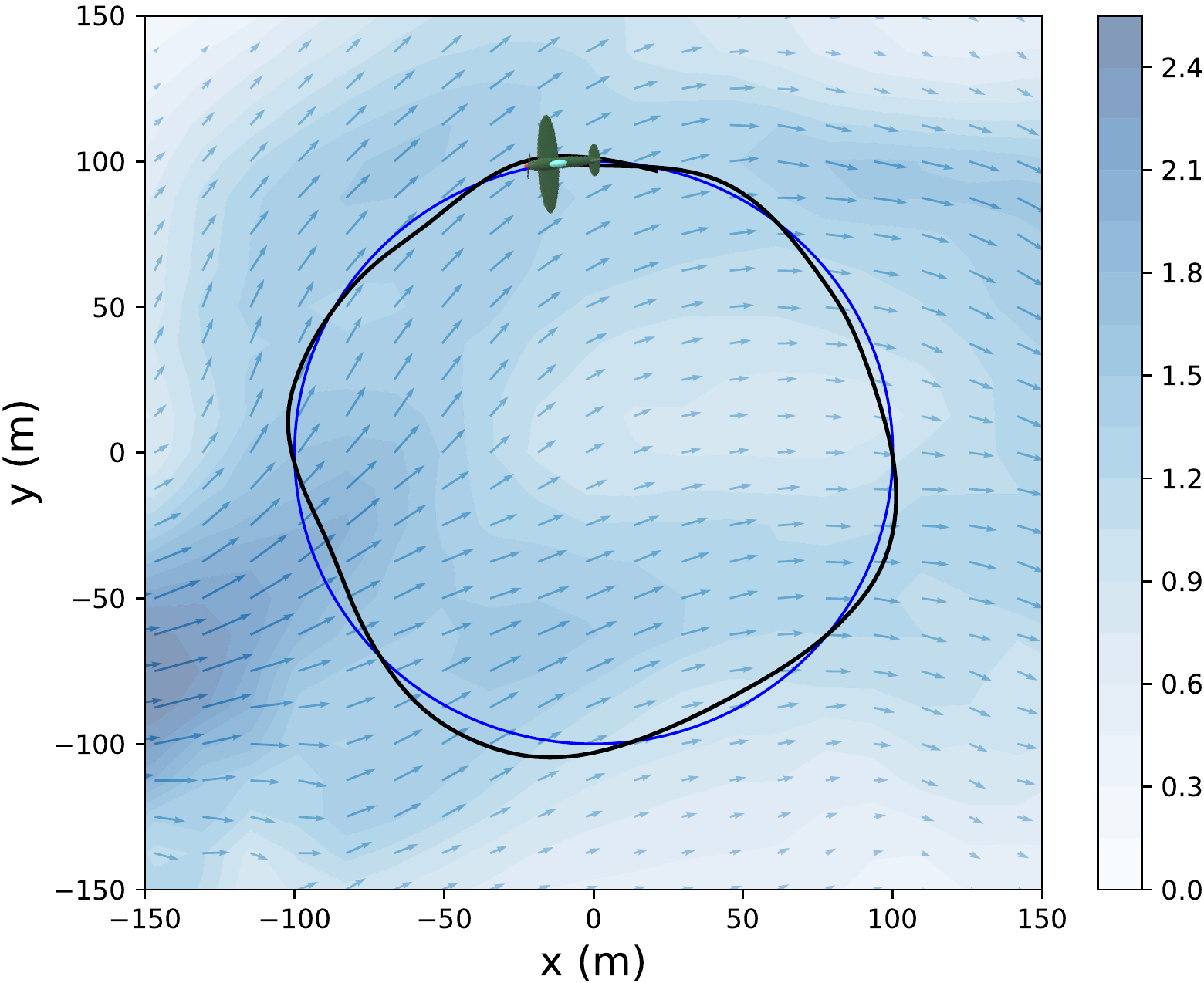}}
	\caption{Resulting trajectories: (a)-(c) belief space iLQG, (d)-(f) proposed algorithm. Blue lines: reference trajectories. Black lines: actual UAV trajectories.}
	\label{fig:ex3_snaptshot}
\end{figure*}

\subsection{Loitering of fixed-wing UAV in a wind field}

In this example, we consider a circular loitering task of a fixed-wing UAV.
Such a task is required for precise communication relay \cite{majumdar2014advanced} as well as persistent monitoring of specific areas \cite{everaerts2008use}.
The true dynamics is given by\cite{guerrero2013quad,al2013wind}:
\begin{equation}
\begin{split}
x_{1,t\!+\!1} &= x_{1,t} + \big(V \cos x_{3,t} + W_1(x_{1,t},x_{2,t})\big)dt + w_{f,1},\\
x_{2,t\!+\!1} &= x_{2,t} + \big(V \sin x_{3,t} + W_2(x_{1,t},x_{2,t})\big)dt + w_{f,2},\\
x_{3,t\!+\!1} &= x_{3,t} + \frac{g_a \tan x_{4,t}}{V}dt + w_{f,3},\\
x_{4,t\!+\!1} &= x_{4,t} + u_t dt + w_{f,4},\\
y_{1,t} &= x_{1,t} + w_{g,1},\\
y_{2,t} &= x_{2,t} + w_{g,2},
\end{split}
\end{equation}
where $[x_{1},x_{2}]$, $x_3$, $x_4$, and $V$ are the position, the heading angle, and the bank angle of the UAV and the airspeed (5m/s), respectively. The gravity acceleration is denoted as $g_a$. The time interval between two successive states is set as $dt = 0.1s$.
$W_{1}$ and $W_{2}$ are the wind magnitude for the $x$ and $y$ directions, respectively.
The wind cannot be directly observed, and its effect on the dynamics is assumed unknown. A UAV can only measure its position information from GPS signals, which inherently contains noise\cite{misra2006global} ($\Sigma_{g,1} = \Sigma_{g,2} = 1$). Furthermore, we set $\Sigma_{f,1} = \Sigma_{f,2} = 0.1$ and $\Sigma_{f,3} = \Sigma_{f,4} = 0.01\times180/\pi$.

The cost function and weights are given by:
\begin{equation}\small
\begin{split}
Q_T = &Q = 2\boldsymbol{I}, \enspace \tilde{Q}_T=\boldsymbol{0}, \enspace R = 100, \enspace \\
&\alpha(\boldsymbol{\mu}_{t}) = \big( \sqrt{\mu_{1,t}^2 + \mu_{2,t}^2} - R_{des} \big)^2,
\end{split}
\end{equation}
where $R_{des}$ is the radius of rotation of the circle centered on the origin that the UAV should track continuously.
We used the dynamics without wind field as the mean prior of the GP-SSM.
For the storage, we save the state estimation results every 3 seconds and the measurements of the latest 12 seconds in memory ($\mathcal{M}=120$).
Online learning is performed for four mini-batch training sets consisting of 30 data points. We set $T_p$ and $T_c$ to 10s and 3s, respectively. Model learning and trajectory optimization are performed concurrently.


We verify the effectiveness of the proposed online learning and planning algorithm by comparing it with the belief space iLQG\cite{van2012BiLQG} in the same environment.
The belief space iLQG computes and executes the control inputs in the same RHC scheme as the proposed framework without knowing the influence of the wind. Fig. \ref{fig:ex3_snaptshot} shows some snapshots of the resulting trajectories from both algorithms.
The belief space iLQG combined with the RHC scheme continually reflects changes in the position of the UAV to generate the control inputs. However, an inaccurate system model causes a difference between the planned trajectory and the actual trajectory, which eventually hinders the performance of the given task (see Fig. \ref{fig:ex3_snaptshot} (a)-(c)).
On the other hand, as shown in Fig. \ref{fig:ex3_snaptshot} (d)-(f), the proposed framework shows improved performance as the task is repeated by learning the system model in an online manner while interacting with an unknown environment.

\section{Conclusion}\label{sec:conclusion}
In this paper, we proposed an online inference method for learning GP-SSM (onlineGPSSM) that  handles sequentially arriving data without catastrophic forgetting and supports adaptation to changes in a system and/or environment.
The onlineGPSSM was shown to mitigate the memory and computation time issues.
By integrating the onlineGPSSM with trajectory optimization and Bayesian filtering algorithms, the GP-SSM-based reinforcement learning (RL) framework has been developed, which can generate a robust control policy for control/planning.
Numerical experiments were presented to demonstrate the applicability and validity of the onlineGPSSM and GP-SSM-based RL framework.


	\addtolength{\textheight}{-1cm}   
	



	\bibliographystyle{IEEEtran}
	\bibliography{manuscript}

\begin{thebibliography}{10}
\providecommand{\url}[1]{#1}
\csname url@samestyle\endcsname
\providecommand{\newblock}{\relax}
\providecommand{\bibinfo}[2]{#2}
\providecommand{\BIBentrySTDinterwordspacing}{\spaceskip=0pt\relax}
\providecommand{\BIBentryALTinterwordstretchfactor}{4}
\providecommand{\BIBentryALTinterwordspacing}{\spaceskip=\fontdimen2\font plus
\BIBentryALTinterwordstretchfactor\fontdimen3\font minus
  \fontdimen4\font\relax}
\providecommand{\BIBforeignlanguage}[2]{{%
\expandafter\ifx\csname l@#1\endcsname\relax
\typeout{** WARNING: IEEEtran.bst: No hyphenation pattern has been}%
\typeout{** loaded for the language `#1'. Using the pattern for}%
\typeout{** the default language instead.}%
\else
\language=\csname l@#1\endcsname
\fi
#2}}
\providecommand{\BIBdecl}{\relax}
\BIBdecl

\bibitem{ljung1998system}
L.~Ljung, ``System identification,'' in \emph{Signal analysis and
  prediction}.\hskip 1em plus 0.5em minus 0.4em\relax Springer, 1998, pp.
  163--173.

\bibitem{li2017identification}
F.~Li, J.~Li, and D.~Peng, ``Identification method of neuro-fuzzy-based
  hammerstein model with coloured noise,'' \emph{IET Control Theory \&
  Applications}, vol.~11, no.~17, pp. 3026--3037, 2017.

\bibitem{li2017neuro}
F.~Li, L.~Jia, D.~Peng, and C.~Han, ``Neuro-fuzzy based identification method
  for hammerstein output error model with colored noise,''
  \emph{Neurocomputing}, vol. 244, pp. 90--101, 2017.

\bibitem{eleftheriadis2017identification}
S.~Eleftheriadis, T.~Nicholson, M.~Deisenroth, and J.~Hensman, ``Identification
  of {Gaussian} process state space models,'' in \emph{Advances in Neural
  Information Processing Systems}, 2017, pp. 5309--5319.

\bibitem{ko2009gpkf}
J.~Ko and D.~Fox, ``{GP-BayesFilters}: {Bayesian} filtering using {Gaussian}
  process prediction and observation models,'' \emph{Autonomous Robots},
  vol.~27, no.~1, pp. 75--90, 2009.

\bibitem{boedecker2014approximate}
J.~Boedecker, J.~T. Springenberg, J.~W{\"u}lfing, and M.~Riedmiller,
  ``Approximate real-time optimal control based on sparse {Gaussian} process
  models,'' in \emph{Adaptive Dynamic Programming and Reinforcement Learning
  (ADPRL), 2014 IEEE Symposium on}.\hskip 1em plus 0.5em minus 0.4em\relax
  IEEE, 2014, pp. 1--8.

\bibitem{pan2014probabilistic}
Y.~Pan and E.~Theodorou, ``Probabilistic differential dynamic programming,'' in
  \emph{Advances in Neural Information Processing Systems}, 2014, pp.
  1907--1915.

\bibitem{deisenroth2015PILCO}
M.~P. Deisenroth, D.~Fox, and C.~E. Rasmussen, ``{Gaussian} processes for
  data-efficient learning in robotics and control,'' \emph{IEEE Transactions on
  Pattern Analysis and Machine Intelligence}, vol.~37, no.~2, pp. 408--423,
  2015.

\bibitem{mcallister2017data}
R.~McAllister and C.~E. Rasmussen, ``Data-efficient reinforcement learning in
  continuous state-action {Gaussian-POMDPs},'' in \emph{Advances in Neural
  Information Processing Systems}, 2017, pp. 2037--2046.

\bibitem{doerr2018PRSSM}
A.~Doerr, C.~Daniel, M.~Schiegg, N.-T. Duy, S.~Schaal, M.~Toussaint, and
  T.~Sebastian, ``Probabilistic recurrent state-space models,'' in
  \emph{Proceedings of the 35th International Conference on Machine Learning},
  2018, pp. 1280--1289.

\bibitem{sternberg2017identification}
W.~Sternberg and M.~P. Deisenroth, ``Identification of {Gaussian} process
  state-space models,'' 2017.

\bibitem{frigola2014variational}
R.~Frigola, Y.~Chen, and C.~E. Rasmussen, ``Variational {Gaussian} process
  state-space models,'' in \emph{Advances in Neural Information Processing
  Systems}, 2014, pp. 3680--3688.

\bibitem{brown1998statistical}
E.~N. Brown, L.~M. Frank, D.~Tang, M.~C. Quirk, and M.~A. Wilson, ``A
  statistical paradigm for neural spike train decoding applied to position
  prediction from ensemble firing patterns of rat hippocampal place cells,''
  \emph{Journal of Neuroscience}, vol.~18, no.~18, pp. 7411--7425, 1998.

\bibitem{polydoros2017survey}
A.~S. Polydoros and L.~Nalpantidis, ``Survey of model-based reinforcement
  learning: Applications on robotics,'' \emph{Journal of Intelligent \& Robotic
  Systems}, vol.~86, no.~2, pp. 153--173, 2017.

\bibitem{frigola2013bayesian}
R.~Frigola, F.~Lindsten, T.~B. Sch{\"o}n, and C.~E. Rasmussen, ``{Bayesian}
  inference and learning in gaussian process state-space models with particle
  {MCMC},'' in \emph{Advances in Neural Information Processing Systems}, 2013,
  pp. 3156--3164.

\bibitem{frigola2014identification}
------, ``Identification of {Gaussian} process state-space models with particle
  stochastic approximation {EM},'' \emph{IFAC Proceedings Volumes}, vol.~47,
  no.~3, pp. 4097--4102, 2014.

\bibitem{svensson2017flexible}
A.~Svensson and T.~B. Sch{\"o}n, ``A flexible state--space model for learning
  nonlinear dynamical systems,'' \emph{Automatica}, vol.~80, pp. 189--199,
  2017.

\bibitem{ghahramani2000online}
Z.~Ghahramani and H.~Attias, ``Online variational {Bayesian} learning,'' in
  \emph{Slides from talk presented at NIPS workshop on Online Learning}, 2000.

\bibitem{sato2001online}
M.-A. Sato, ``Online model selection based on the variational bayes,''
  \emph{Neural computation}, vol.~13, no.~7, pp. 1649--1681, 2001.

\bibitem{broderick2013streaming}
T.~Broderick, N.~Boyd, A.~Wibisono, A.~C. Wilson, and M.~I. Jordan, ``Streaming
  variational {Bayes},'' in \emph{Advances in Neural Information Processing
  Systems}, 2013, pp. 1727--1735.

\bibitem{salimbeni2017doubly}
H.~Salimbeni and M.~Deisenroth, ``Doubly stochastic variational inference for
  deep gaussian processes,'' in \emph{Advances in Neural Information Processing
  Systems}, 2017, pp. 4588--4599.

\bibitem{nguyen2017variational}
C.~V. Nguyen, Y.~Li, T.~D. Bui, and R.~E. Turner, ``Variational continual
  learning,'' \emph{arXiv preprint arXiv:1710.10628}, 2017.

\bibitem{bui2017streaming}
T.~D. Bui, C.~Nguyen, and R.~E. Turner, ``Streaming sparse {Gaussian} process
  approximations,'' in \emph{Advances in Neural Information Processing
  Systems}, 2017, pp. 3301--3309.

\bibitem{berntorp2019recursive}
K.~Berntorp, ``Recursive bayesian inference and learning of gaussian-process
  state-space models,'' in \emph{2019 18th European Control Conference
  (ECC)}.\hskip 1em plus 0.5em minus 0.4em\relax IEEE, 2019, pp. 1866--1871.

\bibitem{kingma2013auto}
D.~P. Kingma and M.~Welling, ``Auto-encoding variational bayes,'' \emph{arXiv
  preprint arXiv:1312.6114}, 2013.

\bibitem{kober2013reinforcement}
J.~Kober, J.~A. Bagnell, and J.~Peters, ``Reinforcement learning in robotics: A
  survey,'' \emph{The International Journal of Robotics Research}, vol.~32,
  no.~11, pp. 1238--1274, 2013.

\bibitem{mitrovic2010adaptive}
D.~Mitrovic, S.~Klanke, and S.~Vijayakumar, ``Adaptive optimal feedback control
  with learned internal dynamics models,'' in \emph{From Motor Learning to
  Interaction Learning in Robots}.\hskip 1em plus 0.5em minus 0.4em\relax
  Springer, 2010, pp. 65--84.

\bibitem{morimoto2003minimax}
J.~Morimoto, G.~Zeglin, and C.~G. Atkeson, ``Minimax differential dynamic
  programming,'' in \emph{SICE Annual Conference Program and Abstracts SICE
  Annual Conference 2003}.\hskip 1em plus 0.5em minus 0.4em\relax The Society
  of Instrument and Control Engineers, 2003, pp. 106--106.

\bibitem{park2019bayesian}
S.~S. Park, Y.-J. Park, and H.-L. Choi, ``A bayesian approach to learning and
  planning for partially observable dynamical systems,'' in \emph{AIAA Scitech
  2019 Forum}, 2019, p. 0398.

\bibitem{van2012BiLQG}
J.~Van Den~Berg, S.~Patil, and R.~Alterovitz, ``Motion planning under
  uncertainty using iterative local optimization in belief space,'' \emph{The
  International Journal of Robotics Research}, vol.~31, no.~11, pp. 1263--1278,
  2012.

\bibitem{frigola2015bayesian}
R.~Frigola-Alcade, ``{Bayesian} time series learning with {Gaussian}
  processes,'' \emph{Uni-versity of Cambridge}, 2015.

\bibitem{GP}
C.~E. Rasmussen, ``{Gaussian} processes in machine learning,'' in
  \emph{Advanced lectures on machine learning}.\hskip 1em plus 0.5em minus
  0.4em\relax Springer, 2004, pp. 63--71.

\bibitem{park2018deep}
Y.-J. Park, P.~M. Tagade, and H.-L. Choi, ``Deep {Gaussian} process-based
  bayesian inference for contaminant source localization,'' \emph{IEEE Access},
  vol.~6, pp. 49\,432--49\,449, 2018.

\bibitem{titsias2009variational}
M.~Titsias, ``Variational learning of inducing variables in sparse {Gaussian}
  processes,'' in \emph{Artificial Intelligence and Statistics}, 2009, pp.
  567--574.

\bibitem{chen2003bayesian}
Z.~Chen \emph{et~al.}, ``{Bayesian} filtering: From {Kalman} filters to
  particle filters, and beyond,'' \emph{Statistics}, vol. 182, no.~1, pp.
  1--69, 2003.

\bibitem{jacobson1970DDP}
D.~H. Jacobson and D.~Q. Mayne, ``Differential dynamic programming,'' 1970.

\bibitem{kuwata2004three}
Y.~Kuwata and J.~How, ``Three dimensional receding horizon control for uavs,''
  in \emph{AIAA Guidance, Navigation, and Control Conference and Exhibit},
  2004, p. 5144.

\bibitem{park2019distributed}
S.-S. Park, Y.~Min, J.-S. Ha, D.-H. Cho, and H.-L. Choi, ``A distributed admm
  approach to non-myopic path planning for multi-target tracking,'' \emph{IEEE
  Access}, vol.~7, pp. 163\,589--163\,603, 2019.

\bibitem{awasthi2019continual}
A.~Awasthi and S.~Sarawagi, ``Continual learning with neural networks: A
  review,'' in \emph{Proceedings of the ACM India Joint International
  Conference on Data Science and Management of Data}, 2019, pp. 362--365.

\bibitem{hoang2019collective}
T.~N. Hoang, Q.~M. Hoang, K.~H. Low, and J.~How, ``Collective online learning
  of gaussian processes in massive multi-agent systems,'' in \emph{Proceedings
  of the AAAI Conference on Artificial Intelligence}, vol.~33, 2019, pp.
  7850--7857.

\bibitem{majumdar2014advanced}
A.~K. Majumdar, \emph{Advanced Free Space Optics ({FSO}): A Systems
  Approach}.\hskip 1em plus 0.5em minus 0.4em\relax Springer, 2014, vol. 186.

\bibitem{everaerts2008use}
J.~Everaerts \emph{et~al.}, ``The use of unmanned aerial vehicles {(UAVs)} for
  remote sensing and mapping,'' \emph{The International Archives of the
  Photogrammetry, Remote Sensing and Spatial Information Sciences}, vol.~37,
  no. 2008, pp. 1187--1192, 2008.

\bibitem{guerrero2013quad}
J.~A. Guerrero, J.-A. Escare{\~n}o, and Y.~Bestaoui, ``Quad-rotor {MAV}
  trajectory planning in wind fields,'' in \emph{2013 IEEE International
  Conference on Robotics and Automation}.\hskip 1em plus 0.5em minus
  0.4em\relax IEEE, 2013, pp. 778--783.

\bibitem{al2013wind}
W.~H. Al-Sabban, L.~F. Gonzalez, and R.~N. Smith, ``Wind-energy based path
  planning for unmanned aerial vehicles using markov decision processes,'' in
  \emph{2013 IEEE International Conference on Robotics and Automation}.\hskip
  1em plus 0.5em minus 0.4em\relax IEEE, 2013, pp. 784--789.

\bibitem{misra2006global}
P.~Misra and P.~Enge, ``Global positioning system: signals, measurements and
  performance second edition,'' \emph{Massachusetts: Ganga-Jamuna Press}, 2006.

\end{thebibliography}

\newpage	
\onecolumn

\section*{Appendix A. Derivation of the Evidence Lower Bound (ELBO) in Section~IV}
The full joint distribution of the GP-SSM is given by:
\begin{equation}
\begin{split}
p(\boldsymbol{y}_{0:T\!-\!1},\boldsymbol{x}_{0:T},\boldsymbol{f}_{1:T}, \boldsymbol{z} \mid \vartheta) =
p(\boldsymbol{x}_{1:T}, \boldsymbol{f}_{1:T} \mid \boldsymbol{z},\boldsymbol{x}_{0},\vartheta)p(\boldsymbol{z} \mid \theta)p(\boldsymbol{x}_{0})
\bigg[\prod_{t=0}^{T\!-\!1} p(\boldsymbol{y}_{t} \mid \boldsymbol{x}_{t})\bigg].
\label{eq:appen1}
\end{split}
\end{equation}
With the approximate distribution for the variational posterior defined in (\ref{eq:variationdis}), the variational distribution is defined by:
\begin{equation}
q(\boldsymbol{x}_{0:T},\boldsymbol{f}_{1:T}, \boldsymbol{z} \mid \vartheta) =
p(\boldsymbol{x}_{1:T}, \boldsymbol{f}_{1:T} \mid \boldsymbol{z},\boldsymbol{x}_{0}, \vartheta)q(\boldsymbol{z}) q(\boldsymbol{x}_{0}).\label{eq:appen2}
\end{equation}
Based on equations (\ref{eq:appen1}) and (\ref{eq:appen2}), the ELBO is derived by:
\begin{equation}
\begin{split}
&\log p(\boldsymbol{y}_{0:T\!-\!1} \mid \vartheta) \geq \mathbb{E}_{q(\boldsymbol{x}_{0:T},\boldsymbol{f}_{1:T},\boldsymbol{z} \mid \vartheta)}
\bigg[\log \dfrac{p(\boldsymbol{y}_{0:T\!-\!1},\boldsymbol{x}_{0:T},\boldsymbol{f}_{1:T}, \boldsymbol{z} \mid \vartheta)}{q(\boldsymbol{x}_{0:T},\boldsymbol{f}_{1:T}, \boldsymbol{z} \mid \vartheta)}  \bigg] \\
&= \mathbb{E}_{q(\boldsymbol{x}_{0:T},\boldsymbol{f}_{1:T}, \boldsymbol{z} \mid \vartheta)}
\begin{bmatrix} \log
\dfrac{p(\boldsymbol{x}_{0}) p(\boldsymbol{z} \mid \theta) \bcancel{p(\boldsymbol{x}_{1:T}, \boldsymbol{f}_{1:T} \mid \boldsymbol{z},\boldsymbol{x}_{0}, \vartheta)}
	\bigg[\prod_{t=0}^{T\!-\!1} p(\boldsymbol{y}_{t} \mid \boldsymbol{x}_{t})\bigg]}
{q(\boldsymbol{x}_{0}) q(\boldsymbol{z}) \bcancel{p(\boldsymbol{x}_{1:T}, \boldsymbol{f}_{1:T} \mid \boldsymbol{z},\boldsymbol{x}_{0}, \vartheta)}
}
\end{bmatrix}\\
&= \mathbb{E}_{q(\boldsymbol{x}_{0:T},\boldsymbol{f}_{1:T}, \boldsymbol{z} \mid \vartheta)} \bigg[\log \prod_{t=0}^{T\!-\!1} p(\boldsymbol{y}_{t} \mid \boldsymbol{x}_{t})\bigg]
- \mathbb{E}_{q(\boldsymbol{x}_{0:T},\boldsymbol{f}_{1:T}, \boldsymbol{z} \mid \vartheta)} \bigg[\log \frac{q(\boldsymbol{x}_{0})}{p(\boldsymbol{x}_{0})} \bigg]
- \mathbb{E}_{q(\boldsymbol{x}_{0:T},\boldsymbol{f}_{1:T}, \boldsymbol{z} \mid \vartheta)} \bigg[\log \frac{q(\boldsymbol{z})}{p(\boldsymbol{z} \mid \theta)} \bigg]\\
&= \mathbb{E}_{q(\boldsymbol{x}_{0:T})} \bigg[\log \prod_{t=0}^{T\!-\!1} p(\boldsymbol{y}_{t} \mid \boldsymbol{x}_{t})\bigg]
- \mathbb{E}_{q(\boldsymbol{x}_{0})} \bigg[\log \frac{q(\boldsymbol{x}_{0})}{p(\boldsymbol{x}_{0})} \bigg]
- \mathbb{E}_{q(\boldsymbol{z})} \bigg[\log \frac{q(\boldsymbol{z})}{p(\boldsymbol{z} \mid \theta)} \bigg]\\
&= \sum_{t=0}^{T\!-\!1} \mathbb{E}_{q(\boldsymbol{x}_{t})} \big[\log p(\boldsymbol{y}_{t} \mid \boldsymbol{x}_{t}) \big]
-\mathcal{D}_{KL}(q(\boldsymbol{x}_{0}) \lVert p(\boldsymbol{x}_{0}))
-\mathcal{D}_{KL}(q(\boldsymbol{z}) \lVert p(\boldsymbol{z} \mid \theta)).
\end{split}
\label{eq:appen_ELBO}
\end{equation}
For a more detailed explanation, refer to \cite{salimbeni2017doubly,doerr2018PRSSM} and
the references therein.	
	
\section*{Appendix B. Derivation of the negative ELBO in Section~V}
The new approximate posterior $q(\boldsymbol{x}_{0:T'},\boldsymbol{f}_{1:T'},\boldsymbol{z},\boldsymbol{z'} \!\mid\! \vartheta') = 
p(\boldsymbol{x}_{0:T\!-\!1},\boldsymbol{f}_{1:T} ,\boldsymbol{z} \!\mid\! \boldsymbol{z'}, \vartheta')
p(\boldsymbol{x}_{T\!+\!1:T'},\boldsymbol{f}_{T\!+\!1:T'} \!\mid\! \boldsymbol{x}_{0:T}, \boldsymbol{f}_{1:T}, \boldsymbol{z}, \boldsymbol{z'},\vartheta') q(\boldsymbol{z'}) q(\boldsymbol{x}_{T})$
is optimized by minimizing the KL divergence:
\begin{equation}
\begin{split}
&\mathcal{D}_{KL}(q(\boldsymbol{x}_{0:T'},\boldsymbol{f}_{1:T'},\boldsymbol{z},\boldsymbol{z'} \mid \vartheta') \lVert
\hat{p}(\boldsymbol{x}_{0:T'},\boldsymbol{f}_{1:T'}, \boldsymbol{z}, \boldsymbol{z'} \mid \boldsymbol{y}_{0:T'},\vartheta'))\\
&= \log \dfrac{p(\boldsymbol{y}_{0:T'\!-\!1} \mid \vartheta')}{p(\boldsymbol{y}_{0:T\!-\!1} \mid \vartheta)}
+ \mathbb{E}_{q(\boldsymbol{x}_{0:T'},\boldsymbol{f}_{1:T'},\boldsymbol{z},\boldsymbol{z'} \mid \vartheta')}
\left[ \log \dfrac{p(\boldsymbol{x}_{0:T},\boldsymbol{f}_{1:T}, \boldsymbol{z}_0, \boldsymbol{z} \mid \vartheta)}
{q(\boldsymbol{x}_{0:T},\boldsymbol{f}_{1:T}, \boldsymbol{z}_0, \boldsymbol{z} \mid \vartheta)}
\right]\\
&+ \mathbb{E}_{q(\boldsymbol{x}_{0:T'},\boldsymbol{f}_{1:T'},\boldsymbol{z},\boldsymbol{z'} \mid \vartheta')}
\left[
\log \dfrac{q(\boldsymbol{x}_{0:T'},\boldsymbol{f}_{1:T'},\boldsymbol{z},\boldsymbol{z'} \mid \vartheta')}
{p(\boldsymbol{x}_{0:T\!-\!1}, \boldsymbol{f}_{1:T}, \boldsymbol{z}, \boldsymbol{z'} \!\mid\! \vartheta') \hat{p}(\boldsymbol{x}_{T})p(\boldsymbol{x}_{T\!+\!1:T'}, \boldsymbol{f}_{T\!+\!1:T'} \mid \boldsymbol{x}_{0:T}, \boldsymbol{f}_{1:T}, \boldsymbol{z}, \boldsymbol{z'},\vartheta')
	p(\boldsymbol{y}_{T:T'\!-\!1} \mid \boldsymbol{x}_{T:T'\!-\!1})}
\right].
\end{split} \label{eq:appen_KL}
\end{equation}
Since the KL divergence is non-negative, the expectation terms in (\ref{eq:appen_KL}) are the negative approximate lower bound of the log marginal likelihood.
Thus, the negative evidence lower bound (negative ELBO) of the approximate online log marginal likelihood is represented by: 	
\begin{equation}
\begin{split}
&\mathcal{NL}(\vartheta')
= \mathbb{E}_{q(\boldsymbol{x}_{0:T'},\boldsymbol{f}_{1:T'},\boldsymbol{z},\boldsymbol{z'} \mid \vartheta')}
\left[ \log \dfrac{p(\boldsymbol{x}_{0:T},\boldsymbol{f}_{1:T}, \boldsymbol{z}_0, \boldsymbol{z} \mid \vartheta)}
{q(\boldsymbol{x}_{0:T},\boldsymbol{f}_{1:T}, \boldsymbol{z}_0, \boldsymbol{z} \mid \vartheta)}
\right]\\
&+ \mathbb{E}_{q(\boldsymbol{x}_{0:T'},\boldsymbol{f}_{1:T'},\boldsymbol{z},\boldsymbol{z'} \mid \vartheta')}
\left[
\log \dfrac{q(\boldsymbol{x}_{0:T'},\boldsymbol{f}_{1:T'},\boldsymbol{z},\boldsymbol{z'} \mid \vartheta')}
{p(\boldsymbol{x}_{0:T\!-\!1}, \boldsymbol{f}_{1:T}, \boldsymbol{z}, \boldsymbol{z'}  \!\mid\! \vartheta') \hat{p}(\boldsymbol{x}_{T})p(\boldsymbol{x}_{T\!+\!1:T'}, \boldsymbol{f}_{T\!+\!1:T'} \mid \boldsymbol{x}_{0:T}, \boldsymbol{f}_{1:T}, \boldsymbol{z}, \boldsymbol{z'},\vartheta')
	p(\boldsymbol{y}_{T:T'\!-\!1} \mid \boldsymbol{x}_{T:T'\!-\!1})}
\right]
.
\end{split}
\label{eq:appen_NELBO}
\end{equation}
{The second term of (\ref{eq:appen_NELBO}) is obtained in the same way as in (\ref{eq:appen_ELBO}):
	\begin{equation}
	\begin{split}
	&\mathbb{E}_{q(\boldsymbol{x}_{0:T'},\boldsymbol{f}_{1:T'},\boldsymbol{z},\boldsymbol{z'} \mid \vartheta')}
	\left[
	\log \dfrac{q(\boldsymbol{x}_{0:T'},\boldsymbol{f}_{1:T'},\boldsymbol{z},\boldsymbol{z'} \mid \vartheta')}
	{p(\boldsymbol{x}_{0:T\!-\!1}, \boldsymbol{f}_{1:T}, \boldsymbol{z}, \boldsymbol{z'}  \!\mid\! \vartheta') \hat{p}(\boldsymbol{x}_{T})p(\boldsymbol{x}_{T\!+\!1:T'}, \boldsymbol{f}_{T\!+\!1:T'} \mid \boldsymbol{x}_{0:T}, \boldsymbol{f}_{1:T}, \boldsymbol{z}, \boldsymbol{z'},\vartheta')
		p(\boldsymbol{y}_{T:T'\!-\!1} \mid \boldsymbol{x}_{T:T'\!-\!1})}
	\right]\\
	&= \mathbb{E}_{q(\cdot \mid \vartheta')}
	\begin{bmatrix} \log
	\dfrac{q(\boldsymbol{x}_{T}) q(\boldsymbol{z'}) \bcancel{p(\boldsymbol{x}_{0:T\!-\!1}, \boldsymbol{f}_{1:T} , \boldsymbol{z} \!\mid\! \boldsymbol{z'}, \vartheta')}\bcancel{p(\boldsymbol{x}_{T\!+\!1:T'}, \boldsymbol{f}_{T\!+\!1:T'} \mid \boldsymbol{x}_{0:T}, \boldsymbol{f}_{1:T}, \boldsymbol{z}, \boldsymbol{z'},\vartheta')}}
	{\hat{p}(\boldsymbol{x}_{T}) p(\boldsymbol{z'}\mid \theta') \bcancel{p(\boldsymbol{x}_{0:T\!-\!1}, \boldsymbol{f}_{1:T}, \boldsymbol{z} \!\mid\! \boldsymbol{z'}, \vartheta')}\bcancel{p(\boldsymbol{x}_{T\!+\!1:T'}, \boldsymbol{f}_{T\!+\!1:T'} \mid \boldsymbol{x}_{0:T}, \boldsymbol{f}_{1:T}, \boldsymbol{z}, \boldsymbol{z'},\vartheta')}
		\bigg[\prod_{t=T}^{T'\!-\!1} p(\boldsymbol{y}_{t} \mid \boldsymbol{x}_{t})\bigg]}
	\end{bmatrix}\\
	&= - \sum_{t=T}^{T'\!-\!1} \mathbb{E}_{q(\boldsymbol{x}_{t})} \big[\log p(\boldsymbol{y}_{t} \mid \boldsymbol{x}_{t}) \big]
	+\mathcal{D}_{KL}(q(\boldsymbol{x}_{T}) \lVert \hat{p}(\boldsymbol{x}_{T}))
	+\mathcal{D}_{KL}(q(\boldsymbol{z'}) \lVert p(\boldsymbol{z'}\mid \theta')).
	\end{split}
	\end{equation}
}
From equations (\ref{eq:GP-SSM}) and (\ref{eq:appen2}), $p(\boldsymbol{x}_{0:T},\boldsymbol{f}_{1:T},\boldsymbol{z}_{0},\boldsymbol{z} \mid \vartheta)/q(\boldsymbol{x}_{0:T},\boldsymbol{f}_{1:T},\boldsymbol{z}_{0},\boldsymbol{z} \mid \vartheta)$ in the first term of (\ref{eq:appen_NELBO}) is given by:
\begin{equation}
\begin{split}
\dfrac{p(\boldsymbol{x}_{0:T},\boldsymbol{f}_{1:T},\boldsymbol{z}_{0},\boldsymbol{z} \mid \vartheta)}
{q(\boldsymbol{x}_{0:T},\boldsymbol{f}_{1:T},\boldsymbol{z}_{0},\boldsymbol{z} \mid \vartheta)}
&= \dfrac{\tilde{p}(\boldsymbol{x}_{T_0})p(\boldsymbol{z} \mid \theta)
	\bcancel{p(\boldsymbol{x}_{0:T_0\!-\!1},\boldsymbol{f}_{1:T_0}, \boldsymbol{z}_0 \!\mid\! \boldsymbol{z}, \vartheta)}
	\bcancel{p(\boldsymbol{x}_{T_0\!+\!1:T},\boldsymbol{f}_{T_0\!+\!1:T} \!\mid\! \boldsymbol{x}_{0:T_0},\boldsymbol{f}_{1:T_0},\boldsymbol{z}_0,\boldsymbol{z}, \vartheta)}
}
{q(\boldsymbol{x}_{T_0}) q(\boldsymbol{z})
	\bcancel{p(\boldsymbol{x}_{0:T_0\!-\!1},\boldsymbol{f}_{1:T_0}, \boldsymbol{z}_0 \!\mid\! \boldsymbol{z}, \vartheta)}
	\bcancel{p(\boldsymbol{x}_{T_0\!+\!1:T},\boldsymbol{f}_{T_0\!+\!1:T} \!\mid\! \boldsymbol{x}_{0:T_0},\boldsymbol{f}_{1:T_0},\boldsymbol{z}_0,\boldsymbol{z}, \vartheta)}
}\\
&= \dfrac{\tilde{p}(\boldsymbol{x}_{T_0}) p(\boldsymbol{z} \mid \theta)}{q(\boldsymbol{x}_{T_0}) q(\boldsymbol{z})},
\end{split}
\end{equation}
where $\tilde{p}(\boldsymbol{x}_{T_0}) = p(\boldsymbol{x}_{T_0} \mid \boldsymbol{x}_{0:T_0\!-\!1},\boldsymbol{f}_{1:T_0},\boldsymbol{z}_0,\boldsymbol{z}, \vartheta)$.
Substituting this, the first term of (\ref{eq:appen_NELBO}) is given by:
\begin{equation}
\begin{split}
\mathbb{E}_{q(\boldsymbol{x}_{0:T'},\boldsymbol{f}_{1:T'},\boldsymbol{z},\boldsymbol{z'} \mid \vartheta')}
\left[\log \frac{p(\boldsymbol{x}_{0:T},\boldsymbol{f}_{1:T},\boldsymbol{z}_0,\boldsymbol{z} \mid \vartheta)}{q(\boldsymbol{x}_{0:T},\boldsymbol{f}_{1:T},\boldsymbol{z}_0,\boldsymbol{z} \mid \vartheta)} \right]
&=\mathbb{E}_{q(\boldsymbol{x}_{0:T'},\boldsymbol{f}_{1:T'},\boldsymbol{z},\boldsymbol{z'} \mid \vartheta')}
\left[\log\dfrac{\tilde{p}(\boldsymbol{x}_{T_0}) p(\boldsymbol{z} \mid \theta)}{q(\boldsymbol{x}_{T_0}) q(\boldsymbol{z})}\right]\\
&=\mathbb{E}_{q(\boldsymbol{x}_{0:T'},\boldsymbol{f}_{1:T'},\boldsymbol{z},\boldsymbol{z'} \mid \vartheta')}
\left[\log\dfrac{\tilde{p}(\boldsymbol{x}_{T_0})}{q(\boldsymbol{x}_{T_0})} \dfrac{q'(\boldsymbol{z})}{q(\boldsymbol{z})} \dfrac{p(\boldsymbol{z} \mid \theta)}{q'(\boldsymbol{z})}\right]\\
&=
\mathbb{E}_{\tilde{p}(\boldsymbol{x}_{T_0})} \left[\log \dfrac{\tilde{p}(\boldsymbol{x}_{T_0})}{q(\boldsymbol{x}_{T_0})}\right]+
\mathbb{E}_{q'(\boldsymbol{z})} \left[\log \dfrac{q'(\boldsymbol{z})}{q(\boldsymbol{z})}\right]
-\mathbb{E}_{q'(\boldsymbol{z})} \left[\log \dfrac{q'(\boldsymbol{z})}{p(\boldsymbol{z} \mid \theta)}\right]\\
&=
\mathcal{D}_{KL}(\tilde{p}(\boldsymbol{x}_{T_0}) \lVert q(\boldsymbol{x}_{T_0}))
+\mathcal{D}_{KL}(q'(\boldsymbol{z}) \lVert q(\boldsymbol{z}))
-\mathcal{D}_{KL}(q'(\boldsymbol{z}) \lVert p(\boldsymbol{z} \mid \theta)),
\end{split} \label{eq:append_online_term}
\end{equation}
where
\begin{equation}
\begin{split}
q'(\boldsymbol{z})
&= \int\int\int q(\boldsymbol{x}_{0:T'},\boldsymbol{f}_{1:T'},\boldsymbol{z},\boldsymbol{z'} \mid \vartheta')
d\boldsymbol{x}_{0:T'}d\boldsymbol{f}_{1:T'}d\boldsymbol{z'}\\
&= \int\int\int q(\boldsymbol{x}_{T}) q(\boldsymbol{z'}) 
p(\boldsymbol{x}_{0:T\!-\!1},\boldsymbol{f}_{1:T} ,\boldsymbol{z} \!\mid\! \boldsymbol{z'}, \vartheta')
p(\boldsymbol{x}_{T\!+\!1:T'},\boldsymbol{f}_{T\!+\!1:T'} \!\mid\! \boldsymbol{x}_{0:T}, \boldsymbol{f}_{1:T}, \boldsymbol{z}, \boldsymbol{z'},\vartheta')
d\boldsymbol{x}_{0:T'}d\boldsymbol{f}_{1:T'}d\boldsymbol{z'}\\
&= \int p(\boldsymbol{z} \mid \boldsymbol{z'},\theta')q(\boldsymbol{z'}) d\boldsymbol{z'} \\
&= \int \mathcal{MN}(\boldsymbol{z};\boldsymbol{m}_{\boldsymbol{z} \mid \boldsymbol{z'}}, \boldsymbol{S}_{\boldsymbol{z} \mid \boldsymbol{z'}}, \boldsymbol{V'})
\mathcal{MN}(\boldsymbol{z'};\boldsymbol{\mu}_{\boldsymbol{z'}}, \boldsymbol{\Sigma}_{\boldsymbol{z'}}, \boldsymbol{V'}) d\boldsymbol{z'} \\
&= \mathcal{MN}(\boldsymbol{z} ; \boldsymbol{m}_{\boldsymbol{\eta}} +
\boldsymbol{K}_{\boldsymbol{\eta} \boldsymbol{\eta}'} \boldsymbol{K}_{\boldsymbol{\eta}' \boldsymbol{\eta}'}^{-1}
\big[\boldsymbol{\mu}_{\boldsymbol{z'}}-\boldsymbol{m}_{\boldsymbol{\eta}'}\big],
\boldsymbol{S}_{\boldsymbol{z} \mid \boldsymbol{z'}}+
\boldsymbol{K}_{\boldsymbol{\eta} \boldsymbol{\eta}'}
\boldsymbol{K}_{\boldsymbol{\eta}'\boldsymbol{\eta}'}^{-1}
\boldsymbol{K}_{\boldsymbol{\eta}'\boldsymbol{\eta}}, \boldsymbol{V'}),\\
p(\boldsymbol{z} \mid \boldsymbol{z'},\theta') &= \mathcal{MN}(\boldsymbol{z};\boldsymbol{m}_{\boldsymbol{z} \mid \boldsymbol{z'}}, \boldsymbol{S}_{\boldsymbol{z} \mid \boldsymbol{z'}}, \boldsymbol{V'}) \\
&= \mathcal{MN}(\boldsymbol{z};\boldsymbol{m}_{\boldsymbol{\eta}} +
\boldsymbol{K}_{\boldsymbol{\eta}\boldsymbol{\eta}'} \boldsymbol{K}_{\boldsymbol{\eta}'\boldsymbol{\eta}'}^{-1} \big[\boldsymbol{z'}-\boldsymbol{m}_{\boldsymbol{\eta}'}\big], \boldsymbol{K}_{\boldsymbol{\eta}\boldsymbol{\eta}} - \boldsymbol{K}_{\boldsymbol{\eta}\boldsymbol{\eta}'} \boldsymbol{K}_{\boldsymbol{\eta}'\boldsymbol{\eta}'}^{-1} \boldsymbol{K}_{\boldsymbol{\eta}'\boldsymbol{\eta}}, \boldsymbol{V'}),\\
q(\boldsymbol{z'}) &= \mathcal{MN}(\boldsymbol{z'};\boldsymbol{\mu}_{\boldsymbol{z'}}, \boldsymbol{\Sigma}_{\boldsymbol{z'}}, \boldsymbol{V'}). \\
\end{split}
\end{equation}
The first term in (\ref{eq:append_online_term}) can be neglected if $\tilde{p}(\boldsymbol{x}_{T_0})\approx q(\boldsymbol{x}_{T_0})$. Thus, 
Finally, the negative evidence lower bound (negative ELBO) of the approximate online log marginal likelihood is represented by:
\begin{align}
\begin{split}
\mathcal{NL}(\vartheta') =
&- \sum_{t=T}^{T'\!-\!1} \mathbb{E}_{q({\boldsymbol{x}_{t}})} \big[\log p(\boldsymbol{y}_{t} \mid \boldsymbol{x}_{t})\big]
+ \mathcal{D}_{KL}(q(\boldsymbol{x}_{T}) \lVert \hat{p}(\boldsymbol{x}_{T}))
+ \mathcal{D}_{KL}(q(\boldsymbol{z'}) \lVert p(\boldsymbol{z'} \mid \theta'))\\		
&+ \mathcal{D}_{KL}(q'(\boldsymbol{z}) \lVert q(\boldsymbol{z}))
-  \mathcal{D}_{KL}(q'(\boldsymbol{z}) \lVert p(\boldsymbol{z} \mid \theta)).
\end{split}
\end{align}

\newpage

\section*{Appendix C. Additional Simulation Results in Section~VII-A}

\begin{figure}[ht]
	\centering
	\subfigure[Step 1]{
		\includegraphics[trim={0.1cm 0.2cm 1cm 0.2cm},clip,width=0.25\columnwidth]{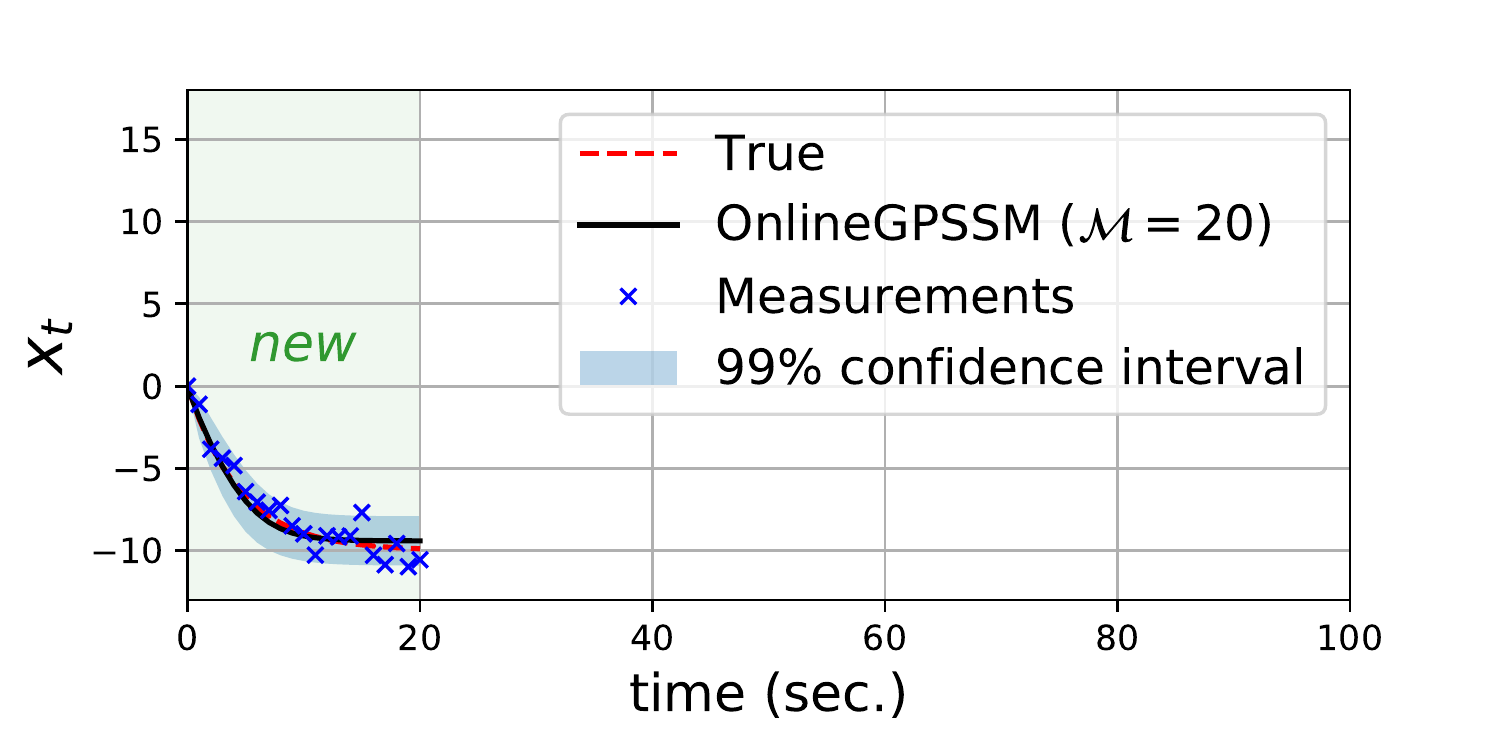}}
	\hspace*{-0.4cm}
	\subfigure[Step 1]{
		\includegraphics[trim={0.1cm 0.2cm 1cm 0.2cm},clip,width=0.25\columnwidth]{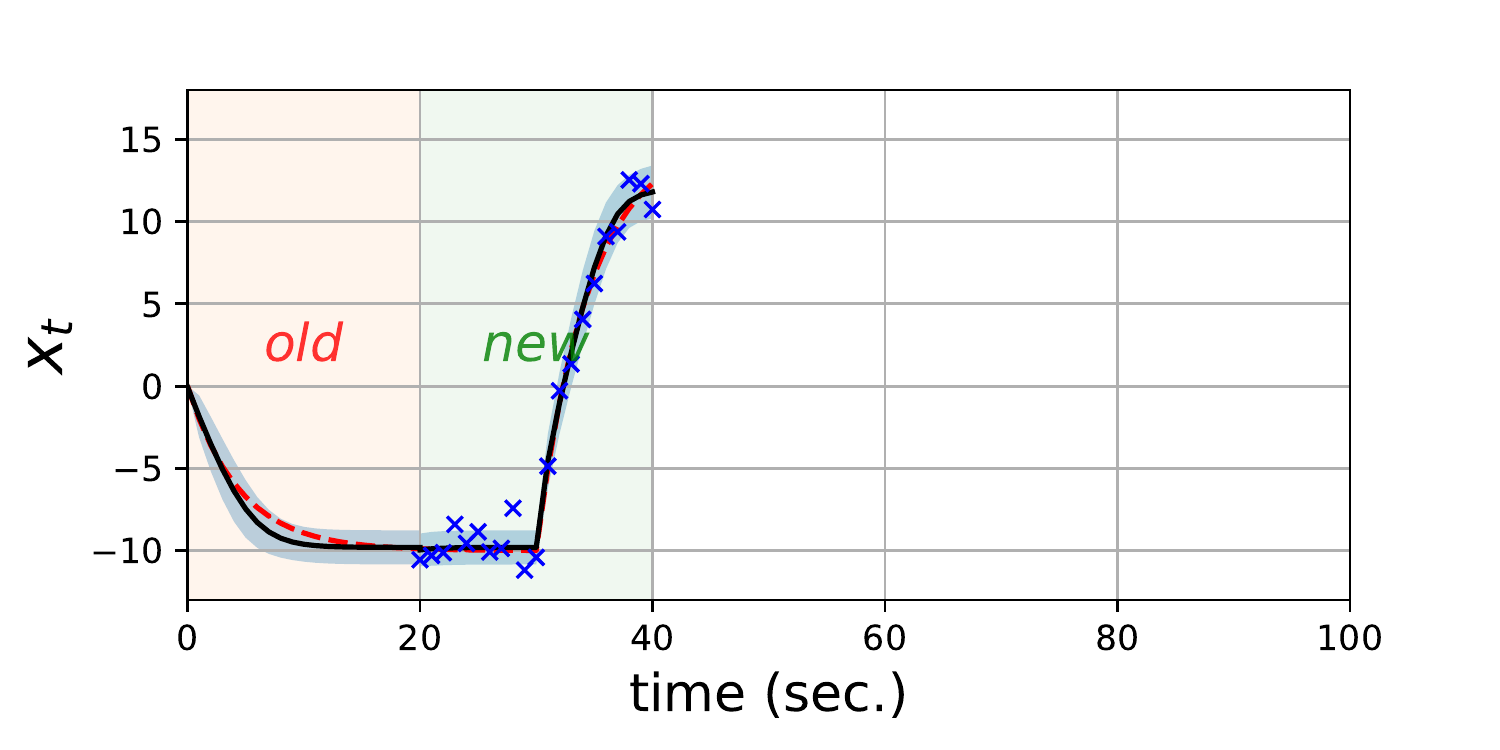}}	
	\hspace*{-0.4cm}
	\subfigure[Step 2]{
		\includegraphics[trim={0.1cm 0.2cm 1cm 0.2cm},clip,width=0.25\columnwidth]{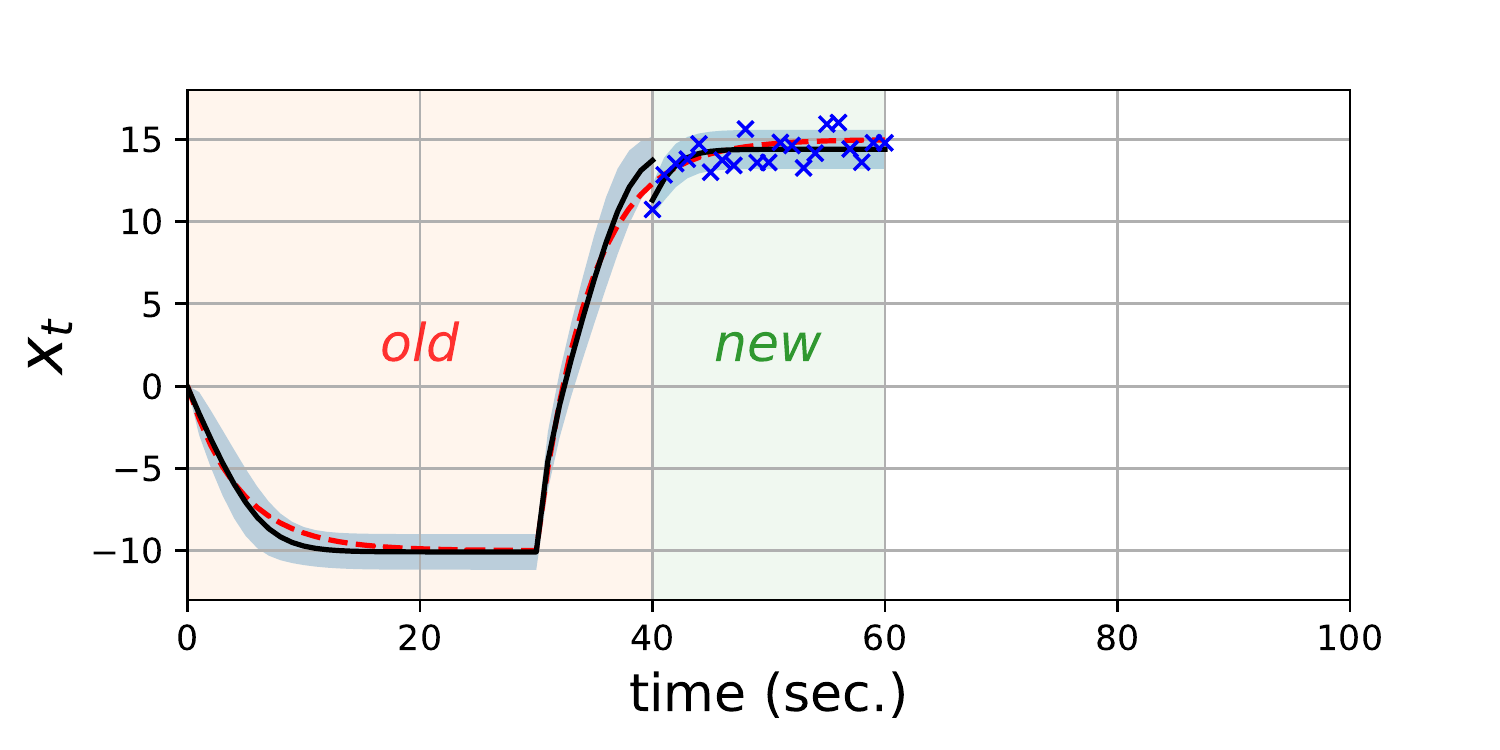}}
	\hspace*{-0.4cm}
	\subfigure[Step 2]{
		\includegraphics[trim={0.1cm 0.2cm 1cm 0.2cm},clip,width=0.25\columnwidth]{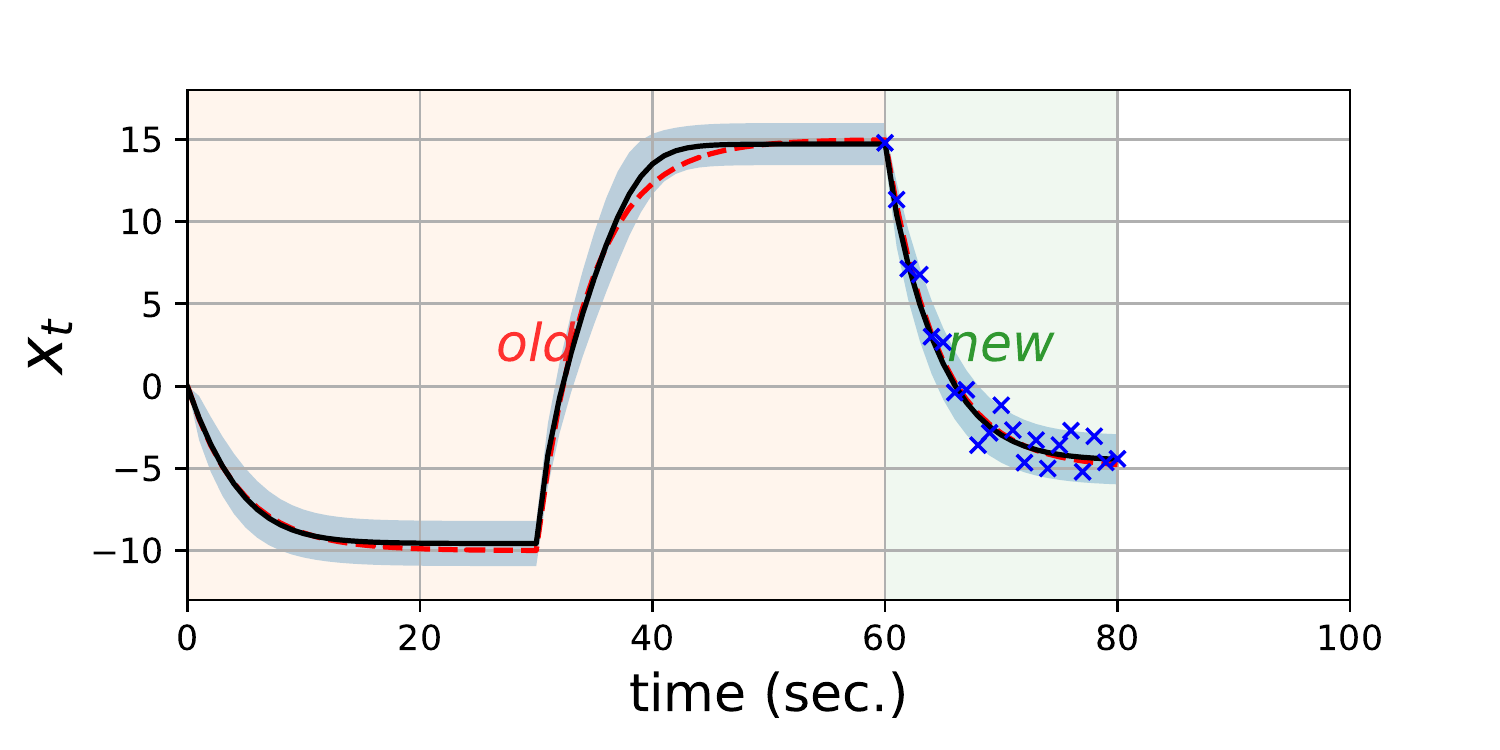}}\\
	\subfigure[Step 3]{
		\includegraphics[trim={0.1cm 0.2cm 1cm 0.2cm},clip,width=0.25\columnwidth]{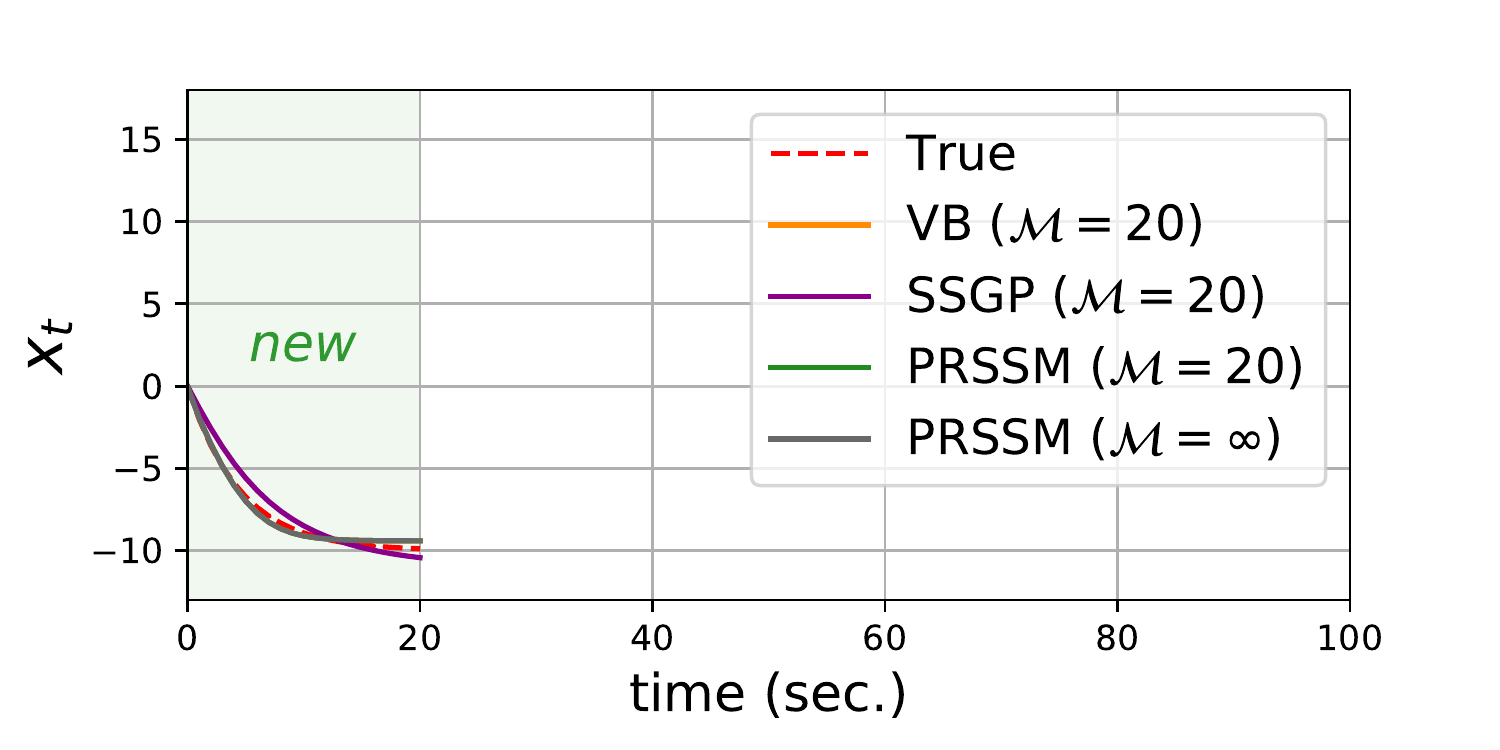}}
	\hspace*{-0.4cm}
	\subfigure[Step 3]{
		\includegraphics[trim={0.1cm 0.2cm 1cm 0.2cm},clip,width=0.25\columnwidth]{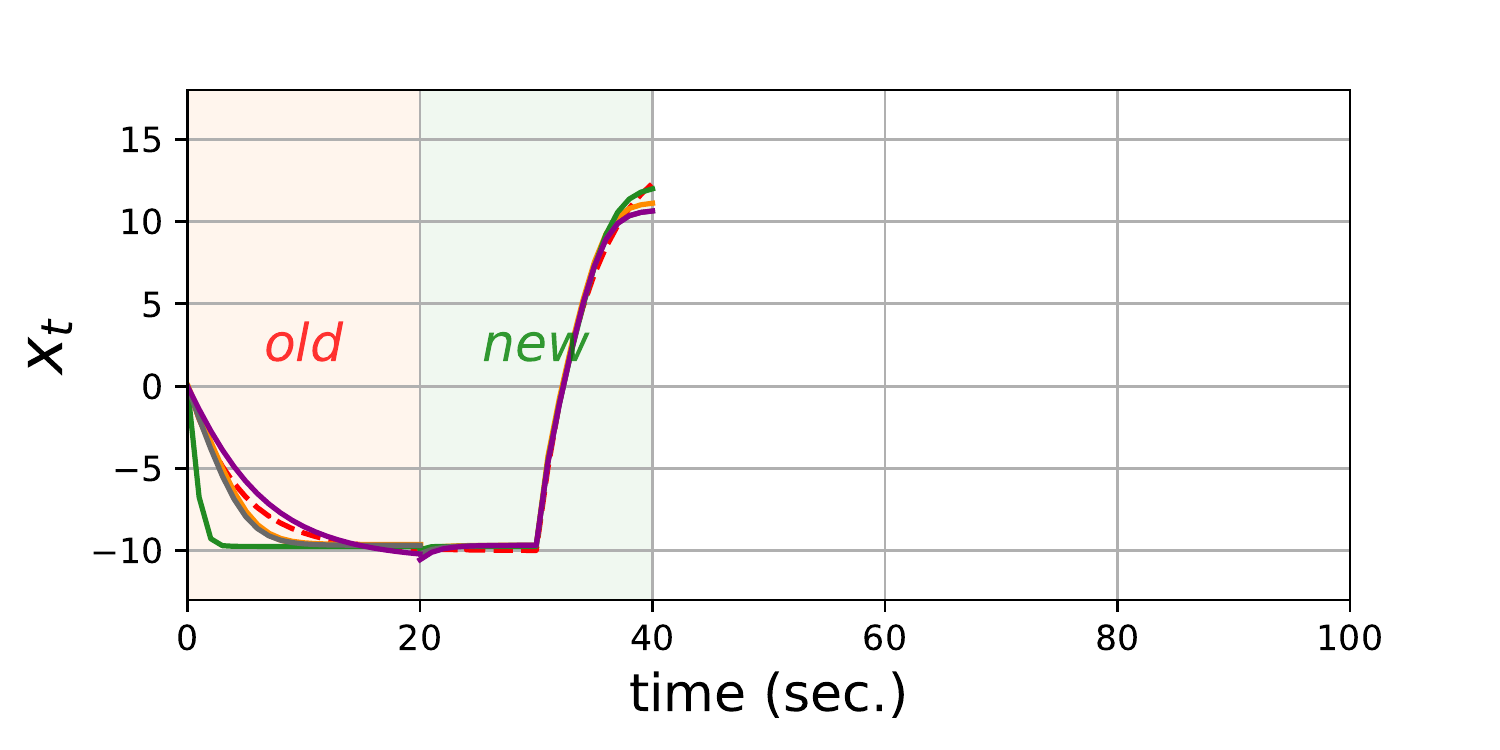}}
	\hspace*{-0.4cm}
	\subfigure[Step 4]{
		\includegraphics[trim={0.1cm 0.2cm 1cm 0.2cm},clip,width=0.25\columnwidth]{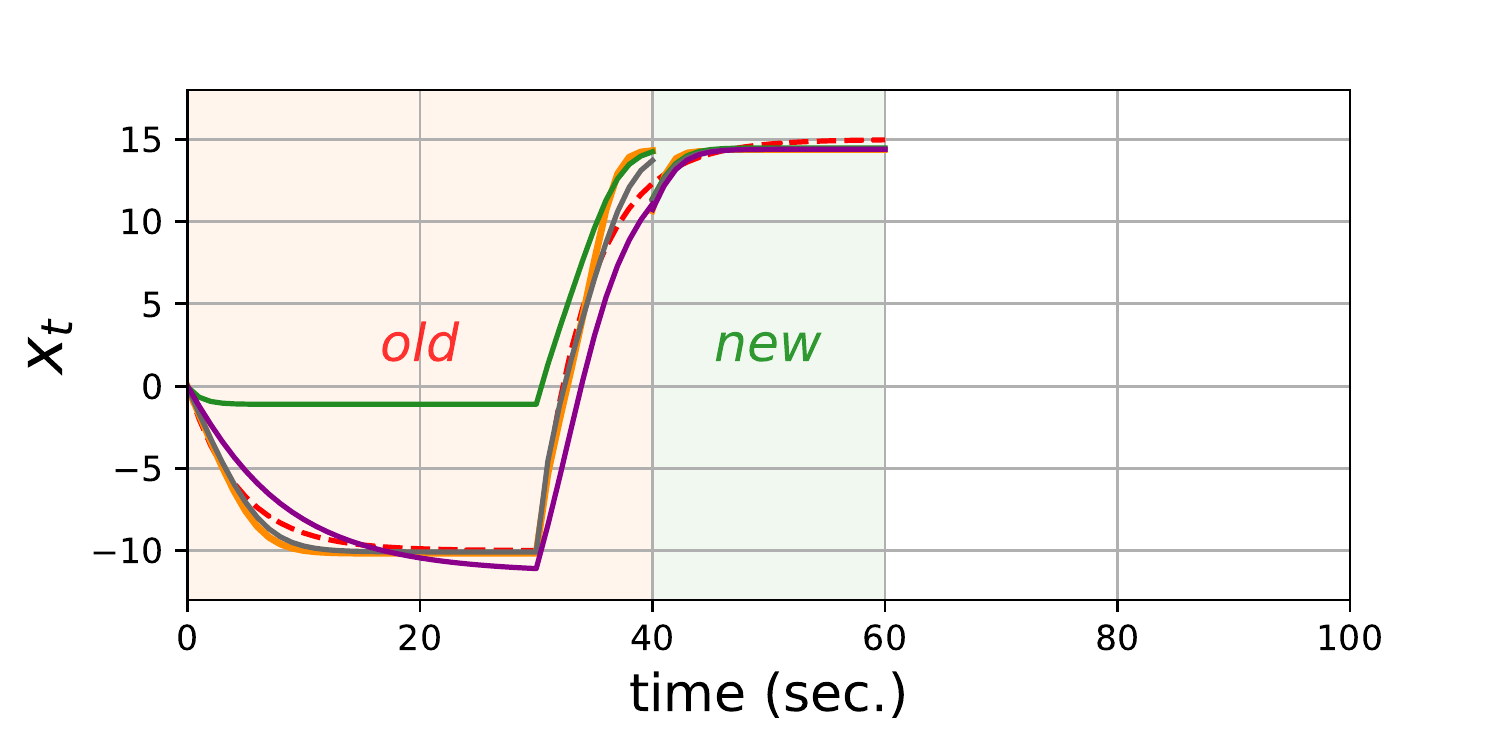}}
	\hspace*{-0.4cm}
	\subfigure[Step 4]{
		\includegraphics[trim={0.1cm 0.2cm 1cm 0.2cm},clip,width=0.25\columnwidth]{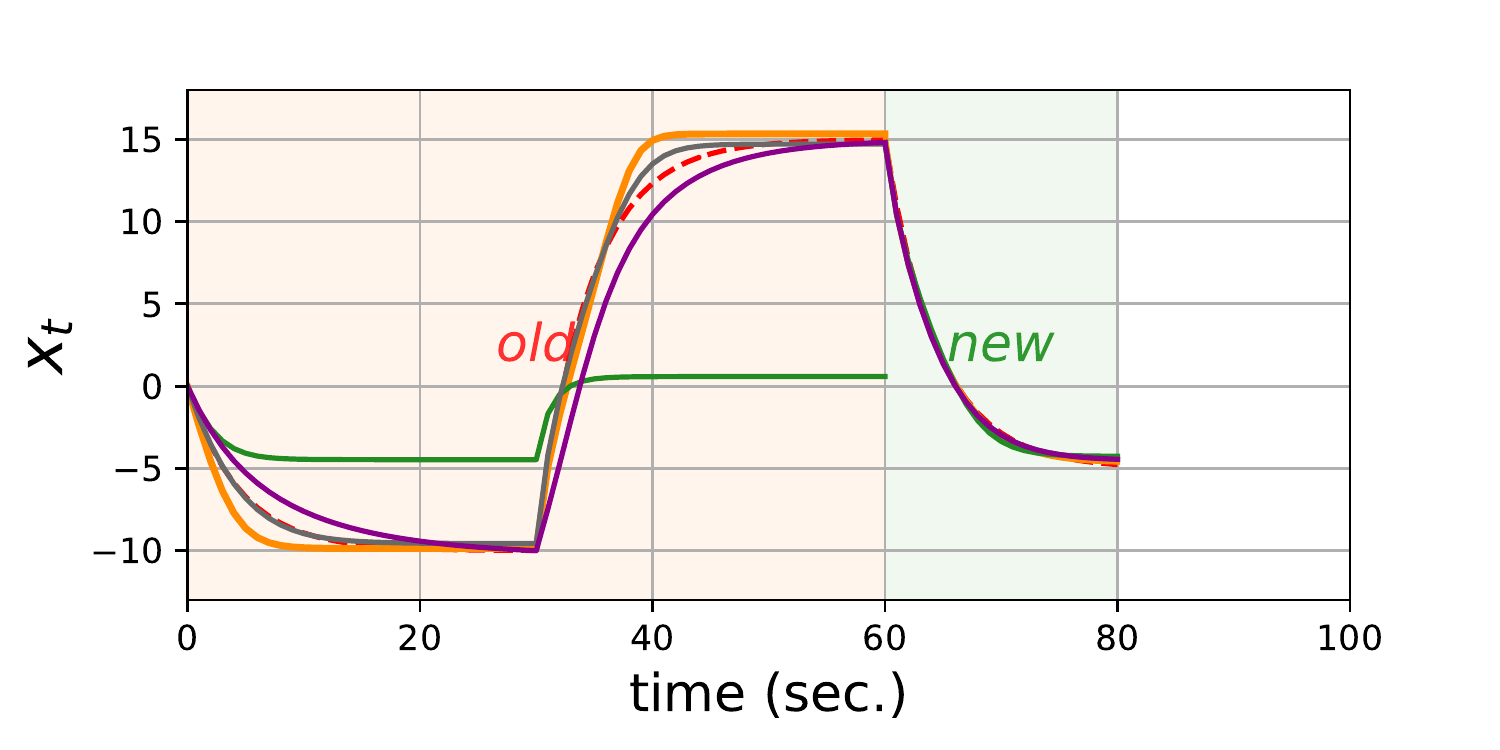}}	
	\caption{Reconstruction results after each online learning.}
\end{figure}

\begin{figure}[ht]
	\centering
	\subfigure[Step 5]{
		\includegraphics[trim={.5cm 1.cm 1cm 1.5cm},clip,width=0.30\columnwidth]{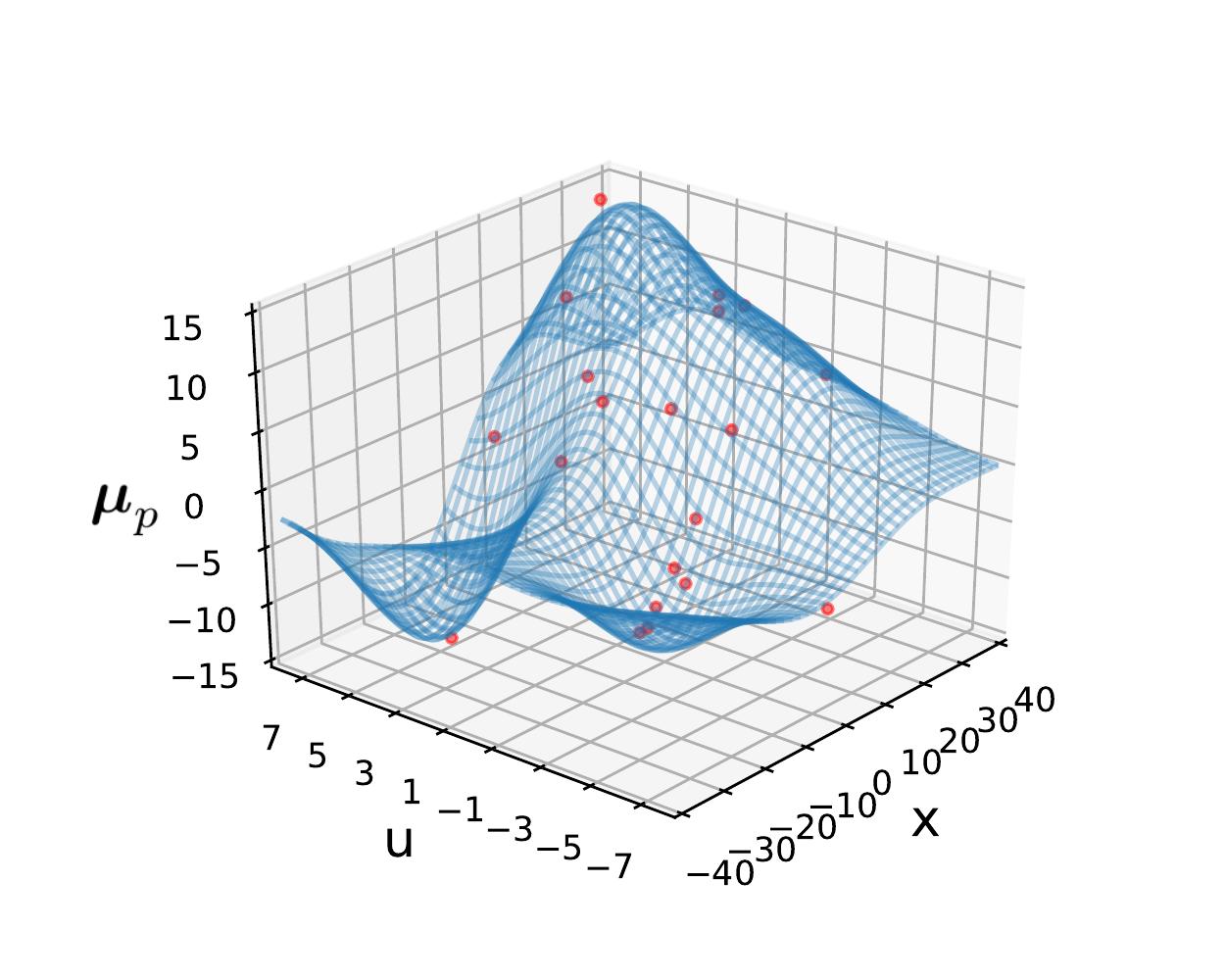}}
	\subfigure[Step 5]{
		\includegraphics[trim={.5cm 1.cm 1cm 1.5cm},clip,width=0.30\columnwidth]{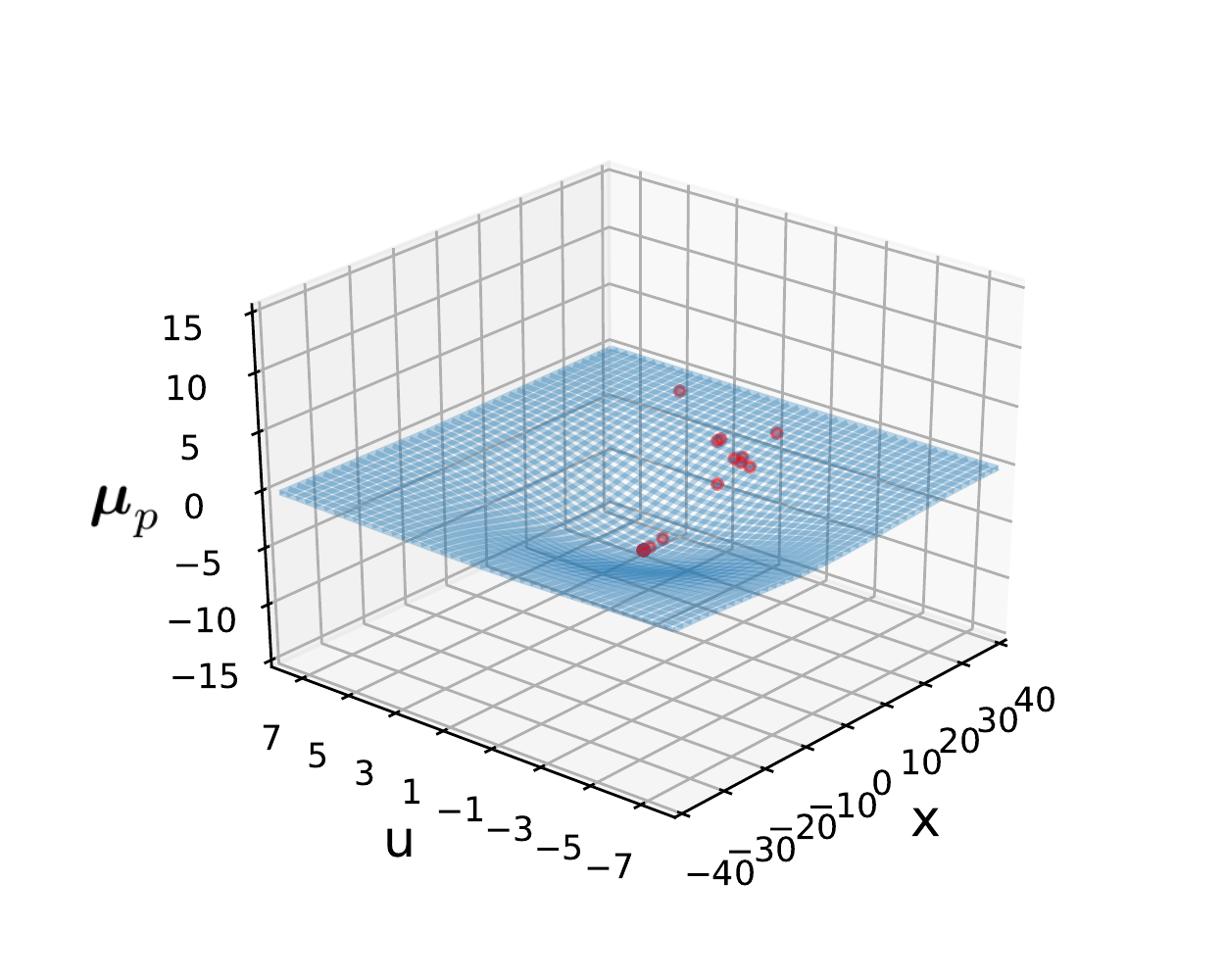}}
	\vspace*{-0.3cm}
	\subfigure[Step 5]{
		\includegraphics[trim={.5cm 1.cm 1cm 1.5cm},clip,width=0.3\columnwidth]{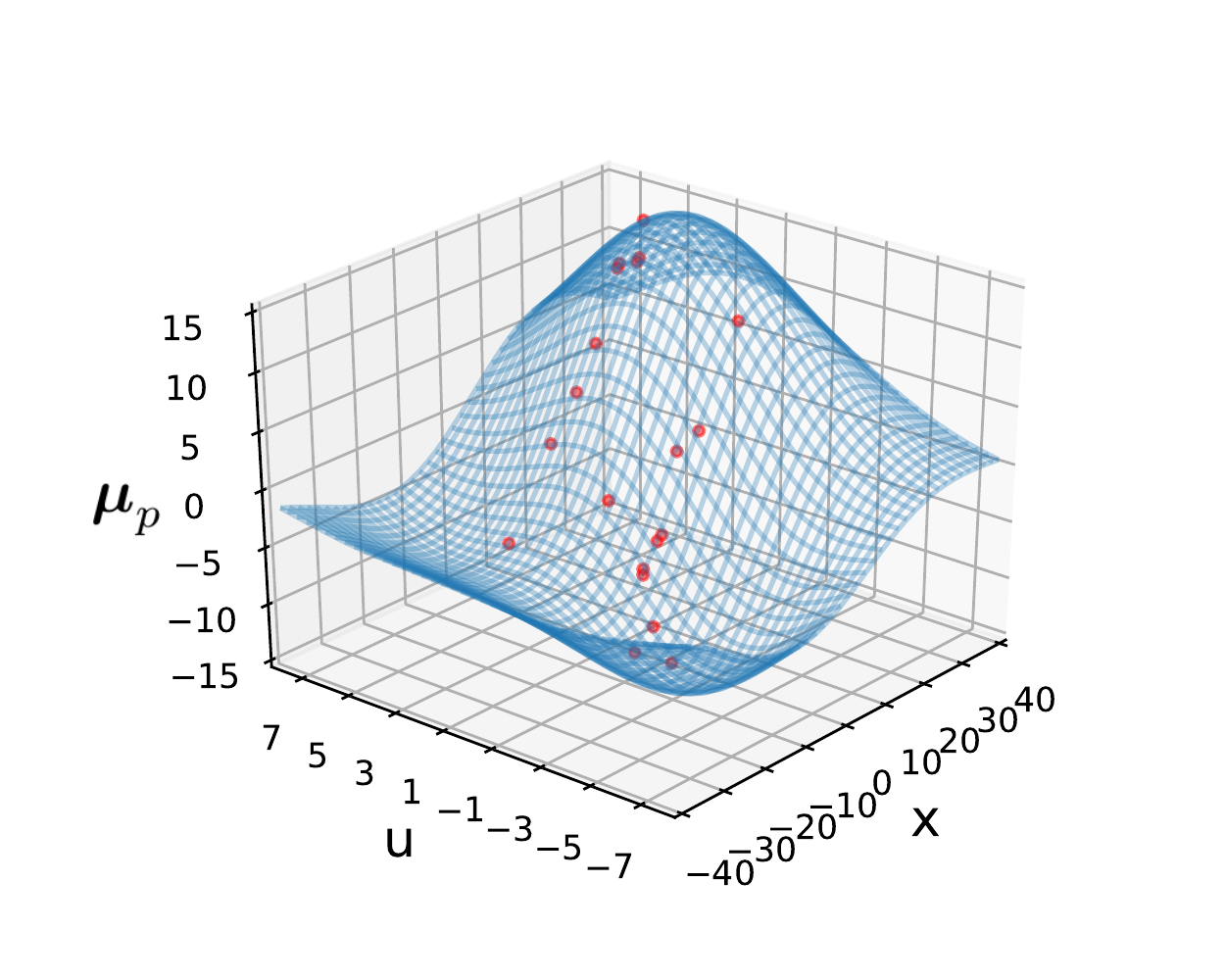}}\\
	\subfigure[Step 5]{
		\includegraphics[trim={.5cm 1.cm 1cm 1.5cm},clip,width=0.30\columnwidth]{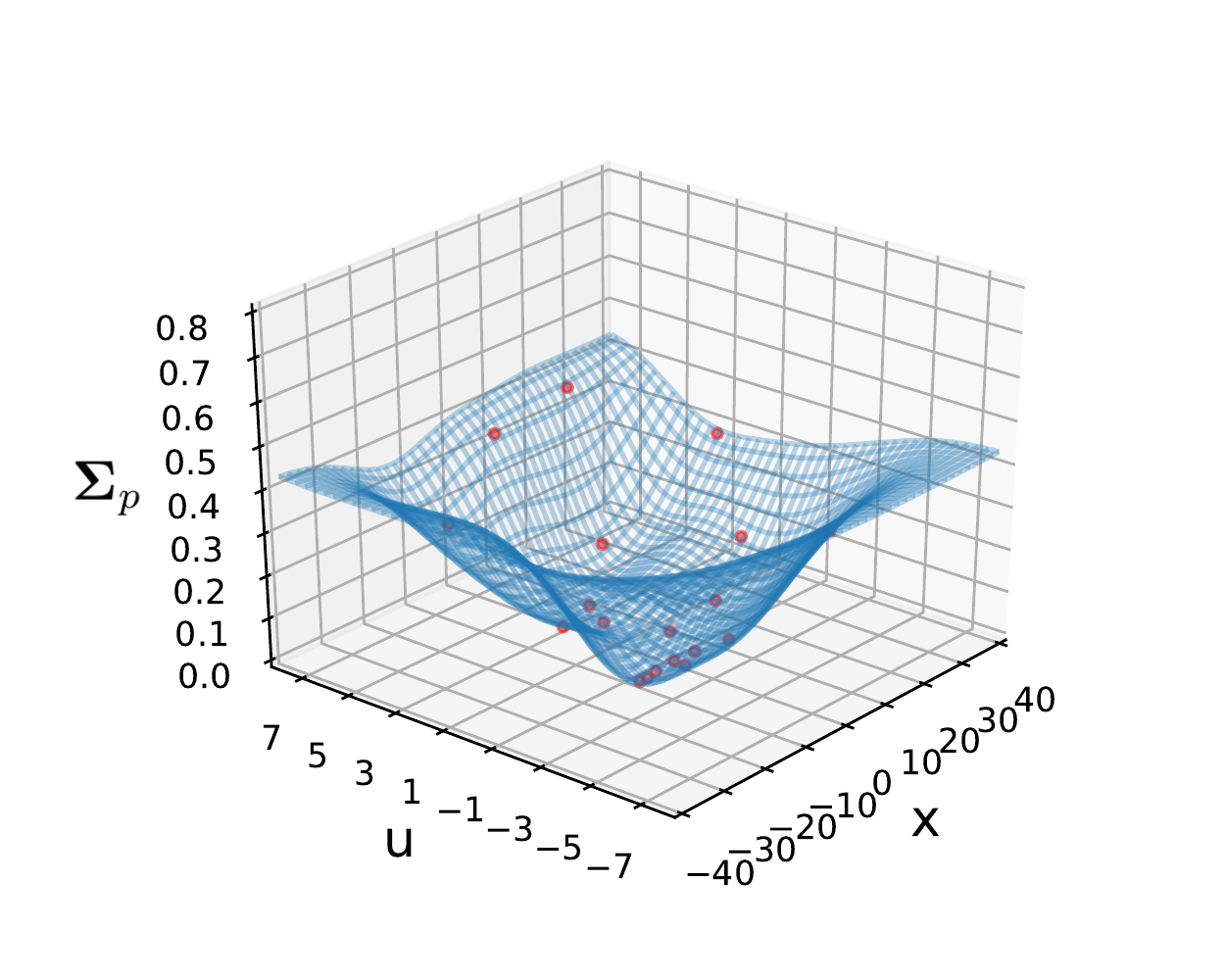}}
	\vspace*{-0.3cm}
	\subfigure[Step 5]{
		\includegraphics[trim={.5cm 1.cm 1cm 1.5cm},clip,width=0.30\columnwidth]{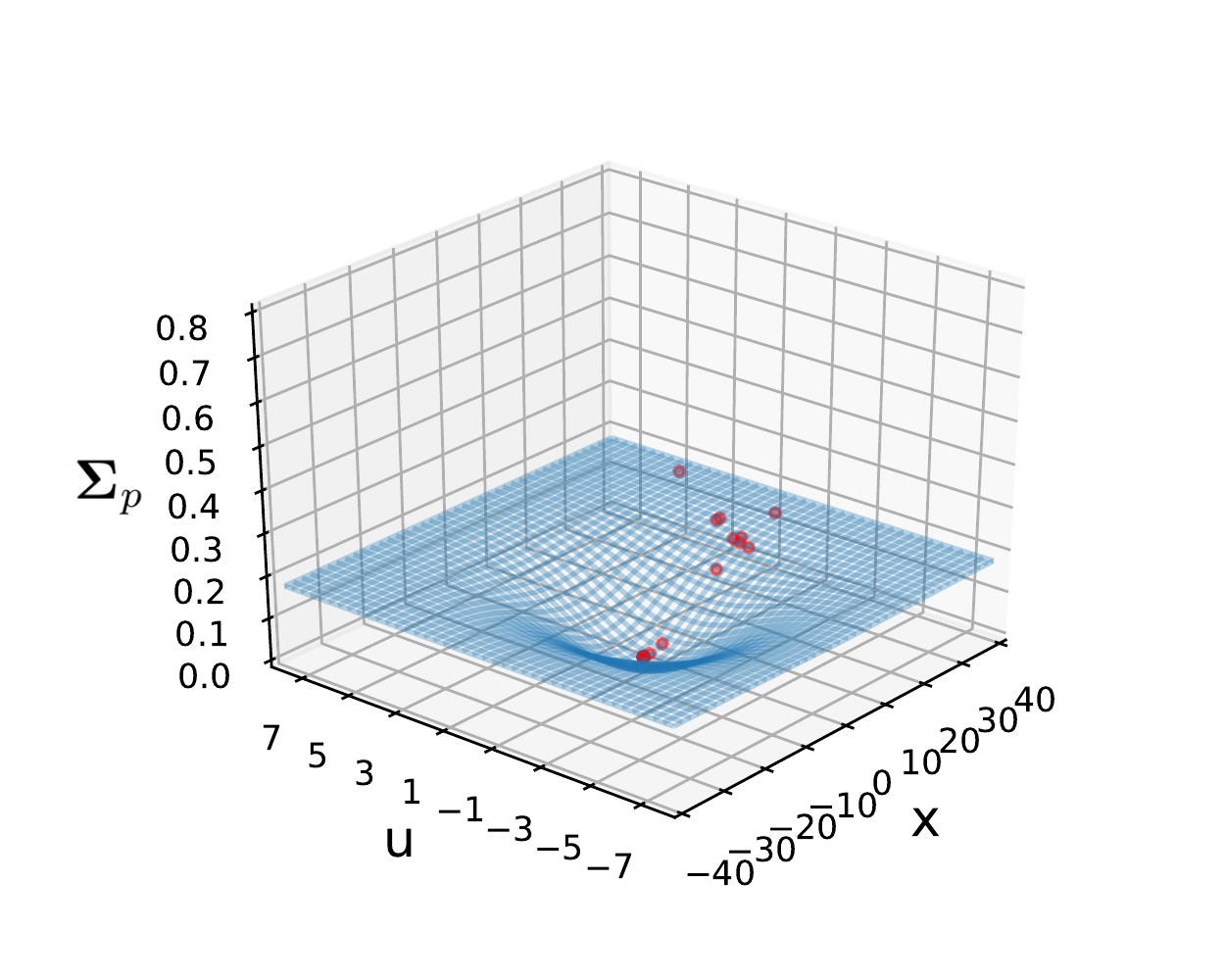}}
	\subfigure[Step 5]{
		\includegraphics[trim={.5cm 1.cm 1cm 1.5cm},clip,width=0.30\columnwidth]{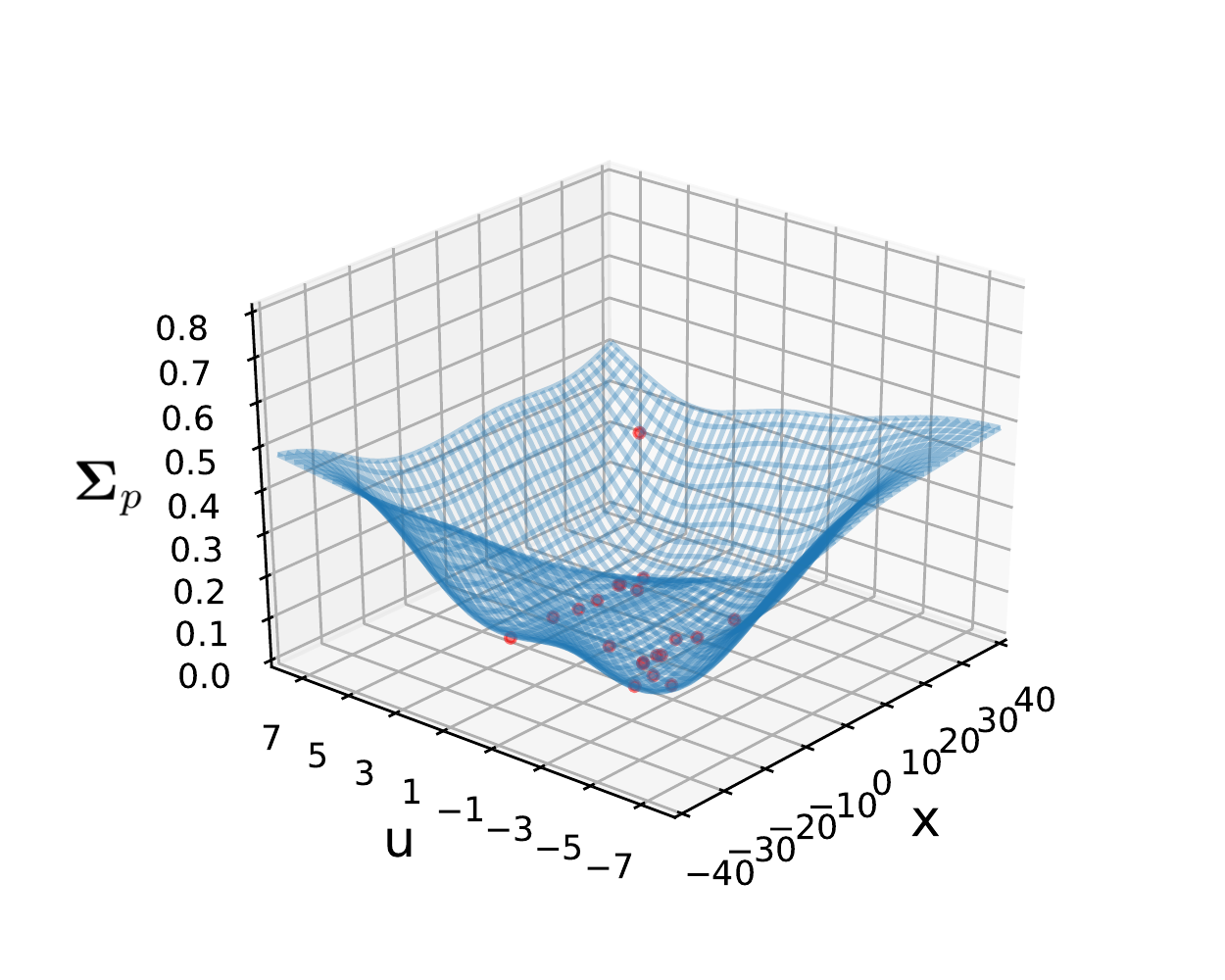}}
	\caption{Prediction results of (a)(d) VB ($\mathcal{M}$=20), (b)(e) PRSSM ($\mathcal{M}$=20), (c)(f) PRSSM ($\mathcal{M}$=$\infty$) after 5-step online learning.}
\end{figure}

\end{document}